\documentclass[10pt,twocolumn,letterpaper]{article}
\usepackage[pagenumbers]{cvpr}
\usepackage{cuted}

\usepackage{currfile}

\usepackage{caption}

\usepackage{tabularx}
\usepackage{multirow}
\usepackage{microtype}
\usepackage{placeins}
\usepackage{iftex}
\iftutex
  \usepackage{fontspec}
\fi
\usepackage{amsmath,amsfonts,bm}

\def\1{\bm{1}}

\DeclareMathAlphabet{\mathsfit}{\encodingdefault}{\sfdefault}{m}{sl}
\SetMathAlphabet{\mathsfit}{bold}{\encodingdefault}{\sfdefault}{bx}{n}

\usepackage{array,booktabs,tabularx,multirow,colortbl,stfloats}
\usepackage{graphicx}
\usepackage{pifont,enumitem,needspace,float}

\newlist{contributions}{enumerate}{1}
\setlist[contributions]{label=\protect\ding{\numexpr181+\value{contributionsi}\relax},
  leftmargin=1.45em,labelsep=0.4em,itemsep=3pt,topsep=4pt,parsep=0pt}
\expandafter\newcommand\csname OmniLogo@gemini\endcsname{\includegraphics[height=1.00em]{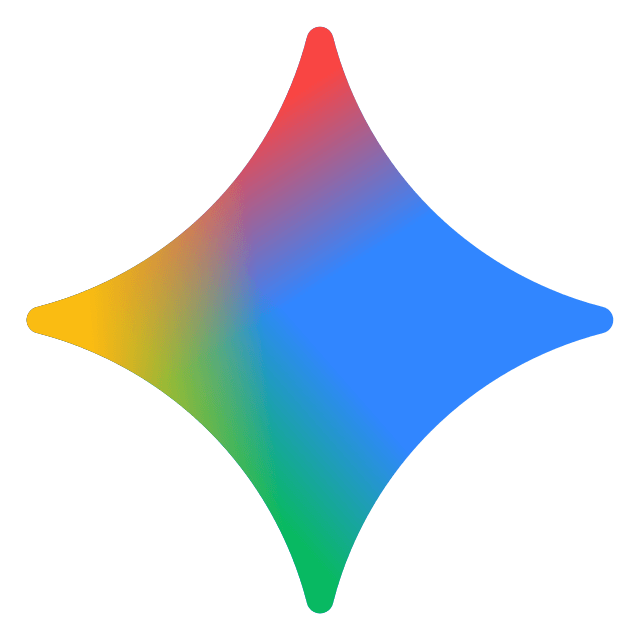}}
\expandafter\newcommand\csname OmniLogo@gemma\endcsname{\includegraphics[height=1.00em]{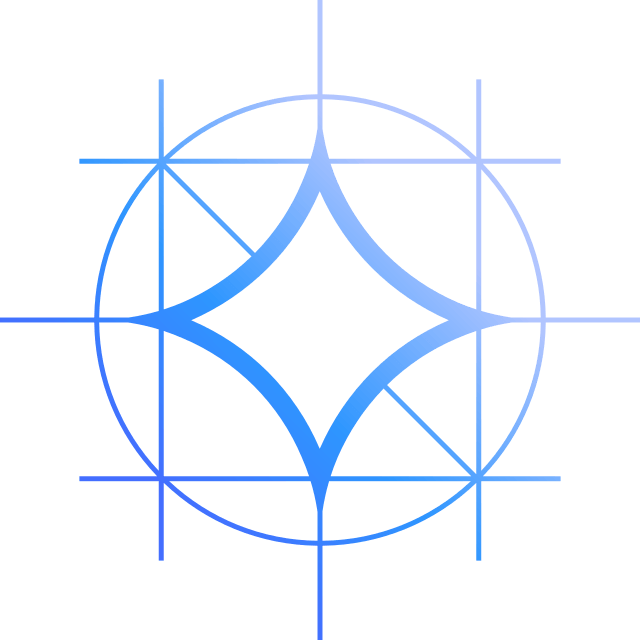}}
\expandafter\newcommand\csname OmniLogo@meta\endcsname{\includegraphics[height=1.00em]{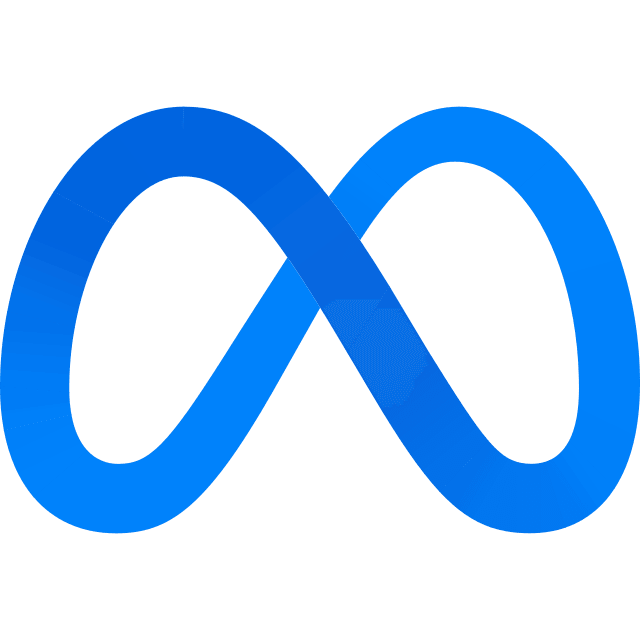}}
\expandafter\newcommand\csname OmniLogo@openai\endcsname{\includegraphics[height=1.00em]{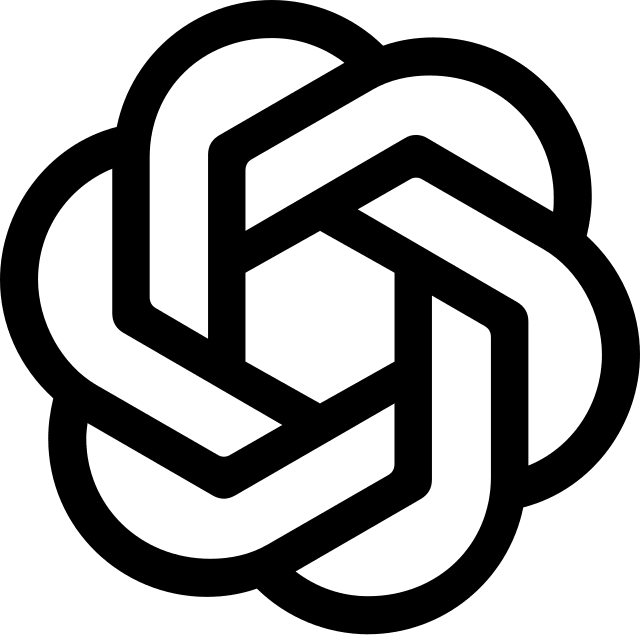}}
\expandafter\newcommand\csname OmniLogo@qwen\endcsname{\includegraphics[height=1.00em]{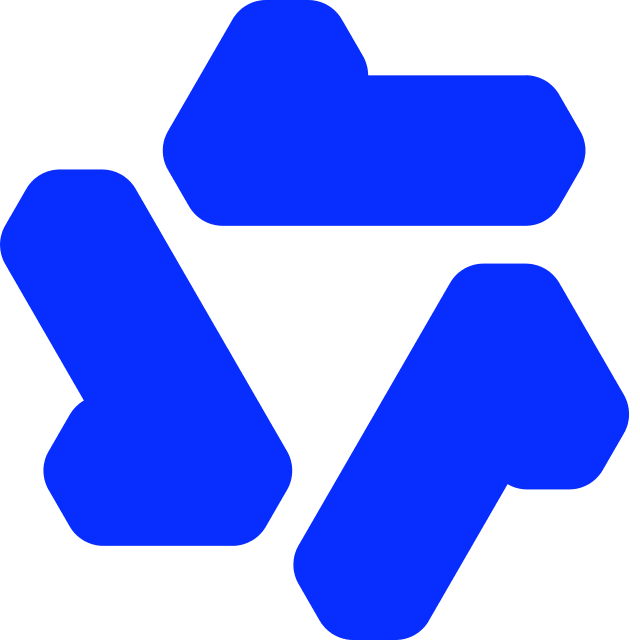}}
\newcommand{\OmniModelLogo}[1]{%
  \nobreak\hspace{0.35em}%
  \raisebox{-0.15em}{\csname OmniLogo@#1\endcsname}%
  \nobreak\hspace{0.15em}%
}
\usepackage[breakable]{tcolorbox}
\newtcolorbox{promptcard}{
  breakable,
  colback=black!4,
  colframe=black!30,
  arc=2.5mm,
  boxrule=0.6pt,
  left=11pt, right=11pt, top=8pt, bottom=9pt,
  before skip=4pt, after skip=12pt
}
\newenvironment{promptlist}{%
  \begin{list}{\textbullet}{%
    \setlength{\leftmargin}{1.15em}%
    \setlength{\itemsep}{1.5pt}%
    \setlength{\parsep}{0pt}%
    \setlength{\topsep}{3pt}%
    \setlength{\partopsep}{0pt}%
  }%
}{\end{list}}
\definecolor{omniHeader}{RGB}{235,235,235}
\definecolor{omniStripe}{RGB}{246,246,248}
\definecolor{omniOurs}{RGB}{255,252,229}
\newcommand{\OmniTableSetup}{\setlength{\tabcolsep}{4pt}\renewcommand{\arraystretch}{1.13}\arrayrulecolor{black!70}}
\newcommand{\OmniGroup}[2]{\rowcolor{white}\multicolumn{#1}{@{}l}{\itshape #2}\\}
\newcommand{\OmniCheck}{\ensuremath{\checkmark}}
\newcommand{\OmniMeanStd}[2]{\mbox{#1\ensuremath{{}_{\scriptscriptstyle\pm#2}}}}
\definecolor{cvprblue}{rgb}{0.21,0.49,0.74}
\usepackage[pagebackref,breaklinks,colorlinks,allcolors=cvprblue]{hyperref}
\title{See Before You Code: Learning Visual Priors and Spatial Grounding for Educational Animation Generation}
\author{Yuejia Li\textsuperscript{1,2,*,\textdagger}\quad
Ke He\textsuperscript{3,4,1,*}\quad
Jingkang Xia\textsuperscript{1}\quad
Shutong Chen\textsuperscript{5}\quad
Zhiyue Su\textsuperscript{1}\\
Junchi Zhang\textsuperscript{1}\quad
JunHeng Li\textsuperscript{1}\quad
Haoyu Wang\textsuperscript{6}\quad
Canran Xiao\textsuperscript{7}\quad
Pengfei Zhou\textsuperscript{8}\\
Xiyuan Yang\textsuperscript{9}\quad
Xin Sun\textsuperscript{10}\quad
Junjie Huang\textsuperscript{11}\quad
Bj\"orn Schuller\textsuperscript{12}\quad
Jiachen Luo\textsuperscript{13}\\[0.5em]
{\normalsize \textsuperscript{1}Wuhan University\quad
\textsuperscript{2}University of California, Berkeley\quad
\textsuperscript{3}Microsoft\quad
\textsuperscript{4}Synesis}\\
{\normalsize \textsuperscript{5}University of the Chinese Academy of Sciences\quad
\textsuperscript{6}Nankai University\quad
\textsuperscript{7}Sun Yat-sen University}\\
{\normalsize \textsuperscript{8}National University of Singapore\quad
\textsuperscript{9}University of Illinois at Urbana-Champaign}\\
{\normalsize \textsuperscript{10}Xiaohongshu\quad
\textsuperscript{11}University of Science and Technology of China}\\
{\normalsize \textsuperscript{12}Technical University of Munich\quad
\textsuperscript{13}Queen Mary, University of London}\\
{\normalsize \textsuperscript{*}Equal contribution\quad
\textsuperscript{\textdagger}Corresponding author: {\tt\small yuejiali@berkeley.edu}}}

\begin{document}
\maketitle
\setlength{\parfillskip}{0pt plus 0.7\columnwidth}
\begin{abstract}
Large language models can generate executable code for educational animations, yet the rendered outputs often suffer from viewport overflow, element overlap, remnant content, and scale mismatch. These failures expose gaps in planning valid layouts, implementing visual constraints in code, and repairing errors revealed by rendering. We propose \textbf{OmniManim}, an educational animation generation framework that integrates explicit visual planning, spatially grounded code learning, and render-feedback-aware repair. Its \textbf{Vision Agent} predicts sparse keyframe layouts through coarse-to-fine bounding-box denoising with interpolation-aware constraints. \textbf{S}patially \textbf{G}rounded \textbf{C}ode \textbf{S}upervision (\textbf{SGCS}) aligns plan-conditioned code generation and diagnostic-guided repair with these layouts. \textbf{F}ailure-\textbf{L}ocalized \textbf{R}ender \textbf{P}olicy \textbf{O}ptimization (\textbf{FLRPO}) associates rendered failures with objects, time intervals, and code regions, and optimizes local patches using measured rendering gains, global constraint preservation, and edit costs. Together, these components connect visual plans to executable code and targeted self-repair. We also construct \textbf{ManimLayout-1K} for visual-layout and aligned code supervision, and \textbf{EduRequire-500} for evaluating executability, instructional quality, visual quality, and efficiency. Experiments on EduRequire-500, TheoremExplainBench, and VisualEDU cover two agent-only backbones (GPT-5.4 and Gemini) and three SGCS/FLRPO-trained backbones (Qwen3-14B, Gemma, and Llama). On EduRequire-500, agent-only GPT-5.4 improves Manim animation generation over Code2Video, raising final render success from 0.923 to 0.986 and layout quality from 0.742 to 0.931, and reducing time from 148\,s to 72\,s.
\end{abstract}

\section{Introduction}
The demand for high-quality educational animations far exceeds the capacity of manual production. End-to-end video generation models~\citep{ho2022imagenvideo,singer2023makeavideo,blattmann2023alignyourlatents,kondratyuk2023videopoet,bar2024lumiere} achieve impressive visual fidelity, but they function as black-box generators with limited control over symbolic content, logical structure, and compositional precision, all of which are essential in educational settings. Recent studies have therefore turned to \emph{code-centric generation}, in which large language models (LLMs) produce executable Manim~\citep{manim2025} scripts rather than pixel-level videos~\citep{chen2025code2video,ku2025theoremexplainagent}, enabling deterministic rendering, precise control, and editable outputs; benchmarks such as ManimBench~\citep{silva2026manim} and ManiBench~\citep{nabin2026manibench} further formalize execution-level evaluation in this paradigm.
\begin{figure*}[t]
\centering\includegraphics[width=\linewidth]{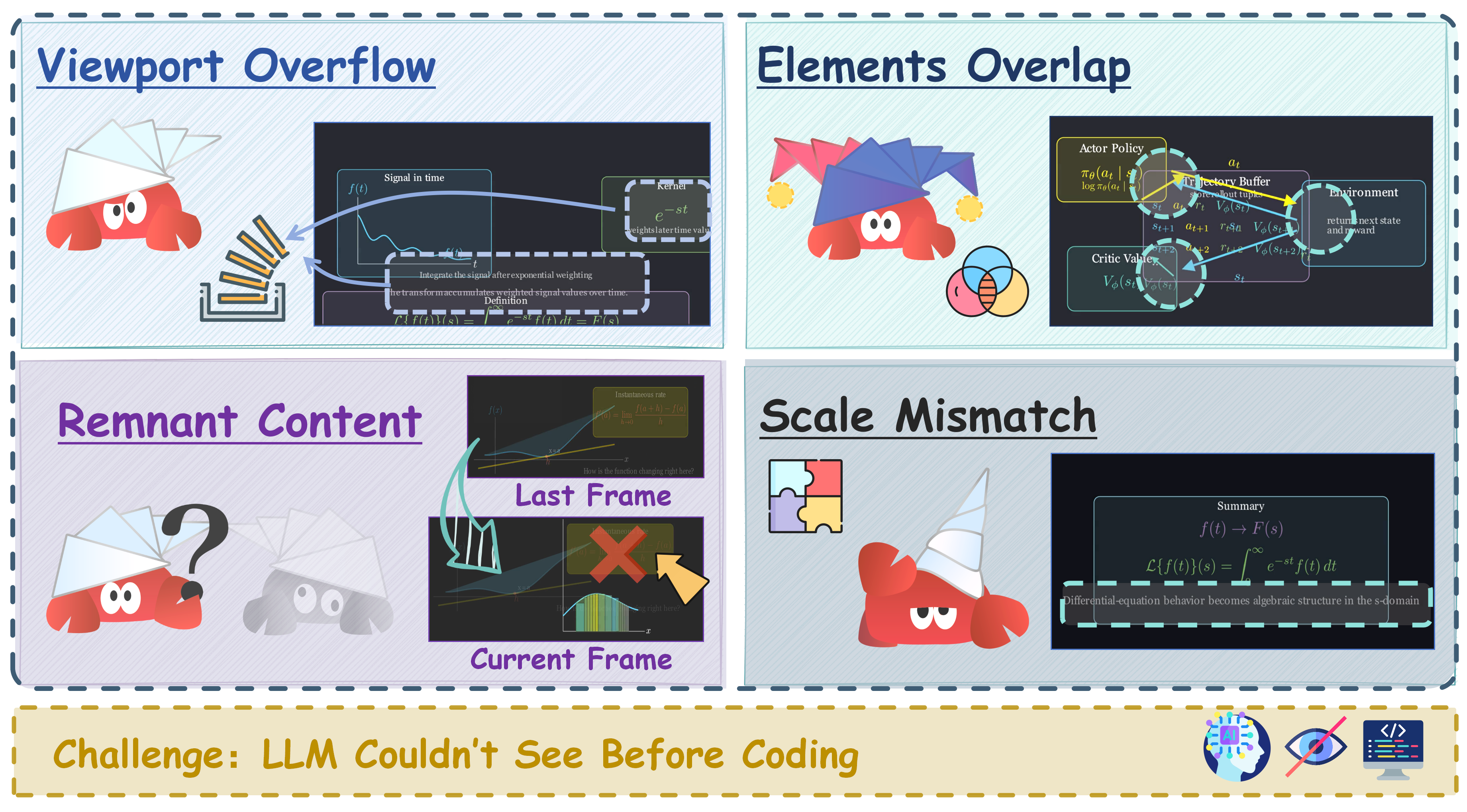}
\caption{Common render-time failure modes in LLM-generated educational animations. Although the generated code is executable, the rendered outputs still exhibit viewport overflow, element overlap, remnant content from previous frames, and scale mismatch. These defects are difficult to detect from code alone and motivate our render-feedback-aware formulation.}\label{fig:motivation}
\end{figure*}
However, \emph{executability does not imply rendering quality}. Figure~\ref{fig:motivation} shows viewport overflow, element overlap, remnant content, and scale mismatch that become evident only after rendering. Three gaps arise: a model may fail to plan a valid layout, fail to implement an otherwise valid plan, or fail to identify and repair the code responsible for a rendered defect. Explicit visual planning addresses the first gap, but supplying coordinates in a prompt does not ensure that a code generator respects object sizes, relations, and timing. Likewise, global visual critique provides limited supervision on which objects, time intervals, and code blocks should change.
OmniManim separates spatial planning from code synthesis through a task-specific \textbf{Vision Agent}. Two valid endpoint layouts can still collide during interpolation; the planner therefore predicts sparse keyframe boxes using coarse-to-fine denoising and an interpolation-aware objective. We couple this planner with a code policy that learns to implement and repair programs under visual constraints. The policy is first trained with \textbf{S}patially \textbf{G}rounded \textbf{C}ode \textbf{S}upervision (\textbf{SGCS}) to implement visual plans and imitate verified local repairs. It is then optimized with \textbf{F}ailure-\textbf{L}ocalized \textbf{R}ender \textbf{P}olicy \textbf{O}ptimization (\textbf{FLRPO}), which uses GRPO~\citep{shao2024deepseekmath} to optimize failure-conditioned patches under a localized rendering reward that scores the effect of each patch on the implicated object and time interval while preserving unrelated constraints.
Our contributions are as follows:
\begin{contributions}
\item \textbf{\textit{Render-aware code generation.}} We formalize constrained animation generation: executable code must satisfy spatial and temporal requirements when rendered.
\item \textbf{\textit{Spatially grounded supervision.}} We combine interpolation-aware keyframe planning with \textbf{SGCS}, aligning visual plans with code through spatially weighted generation and diagnostic-guided repair targets.
\item \textbf{\textit{Failure-localized optimization.}} We introduce \textbf{FLRPO}, optimizing patches conditioned on objects and time intervals using measured local rendering gains, global constraint preservation checks, and explicit edit costs.
\item \textbf{\textit{Training and evaluation resources.}} We build \textbf{ManimLayout-1K} and \textbf{EduRequire-500}, deriving aligned code examples and verified repairs exclusively from training sources, excluding evaluation data.
\item \textbf{\textit{Evaluation across multiple models.}} We test agent-only GPT-5.4 and Gemini and SGCS/FLRPO-trained Qwen3-14B~\citep{yang2025qwen3}, Gemma, and Llama across EduRequire-500, \textbf{TheoremExplainBench}, and \textbf{VisualEDU}.
\end{contributions}

\section{Related Work}
\paragraph{Educational Visualization.}
Code2Video revises Manim from a vision-language critique of rendered frames and grid anchors; TheoremExplainAgent retries after execution errors but lacks a layout critic on rendered frames~\citep{chen2025code2video,ku2025theoremexplainagent}. These methods do not learn sparse keyframe boxes, penalize interpolation collisions, or train failure-localized repair. ManimBench and ManiBench benchmark Manim generation~\citep{silva2026manim,nabin2026manibench}. Beyond Manim, DeTikZify~\citep{belouadi2024detikzify} synthesizes TikZ figures, DiagrammerGPT~\citep{zala2023diagrammergpt} plans diagrams, and EduVisAgent~\citep{ji2025eduvisagent} coordinates pedagogical visualization. Plot2Code, ChartMimic, and StarVector translate rendered charts or vector graphics into executable code~\citep{wu2025plot2code,shi2025chartmimic,rodriguez2023starvector}.
\paragraph{Agentic Refinement.}
ChatDev~\citep{qian2024chatdev}, MetaGPT~\citep{hong2024metagpt}, and AutoGen~\citep{wu2023autogen} support specialized agent coordination. Self-Debugging~\citep{chen2023teaching}, Reflexion~\citep{shinn2023reflexion}, Self-Refine~\citep{madaan2023selfrefine}, and ReAct~\citep{yao2023react} guide correction through execution, self-feedback, or tools; MatPlotAgent~\citep{yang2024matplotagent} adds VLM verification. Toolformer~\citep{schick2023toolformer} studies learned tool use, while chain-of-thought and self-consistency improve reasoning~\citep{wei2022cot,wang2023selfconsistency}. We build on these capabilities, with explicit keyframe geometry and intermediate-frame safety as additional modeling targets.
\paragraph{Code and Repair Learning.}
CodeRL, PPOCoder, RLTF, and RLEF optimize programs with execution feedback~\citep{le2022coderl,shojaee2023ppocoder,liu2023rltf,gehring2025rlef}, and GRPO~\citep{shao2024deepseekmath} supplies the group-relative optimizer used below. OmniManim separates supervision for plan-following code from render-feedback-aware optimization of local repairs. SGCS aligns visual plans with code; FLRPO defines failure-conditioned patches and their rewards based on measured rendering outcomes.
\paragraph{Layout Generation.}
LayoutGAN, LayoutVAE, and LayoutTransformer generate static layouts~\citep{li2019layoutgan,jyothi2019layoutvae,gupta2021layouttransformer}. Diffusion approaches include two LayoutDM models~\citep{inoue2023layoutdm,chai2023layoutdm}, LACE~\citep{chen2024lace}, and Zhang et al.'s LayoutDiffusion for graphic layout generation~\citep{zhang2023graphiclayoutdiffusion}. Zheng et al.'s distinct LayoutDiffusion generates images conditioned on layouts~\citep{zheng2023layoutdiffusion}. LayoutGPT and LayoutPrompter prompt an LLM for static layouts~\citep{feng2023layoutgpt,lin2023layoutprompter}; VideoDirectorGPT and LLM-grounded Video Diffusion plan time-varying boxes for pixel video~\citep{lin2024videodirectorgpt,lian2024lvd}. These methods do not emit executable animation code. OmniManim learns object-level boxes, penalizes sampled interpolation collisions, and trains plan-following code and localized repairs with SGCS and FLRPO.

\section{Method}
\subsection{Problem Formulation}
Given a teaching requirement $r$, OmniManim generates executable animation code $c$ that satisfies instructional, spatial, and temporal constraints. A scene parser $\psi$ extracts objects, relations, pedagogical roles, and schedules into $s$, and a visual planner $v_\phi$ predicts sparse keyframe boxes $\mathbf B$:
\begin{equation}\begin{aligned}
z&=(r,s,\mathbf B),\qquad s=\psi(r),\quad \mathbf B=v_\phi(s),\\
c&\sim\pi_\theta(\cdot\mid z,\texttt{generate}).
\end{aligned}\end{equation}
Rendering yields a diagnostic $d$ that conditions a local patch $p\sim\pi_\theta(\cdot\mid z,c,d,\texttt{repair})$, producing $c'=\operatorname{Apply}(c,p)$. Generation and repair share policy parameters in trained configurations, distinguishing the two tasks. Agent-only configurations use the same inference interfaces without SGCS or FLRPO updates to the code-and-repair backbone.
\subsection{Framework Overview}
\begin{figure*}[t]\centering\includegraphics[width=0.90\linewidth]{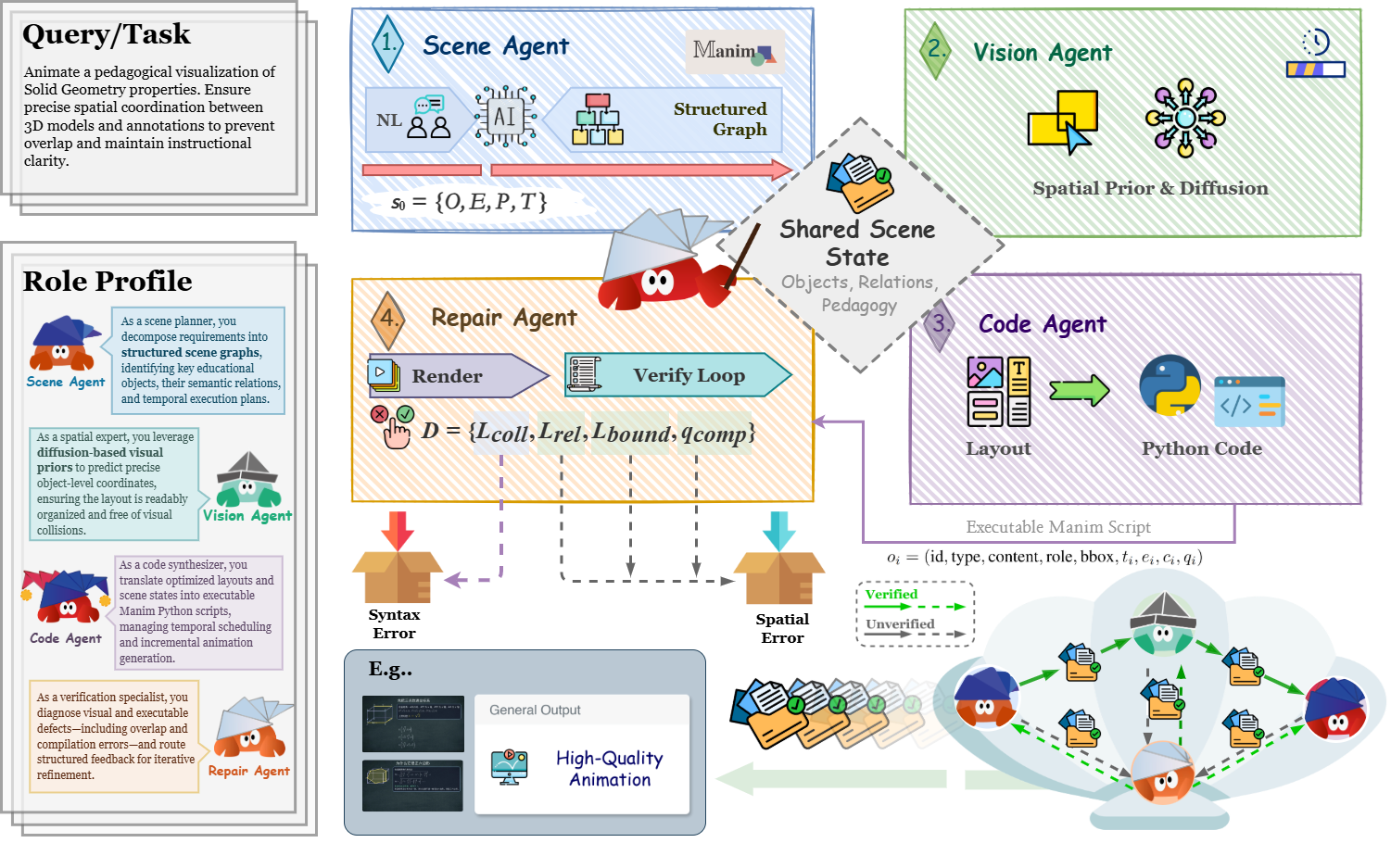}
\caption{OmniManim's inference pipeline. The Scene and Vision Agents specify semantics and keyframe layouts; the Code and Repair Agents generate and revise executable programs through a shared scene state.}\label{fig:overview}\end{figure*}
OmniManim connects four agents through a \textbf{Shared Scene State} containing object identities, content, roles, layouts, schedules, code blocks, and verification status (see Figure~\ref{fig:overview}). The \textbf{Scene Agent} parses requirements, the \textbf{Vision Agent} predicts layouts, the \textbf{Code Agent} implements them, and the \textbf{Repair Agent} patches failures identified by compilation and rendering. Scene programs are written and rendered concurrently, then concatenated into one video. Trained configurations additionally optimize the code-and-repair policy with SGCS and FLRPO. For each normalized box $(x_c,y_c,w_b,h_b)$, the Code Agent maps its center to $X=(x_c-0.5)W_f$, $Y=(0.5-y_c)H_f$, where $(W_f,H_f)=(14.22,8.0)$, and uniformly scales an object of native size $(W_i,H_i)$ by $\min(w_b W_f/W_i,h_b H_f/H_i)$, preserving its aspect ratio and identity across keyframes.
\paragraph{Deployment.}\label{subsec:deployment}
The scene and layout plans condition program generation. Compiler failures supply error evidence; rendered overlap, relation, and boundary violations are flagged at thresholds of $0.02$, $0.15$, and $0.01$, respectively. A VLM reviews flagged start/mid/end keyframes: \texttt{FAIL} triggers a patch, whereas \texttt{PASS} and \texttt{INTENTIONAL} preserve the output. Deployment uses fixed parameters and at most one repair attempt with one candidate. Group sampling is restricted to training. Offline evaluation scores never guide deployment stopping decisions or candidate reranking.
\subsection{Vision Agent}\label{subsec:vision_agent}
The Vision Agent predicts normalized center--size boxes $\mathbf B=\{b_i^{(k)}\}$ at sparse keyframes. Object tokens $\mathbf O$ condition a prior $\mathbf P\in[0,1]^{4\times32\times32}$ for title, figure, equation, and annotation regions, supervised by Gaussian heatmaps derived from ground-truth boxes. A Transformer attends to objects and prior features to regress velocities on \(\mathbf{b}^{\tau}=(1-\tau)\mathbf{b}^{\star}+\tau\boldsymbol{\epsilon}\), with \(\boldsymbol{\epsilon}\sim\mathcal{N}(0,\mathbf{I})\) and \(\tau\in(0,1]\)~\citep{lipman2023flowmatching,liu2023rectifiedflow}. From Gaussian noise, inference takes $T_d=16$ projected Euler steps, reusing the denoiser parameters throughout:
\begin{equation}\begin{aligned}
\hat{\mathbf{v}}^{t}&=\Delta_{\phi}\!\left(\mathbf{b}^{t},\mathbf{O},\mathbf{P},t/T_d\right),\\
\mathbf{b}^{t-1}&=\mathrm{clamp}\!\left(\mathbf{b}^{t}-\frac{1}{T_d}\hat{\mathbf{v}}^{t},\,0,\,1\right).
\end{aligned}\end{equation}
Clamping bounds parameter values but does not guarantee exact target recovery or full-box containment. Between keyframes, we sample $\mathcal{U}=\{0.25,0.50,0.75\}$ and define
\[
\tilde b_i^k(u)=(1-u)b_i^{(k)}+u b_i^{(k+1)},\qquad u\in\mathcal U.
\]
Let $\rho(b,b')=[\mathrm{IoU}(b,b')-\tau_{\mathrm{coll}}]_+$. Pair sets $\mathcal E_k$ and $\mathcal E_k(u)$ exclude absent objects and permitted overlaps; interpolation includes only matched objects present at both endpoints:
\begin{equation}\begin{aligned}
\mathcal L_{\mathrm{coll}}&=\sum_k\sum_{(i,j)\in\mathcal E_k}\rho(b_i^{(k)},b_j^{(k)}),\\
\mathcal L_{\mathrm{interp}}&=\sum_{k=1}^{K-1}\sum_{u\in\mathcal U}\sum_{(i,j)\in\mathcal E_k(u)}\rho(\tilde b_i^k(u),\tilde b_j^k(u)).
\end{aligned}\end{equation}
These sampled constraints reduce overlap risk without certifying rendered motion. The Stage~2 objective is
\begin{equation}\begin{aligned}
\mathcal{L}_{\mathrm{vision}}={}&\mathcal{L}_{\mathrm{diff}}+\lambda_1\mathcal{L}_{\mathrm{box}}+\lambda_2\mathcal{L}_{\mathrm{coll}}\\
&+\lambda_3\mathcal{L}_{\mathrm{rel}}+\lambda_4\mathcal{L}_{\mathrm{bound}}+\lambda_5\mathcal{L}_{\mathrm{interp}}.
\end{aligned}\end{equation}
With $\mathbb E_a$ averaging over valid objects and training samples, and $\mathbf b^0$ denoting the 16-step prediction,
\begin{equation}\begin{aligned}
\mathcal{L}_{\mathrm{diff}}&=\mathbb{E}_a\|\hat{\mathbf{v}}^{t}-(\boldsymbol{\epsilon}-\mathbf{b}^{\star})\|_2^2,\\
\mathcal{L}_{\mathrm{box}}&=\mathbb{E}_a\|\mathbf{b}^{0}-\mathbf{b}^{\star}\|_2^2.
\end{aligned}\end{equation}
$\mathcal L_{\mathrm{rel}}$ retains containment and directional constraints; $\mathcal L_{\mathrm{bound}}$ penalizes box-edge overflow. Both use scheduled visibility. Supplementary Section S3 specifies the architecture, masks, prior supervision, and sampling details.
\subsection{SGCS Training}\label{subsec:sgcs}\label{subsec:policy_training}
For trained configurations, SGCS initializes the shared code-and-repair policy before FLRPO, keeping the Vision Agent frozen throughout both stages. It supervises $(z,c^\star)$ pairs whose code renders consistently with the scene and predicted layouts. Object identities align boxes with code variables, placement/scale operations, and animation events. For token $c_t^\star$, we set $w_t=1+\omega m_t$, where $\omega\geq0$ and $m_t$ marks aligned layout, timing, and object-reference tokens. The resulting weighted generation objective is defined as
\begin{equation}
\mathcal L_{\mathrm{code}}=-\mathbb E_{(z,c^\star)}\left[\frac{\sum_t w_t\log\pi_\theta(c_t^\star\mid z,c_{<t}^\star)}{\sum_t w_t}\right].\label{eq:sgcs_code}
\end{equation}
Prompt tokens are excluded, and numeric tokens are weighted only when their aligned roles warrant it. Repair examples $(z,c,d,p^\star)$ pair a failing program and diagnostic with a local patch verified by rendering and preservation checks. Under the repair task marker, we average over verified records and combine both training objectives:
\begin{equation}\begin{aligned}
\mathcal L_{\mathrm{repair}}&=-\mathbb E\left[\frac{1}{|p^\star|}\sum_t\log\pi_\theta(p_t^\star\mid z,c,d,p_{<t}^\star)\right],\\
\mathcal L_{\mathrm{SGCS}}&=\mathcal L_{\mathrm{code}}+\lambda_{\mathrm{repair}}\mathcal L_{\mathrm{repair}}.
\end{aligned}\label{eq:sgcs_joint}\end{equation}
SGCS teaches plan implementation and local repair from verified examples. FLRPO then optimizes the rendered effects of patches while preserving unrelated constraints.
\subsection{FLRPO Training}\label{subsec:flrpo}
FLRPO optimizes patches for a fixed failure state $x=(z,c,d)$. The diagnostic $d=(\mathcal I,\mathcal T,\Omega,e)$ identifies implicated objects, a time interval, candidate code regions, and error evidence. 
Renderer event traces and object-to-code mappings link each defect to the placement, transformation, and lifecycle operations in the abstract syntax tree (AST). If a defect maps to several candidate regions, the sample is flagged as ambiguous and excluded from training. The behavior policy samples $M$ patches $p_j\sim\pi_{\mathrm{old}}(\cdot\mid x)$, each specifying an edit region and replacement code. Each is applied independently to the same $c$ to obtain $c_j'$, and each patch covers the implicated objects and their necessary dependencies. For compile-time failures, $e$ is the compiler traceback, $\Omega$ the AST nodes it references, and $\mathcal I,\mathcal T$ are empty; the local reward term is then dropped. Each candidate is executed in Manim. Training diagnostics score the implicated objects and interval through $Q_{\mathrm{local}}(c;d)$ and the full animation through $Q_{\mathrm{global}}(c;z)$. The local and global quality gains are
\begin{equation}\begin{aligned}
\Delta Q_{\mathrm{local},j}&=Q_{\mathrm{local}}(c_j';d)-Q_{\mathrm{local}}(c;d),\\
\Delta Q_{\mathrm{global},j}&=Q_{\mathrm{global}}(c_j';z)-Q_{\mathrm{global}}(c;z).
\end{aligned}\end{equation}
Let $G(c_j')\in\{0,1\}$ check executability, required content, and instructional order. The patch reward balances these gains against edit costs and invalid outputs:
\begin{equation}\begin{aligned}
R_j={}&G(c_j')\big[\alpha\Delta Q_{\mathrm{local},j}+\beta\Delta Q_{\mathrm{global},j}\big]\\
&-\lambda_{\mathrm{edit}}C_{\mathrm{edit}}(p_j;c,d)-\lambda_{\mathrm{fail}}\big[1-G(c_j')\big].
\end{aligned}\label{eq:flrpo_reward}\end{equation}
Here, $C_{\mathrm{edit}}$ penalizes normalized patch size and AST edits unrelated to $\Omega$, and $\lambda_{\mathrm{fail}}>0$ penalizes invalid candidates. Targets are fixed before sampling, so deleting an object or shifting a defect outside sampled frames cannot evade lifecycle and visibility checks; global checks cover other relations, lifecycles, and event order. Non-renderable starting programs receive a predefined failure-floor score. Coefficients and checks are fixed on training/validation data, independently of held-out visual scores and human judgments.
We optimize with GRPO~\citep{shao2024deepseekmath}, initialized from $\pi_{\mathrm{SGCS}}$. Within each failure state, rewards are normalized as $\hat A_j=(R_j-\bar R)/(\sigma_R+\epsilon_A)$, and the token likelihood ratio is $\rho_{j,t}(\theta)=\pi_\theta(p_{j,t}\mid x,p_{j,<t})/\pi_{\mathrm{old}}(p_{j,t}\mid x,p_{j,<t})$. We combine clipped policy improvement with token-level regularization in the following objective:
\begin{equation}\begin{aligned}
J(\theta)&=\mathbb E\!\left[\frac{1}{M}\sum_{j=1}^M\frac{1}{|p_j|}\sum_t g_{j,t}(\theta)\right],\\
g_{j,t}(\theta)&=\min\!\left(\rho_{j,t}\hat A_j,\,\bar\rho_{j,t}\hat A_j\right)-\eta_{\mathrm{KL}}D_{j,t},\\
\bar\rho_{j,t}&=\operatorname{clip}(\rho_{j,t},1-\epsilon_c,1+\epsilon_c),
\end{aligned}\label{eq:flrpo_grpo}\end{equation}
where $D_{j,t}$ penalizes token-level KL divergence from $\pi_{\mathrm{SGCS}}$. Equal-reward groups have zero relative advantage. Comparing patches for the same program and target localizes the learning signal. GRPO supplies the optimizer; FLRPO defines conditioning on the localized failure, local patch actions, and rendering-based training rewards.

\section{Datasets}
\label{sec:dataset}

We construct two datasets for training and evaluation. \textbf{ManimLayout-1K} contains 1{,}000 educational animations selected from 3{,}246 research-permitted GitHub/community scripts, excluding 1{,}418 rendering failures, 527 non-educational scripts, and 301 visually inadequate samples. Keyframe extraction yields 22{,}579 scene--layout records from 2{,}968 scenes, split into 17{,}773/2{,}681/2{,}125 training/validation/test records at the source-animation level. A 200-keyframe audit finds 93.5\% acceptable boxes and 91.0\% correct temporal matches; we correct or discard errors.

\textbf{EduRequire-500} comprises 500 expert-authored requirements spanning 120 discipline labels and 16 task types. Twelve contributors (four high-school teachers and eight university professors or PhD students) drafted approximately 720 candidate requirements, which were then independently reviewed by two reviewers not involved in drafting for correctness, clarity, and Manim feasibility. LLM assistance was limited to English polishing. A 100-item audit yielded 92\% accept/reject agreement between reviewers. Exact matching and manual inspection of the top-100 embedding-retrieved pairs revealed no confirmed duplicates or semantic near-duplicates with ManimLayout-1K. The supplementary material details data construction and filtering.

\subsection{Supervision Data}
\label{subsec:policy_data}

SGCS supervision is derived after source-level splitting. Execution traces and AST mappings align objects, events, layouts, and code. Targets must render consistently with frozen-planner layouts; mismatches are corrected through verified adaptation or discarded. Keyframe records may be aggregated into scene-level examples and do not represent complete-program or repair-trajectory counts.

Repair supervision is mined from generated failures and controlled code perturbations. Teacher or manual patches are retained only after rendering verifies target correction and preservation of unrelated constraints. Records include code, diagnostics, object/time attribution, patches, and verification outcomes. Invalid patches are excluded from positive supervision but are used as negative samples in RL. All derived examples are grouped by training source. Evaluation data are excluded from all supervision, failure mining, and reward tuning throughout both training stages.

\section{Experiments}\label{sec:experiments}
\begin{table*}[t]
\centering
\captionsetup{width=\textwidth,skip=3pt}
\caption{EduRequire-500 comparison. GPT-5.4 and Gemini use agent-only OmniManim. Qwen3-14B, Gemma, and Llama are trained with SGCS and FLRPO. All eleven metrics are defined in Supplementary Table~S1; Time is in seconds. Entries use mean$_{\pm\mathrm{SD}}$ over three seeds. Bold marks the best displayed mean within each model group.}
\label{tab:main_results}
\fontsize{7.5pt}{8.5pt}\selectfont
\OmniTableSetup
\setlength{\tabcolsep}{1pt}
\renewcommand{\arraystretch}{1.03}
\setlength{\aboverulesep}{1.5pt}
\setlength{\belowrulesep}{1.2pt}
\begin{tabularx}{\textwidth}{@{}>{\raggedright\arraybackslash}p{0.17\textwidth}*{9}{>{\centering\arraybackslash}X}rr@{}}
\toprule
\rowcolor{omniHeader}
\textbf{Method} & \multicolumn{2}{c}{\textbf{Executability} $\uparrow$}
& \multicolumn{3}{c}{\textbf{Instructional quality} $\uparrow$}
& \multicolumn{4}{c}{\textbf{Visual quality} $\uparrow$}
& \multicolumn{2}{c}{\textbf{Efficiency} $\downarrow$}\\
\rowcolor{omniHeader}
& R@1 & R@F & CA & PC & EN & OV & LQ & AC & VC & Tok & Time (s)\\
\midrule
\rowcolor{omniHeader}
\multicolumn{12}{@{}l}{\textbf{Dataset: EduRequire-500}}\\
\midrule
\OmniGroup{12}{Backbone: GPT-5.4\OmniModelLogo{openai} (OmniManim: agent-only)}
\rowcolor{omniStripe}
Direct LLM & \OmniMeanStd{0.812}{.009} & \OmniMeanStd{0.934}{.012} & \OmniMeanStd{0.612}{.017} & \OmniMeanStd{0.642}{.022} & \OmniMeanStd{0.597}{.034} & \OmniMeanStd{0.628}{.013} & \OmniMeanStd{0.778}{.021} & \OmniMeanStd{0.764}{.022} & \OmniMeanStd{0.748}{.009} & \textbf{\OmniMeanStd{56{,}328}{1{,}388}} & \OmniMeanStd{108}{4.5} \\
Self-Refine & \OmniMeanStd{0.829}{.025} & \OmniMeanStd{0.957}{.014} & \OmniMeanStd{0.698}{.021} & \OmniMeanStd{0.687}{.012} & \OmniMeanStd{0.643}{.016} & \OmniMeanStd{0.671}{.005} & \OmniMeanStd{0.826}{.018} & \OmniMeanStd{0.791}{.014} & \OmniMeanStd{0.797}{.003} & \OmniMeanStd{68{,}147}{2{,}336} & \OmniMeanStd{137}{5.7} \\
\rowcolor{omniStripe}
Code2Video & \OmniMeanStd{0.841}{.012} & \OmniMeanStd{0.923}{.011} & \OmniMeanStd{0.724}{.014} & \OmniMeanStd{0.712}{.019} & \OmniMeanStd{0.541}{.023} & \OmniMeanStd{0.598}{.024} & \OmniMeanStd{0.742}{.020} & \OmniMeanStd{0.784}{.007} & \OmniMeanStd{0.748}{.009} & \OmniMeanStd{69{,}532}{769} & \OmniMeanStd{148}{3.7} \\
TheoremExplainAgent & \OmniMeanStd{0.863}{.011} & \OmniMeanStd{0.951}{.027} & \OmniMeanStd{0.807}{.017} & \textbf{\OmniMeanStd{0.786}{.007}} & \OmniMeanStd{0.731}{.002} & \OmniMeanStd{0.694}{.001} & \OmniMeanStd{0.861}{.009} & \textbf{\OmniMeanStd{0.824}{.018}} & \textbf{\OmniMeanStd{0.813}{.019}} & \OmniMeanStd{92{,}386}{5{,}416} & \OmniMeanStd{206}{6.7} \\
\rowcolor{omniOurs}
\textbf{OmniManim} & \textbf{\OmniMeanStd{0.907}{.008}} & \textbf{\OmniMeanStd{0.986}{.019}} & \textbf{\OmniMeanStd{0.836}{.016}} & \OmniMeanStd{0.730}{.011} & \textbf{\OmniMeanStd{0.868}{.008}} & \textbf{\OmniMeanStd{0.763}{.010}} & \textbf{\OmniMeanStd{0.931}{.012}} & \OmniMeanStd{0.802}{.015} & \OmniMeanStd{0.762}{.030} & \OmniMeanStd{73{,}279}{881} & \textbf{\OmniMeanStd{72}{2.4}} \\
\midrule
\OmniGroup{12}{Backbone: Gemini 3.1 Pro Preview\OmniModelLogo{gemini} (OmniManim: agent-only)}
\rowcolor{omniStripe}
Direct LLM & \OmniMeanStd{0.796}{.021} & \OmniMeanStd{0.918}{.009} & \OmniMeanStd{0.584}{.014} & \OmniMeanStd{0.598}{.007} & \OmniMeanStd{0.551}{.007} & \OmniMeanStd{0.578}{.027} & \OmniMeanStd{0.768}{.009} & \OmniMeanStd{0.742}{.011} & \OmniMeanStd{0.782}{.013} & \textbf{\OmniMeanStd{64{,}215}{2{,}418}} & \OmniMeanStd{133}{4.4} \\
Self-Refine & \OmniMeanStd{0.817}{.008} & \OmniMeanStd{0.943}{.020} & \OmniMeanStd{0.669}{.017} & \OmniMeanStd{0.653}{.018} & \OmniMeanStd{0.616}{.013} & \OmniMeanStd{0.637}{.020} & \OmniMeanStd{0.803}{.022} & \OmniMeanStd{0.779}{.007} & \OmniMeanStd{0.809}{.016} & \OmniMeanStd{72{,}683}{1{,}507} & \OmniMeanStd{164}{1.3} \\
\rowcolor{omniStripe}
Code2Video & \OmniMeanStd{0.812}{.019} & \OmniMeanStd{0.901}{.017} & \OmniMeanStd{0.748}{.012} & \OmniMeanStd{0.734}{.015} & \OmniMeanStd{0.558}{.005} & \OmniMeanStd{0.584}{.006} & \OmniMeanStd{0.758}{.018} & \OmniMeanStd{0.764}{.017} & \OmniMeanStd{0.782}{.012} & \OmniMeanStd{68{,}200}{2{,}056} & \OmniMeanStd{142}{3.1} \\
TheoremExplainAgent & \OmniMeanStd{0.846}{.035} & \OmniMeanStd{0.936}{.007} & \OmniMeanStd{0.796}{.033} & \textbf{\OmniMeanStd{0.773}{.019}} & \OmniMeanStd{0.708}{.010} & \OmniMeanStd{0.681}{.011} & \OmniMeanStd{0.832}{.012} & \OmniMeanStd{0.798}{.015} & \textbf{\OmniMeanStd{0.826}{.023}} & \OmniMeanStd{95{,}714}{1{,}437} & \OmniMeanStd{227}{8.5} \\
\rowcolor{omniOurs}
\textbf{OmniManim} & \textbf{\OmniMeanStd{0.884}{.009}} & \textbf{\OmniMeanStd{0.972}{.011}} & \textbf{\OmniMeanStd{0.852}{.023}} & \OmniMeanStd{0.748}{.021} & \textbf{\OmniMeanStd{0.852}{.006}} & \textbf{\OmniMeanStd{0.778}{.006}} & \textbf{\OmniMeanStd{0.942}{.013}} & \textbf{\OmniMeanStd{0.818}{.014}} & \OmniMeanStd{0.781}{.017} & \OmniMeanStd{74{,}800}{2{,}279} & \textbf{\OmniMeanStd{78}{2.2}} \\
\midrule
\OmniGroup{12}{Backbone: Qwen3-14B\OmniModelLogo{qwen} (OmniManim: SGCS + FLRPO)}
\rowcolor{omniStripe}
Direct LLM & \OmniMeanStd{0.174}{.009} & \OmniMeanStd{0.356}{.016} & \OmniMeanStd{0.296}{.011} & \OmniMeanStd{0.184}{.011} & \OmniMeanStd{0.151}{.012} & \OmniMeanStd{0.108}{.020} & \OmniMeanStd{0.181}{.023} & \OmniMeanStd{0.172}{.013} & \OmniMeanStd{0.204}{.017} & \textbf{\OmniMeanStd{28{,}714}{1{,}226}} & \textbf{\OmniMeanStd{56}{2.8}} \\
Self-Refine & \OmniMeanStd{0.219}{.010} & \OmniMeanStd{0.427}{.008} & \OmniMeanStd{0.358}{.014} & \OmniMeanStd{0.267}{.031} & \OmniMeanStd{0.223}{.010} & \OmniMeanStd{0.193}{.022} & \OmniMeanStd{0.281}{.018} & \OmniMeanStd{0.247}{.023} & \OmniMeanStd{0.309}{.014} & \OmniMeanStd{36{,}789}{875} & \OmniMeanStd{89}{2.7} \\
\rowcolor{omniStripe}
Code2Video & \OmniMeanStd{0.496}{.029} & \OmniMeanStd{0.698}{.012} & \OmniMeanStd{0.544}{.009} & \OmniMeanStd{0.472}{.010} & \OmniMeanStd{0.387}{.025} & \OmniMeanStd{0.436}{.020} & \OmniMeanStd{0.612}{.022} & \OmniMeanStd{0.519}{.011} & \OmniMeanStd{0.547}{.022} & \OmniMeanStd{54{,}826}{2{,}944} & \OmniMeanStd{157}{2.2} \\
TheoremExplainAgent & \OmniMeanStd{0.551}{.016} & \OmniMeanStd{0.742}{.015} & \OmniMeanStd{0.632}{.016} & \OmniMeanStd{0.658}{.025} & \OmniMeanStd{0.526}{.011} & \OmniMeanStd{0.487}{.006} & \OmniMeanStd{0.671}{.018} & \OmniMeanStd{0.563}{.016} & \OmniMeanStd{0.586}{.021} & \OmniMeanStd{67{,}193}{1{,}937} & \OmniMeanStd{218}{3.2} \\
\rowcolor{omniOurs}
\textbf{OmniManim} & \textbf{\OmniMeanStd{0.846}{.017}} & \textbf{\OmniMeanStd{0.947}{.009}} & \textbf{\OmniMeanStd{0.776}{.023}} & \textbf{\OmniMeanStd{0.731}{.004}} & \textbf{\OmniMeanStd{0.704}{.013}} & \textbf{\OmniMeanStd{0.804}{.017}} & \textbf{\OmniMeanStd{0.894}{.024}} & \textbf{\OmniMeanStd{0.793}{.008}} & \textbf{\OmniMeanStd{0.815}{.017}} & \OmniMeanStd{43{,}817}{1{,}740} & \OmniMeanStd{86}{2.4} \\
\midrule
\OmniGroup{12}{Backbone: Gemma-3-12B-IT\OmniModelLogo{gemma} (OmniManim: SGCS + FLRPO)}
\rowcolor{omniStripe}
Direct LLM & \OmniMeanStd{0.238}{.005} & \OmniMeanStd{0.419}{.008} & \OmniMeanStd{0.317}{.010} & \OmniMeanStd{0.246}{.014} & \OmniMeanStd{0.189}{.028} & \OmniMeanStd{0.157}{.019} & \OmniMeanStd{0.218}{.017} & \OmniMeanStd{0.263}{.021} & \OmniMeanStd{0.284}{.030} & \textbf{\OmniMeanStd{26{,}483}{1{,}041}} & \textbf{\OmniMeanStd{63}{1.9}} \\
Self-Refine & \OmniMeanStd{0.287}{.008} & \OmniMeanStd{0.476}{.011} & \OmniMeanStd{0.376}{.014} & \OmniMeanStd{0.329}{.013} & \OmniMeanStd{0.271}{.025} & \OmniMeanStd{0.236}{.020} & \OmniMeanStd{0.391}{.012} & \OmniMeanStd{0.337}{.007} & \OmniMeanStd{0.391}{.024} & \OmniMeanStd{35{,}162}{453} & \OmniMeanStd{98}{3.0} \\
\rowcolor{omniStripe}
Code2Video & \OmniMeanStd{0.463}{.003} & \OmniMeanStd{0.674}{.025} & \OmniMeanStd{0.528}{.018} & \OmniMeanStd{0.497}{.011} & \OmniMeanStd{0.416}{.012} & \OmniMeanStd{0.402}{.020} & \OmniMeanStd{0.587}{.027} & \OmniMeanStd{0.548}{.006} & \OmniMeanStd{0.573}{.002} & \OmniMeanStd{51{,}297}{2{,}481} & \OmniMeanStd{174}{10.1} \\
TheoremExplainAgent & \OmniMeanStd{0.526}{.012} & \OmniMeanStd{0.718}{.020} & \OmniMeanStd{0.611}{.017} & \OmniMeanStd{0.682}{.018} & \OmniMeanStd{0.569}{.014} & \OmniMeanStd{0.453}{.019} & \OmniMeanStd{0.649}{.015} & \OmniMeanStd{0.592}{.008} & \OmniMeanStd{0.619}{.021} & \OmniMeanStd{63{,}842}{1{,}941} & \OmniMeanStd{231}{12.2} \\
\rowcolor{omniOurs}
\textbf{OmniManim} & \textbf{\OmniMeanStd{0.818}{.015}} & \textbf{\OmniMeanStd{0.926}{.018}} & \textbf{\OmniMeanStd{0.752}{.033}} & \textbf{\OmniMeanStd{0.759}{.019}} & \textbf{\OmniMeanStd{0.721}{.017}} & \textbf{\OmniMeanStd{0.779}{.018}} & \textbf{\OmniMeanStd{0.868}{.024}} & \textbf{\OmniMeanStd{0.812}{.016}} & \textbf{\OmniMeanStd{0.806}{.019}} & \OmniMeanStd{41{,}753}{1{,}015} & \OmniMeanStd{94}{3.4} \\
\midrule
\OmniGroup{12}{Backbone: Llama-3.1-8B-Instruct\OmniModelLogo{meta} (OmniManim: SGCS + FLRPO)}
\rowcolor{omniStripe}
Direct LLM & \OmniMeanStd{0.197}{.013} & \OmniMeanStd{0.374}{.011} & \OmniMeanStd{0.278}{.009} & \OmniMeanStd{0.213}{.020} & \OmniMeanStd{0.168}{.027} & \OmniMeanStd{0.126}{.009} & \OmniMeanStd{0.193}{.019} & \OmniMeanStd{0.231}{.015} & \OmniMeanStd{0.259}{.024} & \textbf{\OmniMeanStd{24{,}761}{1{,}010}} & \textbf{\OmniMeanStd{49}{0.3}} \\
Self-Refine & \OmniMeanStd{0.251}{.020} & \OmniMeanStd{0.438}{.031} & \OmniMeanStd{0.341}{.012} & \OmniMeanStd{0.286}{.020} & \OmniMeanStd{0.239}{.013} & \OmniMeanStd{0.208}{.015} & \OmniMeanStd{0.307}{.019} & \OmniMeanStd{0.298}{.020} & \OmniMeanStd{0.346}{.014} & \OmniMeanStd{32{,}897}{817} & \OmniMeanStd{82}{3.5} \\
\rowcolor{omniStripe}
Code2Video & \OmniMeanStd{0.419}{.009} & \OmniMeanStd{0.628}{.014} & \OmniMeanStd{0.486}{.015} & \OmniMeanStd{0.451}{.006} & \OmniMeanStd{0.367}{.013} & \OmniMeanStd{0.378}{.005} & \OmniMeanStd{0.543}{.014} & \OmniMeanStd{0.493}{.018} & \OmniMeanStd{0.521}{.026} & \OmniMeanStd{48{,}637}{1{,}082} & \OmniMeanStd{139}{4.2} \\
TheoremExplainAgent & \OmniMeanStd{0.478}{.019} & \OmniMeanStd{0.681}{.026} & \OmniMeanStd{0.574}{.018} & \OmniMeanStd{0.613}{.017} & \OmniMeanStd{0.492}{.007} & \OmniMeanStd{0.429}{.019} & \OmniMeanStd{0.608}{.023} & \OmniMeanStd{0.541}{.016} & \OmniMeanStd{0.562}{.020} & \OmniMeanStd{59{,}128}{1{,}754} & \OmniMeanStd{193}{4.9} \\
\rowcolor{omniOurs}
\textbf{OmniManim} & \textbf{\OmniMeanStd{0.767}{.015}} & \textbf{\OmniMeanStd{0.902}{.019}} & \textbf{\OmniMeanStd{0.728}{.014}} & \textbf{\OmniMeanStd{0.694}{.019}} & \textbf{\OmniMeanStd{0.662}{.015}} & \textbf{\OmniMeanStd{0.751}{.011}} & \textbf{\OmniMeanStd{0.839}{.014}} & \textbf{\OmniMeanStd{0.756}{.013}} & \textbf{\OmniMeanStd{0.778}{.009}} & \OmniMeanStd{39{,}821}{712} & \OmniMeanStd{77}{3.1} \\
\bottomrule
\end{tabularx}
\end{table*}

\paragraph{Configuration scope.} GPT-5.4 and Gemini use the agent-only pipeline: scene and visual planning, code generation, and render-feedback-aware repair, without SGCS/FLRPO updates. Qwen3-14B, Gemma, and Llama use SGCS/FLRPO-trained policies in the main and transfer tables. Qwen ablations remove individual modules. Human evaluation assesses the agent-only GPT-5.4 pipeline.
\subsection{Experimental Setup}\label{sec:external}
\paragraph{Benchmarks and baselines.}
EduRequire-500 is the primary benchmark; TheoremExplainBench (240 tasks)~\citep{ku2025theoremexplainagent} and VisualEDU (200 tasks)~\citep{chen2025visualedu} assess transfer. We compare Direct LLM, Self-Refine~\citep{madaan2023selfrefine}, Code2Video~\citep{chen2025code2video}, TheoremExplainAgent~\citep{ku2025theoremexplainagent}, and OmniManim under the five model configurations in Table~\ref{tab:main_results}; transfer experiments use the trained Qwen, Gemma, and Llama code policies.
\paragraph{Protocol and metrics.}
All methods share the same inputs, diagnostic services, and one repair attempt with one candidate. Results are reported as mean$_{\pm\mathrm{SD}}$ over three seeds. Supplementary Table~S1 and Section~S5 provide metric definitions, evaluation protocols, and repair-auditing procedures.
\begin{table*}[t]
\centering
\captionsetup{width=\textwidth,skip=3pt}
\caption{TheoremExplainBench results. OmniManim uses SGCS/FLRPO-trained Qwen, Gemma, and Llama. All eleven metrics follow Supplementary Table~S1. Time is in seconds; entries report mean$_{\pm\mathrm{SD}}$ over three seeds.}
\label{tab:external_results}
\fontsize{7.5pt}{8.5pt}\selectfont
\OmniTableSetup
\setlength{\tabcolsep}{1pt}
\renewcommand{\arraystretch}{1.03}
\setlength{\aboverulesep}{1.5pt}
\setlength{\belowrulesep}{1.2pt}
\begin{tabularx}{\textwidth}{@{}>{\raggedright\arraybackslash}p{0.17\textwidth}*{9}{>{\centering\arraybackslash}X}rr@{}}
\toprule
\rowcolor{omniHeader}
\textbf{Method} & \multicolumn{2}{c}{\textbf{Executability} $\uparrow$}
& \multicolumn{3}{c}{\textbf{Instructional quality} $\uparrow$}
& \multicolumn{4}{c}{\textbf{Visual quality} $\uparrow$}
& \multicolumn{2}{c}{\textbf{Efficiency} $\downarrow$}\\
\rowcolor{omniHeader}
& R@1 & R@F & CA & PC & EN & OV & LQ & AC & VC & Tok & Time (s)\\
\midrule
\rowcolor{omniHeader}
\multicolumn{12}{@{}l}{\textbf{A. TheoremExplainBench}}\\
\midrule
\OmniGroup{12}{Backbone: Qwen3-14B\OmniModelLogo{qwen} (OmniManim: SGCS + FLRPO)}
\rowcolor{omniStripe}
Direct LLM & \OmniMeanStd{0.139}{.017} & \OmniMeanStd{0.318}{.013} & \OmniMeanStd{0.256}{.019} & \OmniMeanStd{0.184}{.020} & \OmniMeanStd{0.128}{.016} & \OmniMeanStd{0.062}{.012} & \OmniMeanStd{0.125}{.019} & \OmniMeanStd{0.115}{.006} & \OmniMeanStd{0.178}{.016} & \textbf{\OmniMeanStd{39{,}720}{1{,}311}} & \textbf{\OmniMeanStd{86}{3.7}} \\
Self-Refine & \OmniMeanStd{0.172}{.025} & \OmniMeanStd{0.379}{.011} & \OmniMeanStd{0.319}{.024} & \OmniMeanStd{0.259}{.018} & \OmniMeanStd{0.197}{.014} & \OmniMeanStd{0.147}{.008} & \OmniMeanStd{0.319}{.016} & \OmniMeanStd{0.195}{.016} & \OmniMeanStd{0.290}{.008} & \OmniMeanStd{57{,}594}{801} & \OmniMeanStd{135}{4.8} \\
\rowcolor{omniStripe}
Code2Video & \OmniMeanStd{0.467}{.011} & \OmniMeanStd{0.644}{.021} & \OmniMeanStd{0.508}{.019} & \OmniMeanStd{0.471}{.004} & \OmniMeanStd{0.347}{.020} & \OmniMeanStd{0.382}{.021} & \OmniMeanStd{0.568}{.018} & \OmniMeanStd{0.472}{.030} & \OmniMeanStd{0.521}{.018} & \OmniMeanStd{80{,}266}{3{,}343} & \OmniMeanStd{240}{10.4} \\
TheoremExplainAgent & \OmniMeanStd{0.519}{.010} & \OmniMeanStd{0.708}{.015} & \OmniMeanStd{0.619}{.006} & \OmniMeanStd{0.628}{.024} & \OmniMeanStd{0.489}{.008} & \OmniMeanStd{0.445}{.005} & \OmniMeanStd{0.646}{.021} & \OmniMeanStd{0.517}{.015} & \OmniMeanStd{0.566}{.018} & \OmniMeanStd{89{,}242}{3{,}499} & \OmniMeanStd{288}{6.2} \\
\rowcolor{omniOurs}
\textbf{OmniManim} & \textbf{\OmniMeanStd{0.785}{.013}} & \textbf{\OmniMeanStd{0.919}{.013}} & \textbf{\OmniMeanStd{0.750}{.011}} & \textbf{\OmniMeanStd{0.722}{.026}} & \textbf{\OmniMeanStd{0.676}{.022}} & \textbf{\OmniMeanStd{0.767}{.012}} & \textbf{\OmniMeanStd{0.871}{.013}} & \textbf{\OmniMeanStd{0.751}{.012}} & \textbf{\OmniMeanStd{0.772}{.023}} & \OmniMeanStd{58{,}873}{1{,}992} & \OmniMeanStd{130}{3.7} \\
\midrule
\OmniGroup{12}{Backbone: Gemma-3-12B-IT\OmniModelLogo{gemma} (OmniManim: SGCS + FLRPO)}
\rowcolor{omniStripe}
Direct LLM & \OmniMeanStd{0.176}{.021} & \OmniMeanStd{0.406}{.010} & \OmniMeanStd{0.271}{.018} & \OmniMeanStd{0.224}{.014} & \OmniMeanStd{0.171}{.017} & \OmniMeanStd{0.111}{.008} & \OmniMeanStd{0.190}{.027} & \OmniMeanStd{0.200}{.009} & \OmniMeanStd{0.267}{.027} & \textbf{\OmniMeanStd{38{,}483}{304}} & \textbf{\OmniMeanStd{82}{2.2}} \\
Self-Refine & \OmniMeanStd{0.225}{.033} & \OmniMeanStd{0.451}{.013} & \OmniMeanStd{0.353}{.009} & \OmniMeanStd{0.292}{.032} & \OmniMeanStd{0.222}{.003} & \OmniMeanStd{0.179}{.011} & \OmniMeanStd{0.331}{.018} & \OmniMeanStd{0.299}{.015} & \OmniMeanStd{0.381}{.013} & \OmniMeanStd{53{,}794}{2{,}429} & \OmniMeanStd{129}{4.7} \\
\rowcolor{omniStripe}
Code2Video & \OmniMeanStd{0.438}{.026} & \OmniMeanStd{0.621}{.022} & \OmniMeanStd{0.521}{.015} & \OmniMeanStd{0.466}{.010} & \OmniMeanStd{0.371}{.016} & \OmniMeanStd{0.346}{.014} & \OmniMeanStd{0.579}{.024} & \OmniMeanStd{0.519}{.011} & \OmniMeanStd{0.564}{.022} & \OmniMeanStd{69{,}553}{3{,}465} & \OmniMeanStd{239}{8.3} \\
TheoremExplainAgent & \OmniMeanStd{0.476}{.013} & \OmniMeanStd{0.667}{.026} & \OmniMeanStd{0.600}{.011} & \OmniMeanStd{0.683}{.018} & \OmniMeanStd{0.545}{.008} & \OmniMeanStd{0.403}{.009} & \OmniMeanStd{0.622}{.019} & \OmniMeanStd{0.554}{.008} & \OmniMeanStd{0.610}{.025} & \OmniMeanStd{94{,}044}{5{,}261} & \OmniMeanStd{344}{16.0} \\
\rowcolor{omniOurs}
\textbf{OmniManim} & \textbf{\OmniMeanStd{0.763}{.015}} & \textbf{\OmniMeanStd{0.892}{.027}} & \textbf{\OmniMeanStd{0.733}{.022}} & \textbf{\OmniMeanStd{0.736}{.011}} & \textbf{\OmniMeanStd{0.691}{.011}} & \textbf{\OmniMeanStd{0.739}{.029}} & \textbf{\OmniMeanStd{0.861}{.027}} & \textbf{\OmniMeanStd{0.779}{.012}} & \textbf{\OmniMeanStd{0.776}{.010}} & \OmniMeanStd{60{,}394}{2{,}414} & \OmniMeanStd{141}{6.0} \\
\midrule
\OmniGroup{12}{Backbone: Llama-3.1-8B-Instruct\OmniModelLogo{meta} (OmniManim: SGCS + FLRPO)}
\rowcolor{omniStripe}
Direct LLM & \OmniMeanStd{0.165}{.010} & \OmniMeanStd{0.343}{.015} & \OmniMeanStd{0.268}{.010} & \OmniMeanStd{0.186}{.015} & \OmniMeanStd{0.126}{.013} & \OmniMeanStd{0.106}{.027} & \OmniMeanStd{0.164}{.014} & \OmniMeanStd{0.181}{.028} & \OmniMeanStd{0.247}{.014} & \textbf{\OmniMeanStd{31{,}948}{1{,}928}} & \textbf{\OmniMeanStd{67}{2.1}} \\
Self-Refine & \OmniMeanStd{0.229}{.015} & \OmniMeanStd{0.408}{.018} & \OmniMeanStd{0.326}{.015} & \OmniMeanStd{0.270}{.014} & \OmniMeanStd{0.219}{.012} & \OmniMeanStd{0.170}{.015} & \OmniMeanStd{0.272}{.007} & \OmniMeanStd{0.234}{.011} & \OmniMeanStd{0.335}{.016} & \OmniMeanStd{50{,}982}{1{,}070} & \OmniMeanStd{115}{3.9} \\
\rowcolor{omniStripe}
Code2Video & \OmniMeanStd{0.362}{.019} & \OmniMeanStd{0.582}{.013} & \OmniMeanStd{0.444}{.010} & \OmniMeanStd{0.441}{.022} & \OmniMeanStd{0.318}{.008} & \OmniMeanStd{0.328}{.016} & \OmniMeanStd{0.516}{.017} & \OmniMeanStd{0.431}{.032} & \OmniMeanStd{0.501}{.023} & \OmniMeanStd{66{,}875}{2{,}276} & \OmniMeanStd{216}{11.8} \\
TheoremExplainAgent & \OmniMeanStd{0.440}{.006} & \OmniMeanStd{0.668}{.017} & \OmniMeanStd{0.567}{.015} & \OmniMeanStd{0.588}{.013} & \OmniMeanStd{0.472}{.017} & \OmniMeanStd{0.395}{.006} & \OmniMeanStd{0.536}{.012} & \OmniMeanStd{0.489}{.009} & \OmniMeanStd{0.543}{.012} & \OmniMeanStd{89{,}551}{4{,}300} & \OmniMeanStd{302}{2.0} \\
\rowcolor{omniOurs}
\textbf{OmniManim} & \textbf{\OmniMeanStd{0.713}{.018}} & \textbf{\OmniMeanStd{0.853}{.025}} & \textbf{\OmniMeanStd{0.683}{.011}} & \textbf{\OmniMeanStd{0.681}{.009}} & \textbf{\OmniMeanStd{0.613}{.013}} & \textbf{\OmniMeanStd{0.719}{.011}} & \textbf{\OmniMeanStd{0.814}{.011}} & \textbf{\OmniMeanStd{0.696}{.015}} & \textbf{\OmniMeanStd{0.754}{.022}} & \OmniMeanStd{60{,}173}{2{,}722} & \OmniMeanStd{108}{0.8} \\
\bottomrule
\end{tabularx}
\end{table*}

\begin{table*}[t]
\centering
\captionsetup{width=\textwidth,skip=3pt}
\caption{VisualEDU results. OmniManim uses SGCS/FLRPO-trained Qwen, Gemma, and Llama. All eleven metrics follow Supplementary Table~S1. Time is in seconds; entries report mean$_{\pm\mathrm{SD}}$ over three seeds.}
\label{tab:visualedu_results}
\fontsize{7.5pt}{8.5pt}\selectfont
\OmniTableSetup
\setlength{\tabcolsep}{1pt}
\renewcommand{\arraystretch}{1.03}
\setlength{\aboverulesep}{1.5pt}
\setlength{\belowrulesep}{1.2pt}
\begin{tabularx}{\textwidth}{@{}>{\raggedright\arraybackslash}p{0.17\textwidth}*{9}{>{\centering\arraybackslash}X}rr@{}}
\toprule
\rowcolor{omniHeader}
\textbf{Method} & \multicolumn{2}{c}{\textbf{Executability} $\uparrow$}
& \multicolumn{3}{c}{\textbf{Instructional quality} $\uparrow$}
& \multicolumn{4}{c}{\textbf{Visual quality} $\uparrow$}
& \multicolumn{2}{c}{\textbf{Efficiency} $\downarrow$}\\
\rowcolor{omniHeader}
& R@1 & R@F & CA & PC & EN & OV & LQ & AC & VC & Tok & Time (s)\\
\midrule
\rowcolor{omniHeader}
\rowcolor{omniHeader}
\multicolumn{12}{@{}l}{\textbf{B. VisualEDU}}\\

\midrule
\OmniGroup{12}{Backbone: Qwen3-14B\OmniModelLogo{qwen} (OmniManim: SGCS + FLRPO)}
\rowcolor{omniStripe}
Direct LLM & \OmniMeanStd{0.153}{.019} & \OmniMeanStd{0.357}{.028} & \OmniMeanStd{0.293}{.020} & \OmniMeanStd{0.216}{.027} & \OmniMeanStd{0.150}{.027} & \OmniMeanStd{0.081}{.012} & \OmniMeanStd{0.153}{.019} & \OmniMeanStd{0.161}{.009} & \OmniMeanStd{0.232}{.023} & \textbf{\OmniMeanStd{35{,}869}{1{,}327}} & \textbf{\OmniMeanStd{68}{2.0}} \\
Self-Refine & \OmniMeanStd{0.192}{.015} & \OmniMeanStd{0.423}{.008} & \OmniMeanStd{0.357}{.016} & \OmniMeanStd{0.273}{.006} & \OmniMeanStd{0.211}{.013} & \OmniMeanStd{0.161}{.032} & \OmniMeanStd{0.250}{.031} & \OmniMeanStd{0.210}{.017} & \OmniMeanStd{0.324}{.004} & \OmniMeanStd{46{,}164}{1{,}596} & \OmniMeanStd{109}{2.7} \\
\rowcolor{omniStripe}
Code2Video & \OmniMeanStd{0.463}{.029} & \OmniMeanStd{0.683}{.008} & \OmniMeanStd{0.542}{.015} & \OmniMeanStd{0.507}{.015} & \OmniMeanStd{0.363}{.027} & \OmniMeanStd{0.429}{.022} & \OmniMeanStd{0.626}{.021} & \OmniMeanStd{0.487}{.018} & \OmniMeanStd{0.563}{.019} & \OmniMeanStd{64{,}348}{2{,}262} & \OmniMeanStd{176}{3.5} \\
TheoremExplainAgent & \OmniMeanStd{0.520}{.023} & \OmniMeanStd{0.717}{.016} & \OmniMeanStd{0.632}{.008} & \OmniMeanStd{0.676}{.018} & \OmniMeanStd{0.526}{.011} & \OmniMeanStd{0.486}{.012} & \OmniMeanStd{0.682}{.018} & \OmniMeanStd{0.561}{.014} & \OmniMeanStd{0.596}{.008} & \OmniMeanStd{90{,}113}{4{,}029} & \OmniMeanStd{250}{2.6} \\
\rowcolor{omniOurs}
\textbf{OmniManim} & \textbf{\OmniMeanStd{0.830}{.013}} & \textbf{\OmniMeanStd{0.937}{.010}} & \textbf{\OmniMeanStd{0.783}{.010}} & \textbf{\OmniMeanStd{0.730}{.018}} & \textbf{\OmniMeanStd{0.711}{.019}} & \textbf{\OmniMeanStd{0.798}{.015}} & \textbf{\OmniMeanStd{0.893}{.010}} & \textbf{\OmniMeanStd{0.783}{.019}} & \textbf{\OmniMeanStd{0.810}{.009}} & \OmniMeanStd{57{,}533}{3{,}089} & \OmniMeanStd{116}{3.1} \\
\midrule
\OmniGroup{12}{Backbone: Gemma-3-12B-IT\OmniModelLogo{gemma} (OmniManim: SGCS + FLRPO)}
\rowcolor{omniStripe}
Direct LLM & \OmniMeanStd{0.213}{.010} & \OmniMeanStd{0.402}{.013} & \OmniMeanStd{0.312}{.013} & \OmniMeanStd{0.275}{.011} & \OmniMeanStd{0.196}{.006} & \OmniMeanStd{0.144}{.021} & \OmniMeanStd{0.206}{.004} & \OmniMeanStd{0.236}{.024} & \OmniMeanStd{0.307}{.002} & \textbf{\OmniMeanStd{30{,}488}{626}} & \textbf{\OmniMeanStd{69}{1.4}} \\
Self-Refine & \OmniMeanStd{0.263}{.013} & \OmniMeanStd{0.457}{.025} & \OmniMeanStd{0.378}{.010} & \OmniMeanStd{0.349}{.007} & \OmniMeanStd{0.241}{.016} & \OmniMeanStd{0.201}{.020} & \OmniMeanStd{0.327}{.014} & \OmniMeanStd{0.331}{.002} & \OmniMeanStd{0.379}{.021} & \OmniMeanStd{41{,}284}{1{,}427} & \OmniMeanStd{113}{4.5} \\
\rowcolor{omniStripe}
Code2Video & \OmniMeanStd{0.435}{.018} & \OmniMeanStd{0.668}{.013} & \OmniMeanStd{0.553}{.019} & \OmniMeanStd{0.515}{.025} & \OmniMeanStd{0.389}{.007} & \OmniMeanStd{0.375}{.021} & \OmniMeanStd{0.603}{.016} & \OmniMeanStd{0.514}{.022} & \OmniMeanStd{0.575}{.023} & \OmniMeanStd{68{,}658}{1{,}469} & \OmniMeanStd{220}{10.9} \\
TheoremExplainAgent & \OmniMeanStd{0.497}{.023} & \OmniMeanStd{0.718}{.008} & \OmniMeanStd{0.612}{.015} & \OmniMeanStd{0.707}{.019} & \OmniMeanStd{0.544}{.027} & \OmniMeanStd{0.419}{.011} & \OmniMeanStd{0.597}{.025} & \OmniMeanStd{0.584}{.010} & \OmniMeanStd{0.607}{.002} & \OmniMeanStd{85{,}669}{3{,}481} & \OmniMeanStd{264}{14.7} \\
\rowcolor{omniOurs}
\textbf{OmniManim} & \textbf{\OmniMeanStd{0.788}{.013}} & \textbf{\OmniMeanStd{0.932}{.013}} & \textbf{\OmniMeanStd{0.770}{.018}} & \textbf{\OmniMeanStd{0.761}{.021}} & \textbf{\OmniMeanStd{0.717}{.004}} & \textbf{\OmniMeanStd{0.759}{.004}} & \textbf{\OmniMeanStd{0.850}{.012}} & \textbf{\OmniMeanStd{0.782}{.021}} & \textbf{\OmniMeanStd{0.834}{.015}} & \OmniMeanStd{50{,}696}{1{,}526} & \OmniMeanStd{110}{6.2} \\
\midrule
\OmniGroup{12}{Backbone: Llama-3.1-8B-Instruct\OmniModelLogo{meta} (OmniManim: SGCS + FLRPO)}
\rowcolor{omniStripe}
Direct LLM & \OmniMeanStd{0.198}{.008} & \OmniMeanStd{0.342}{.019} & \OmniMeanStd{0.290}{.023} & \OmniMeanStd{0.242}{.023} & \OmniMeanStd{0.152}{.026} & \OmniMeanStd{0.092}{.013} & \OmniMeanStd{0.161}{.009} & \OmniMeanStd{0.190}{.009} & \OmniMeanStd{0.269}{.014} & \textbf{\OmniMeanStd{27{,}265}{330}} & \textbf{\OmniMeanStd{57}{2.1}} \\
Self-Refine & \OmniMeanStd{0.217}{.015} & \OmniMeanStd{0.410}{.013} & \OmniMeanStd{0.357}{.028} & \OmniMeanStd{0.282}{.011} & \OmniMeanStd{0.238}{.013} & \OmniMeanStd{0.174}{.011} & \OmniMeanStd{0.337}{.021} & \OmniMeanStd{0.277}{.017} & \OmniMeanStd{0.346}{.007} & \OmniMeanStd{44{,}103}{2{,}485} & \OmniMeanStd{99}{3.8} \\
\rowcolor{omniStripe}
Code2Video & \OmniMeanStd{0.413}{.008} & \OmniMeanStd{0.633}{.013} & \OmniMeanStd{0.502}{.020} & \OmniMeanStd{0.485}{.011} & \OmniMeanStd{0.369}{.008} & \OmniMeanStd{0.342}{.019} & \OmniMeanStd{0.561}{.025} & \OmniMeanStd{0.460}{.025} & \OmniMeanStd{0.523}{.032} & \OmniMeanStd{63{,}397}{1{,}981} & \OmniMeanStd{156}{0.4} \\
TheoremExplainAgent & \OmniMeanStd{0.472}{.020} & \OmniMeanStd{0.657}{.020} & \OmniMeanStd{0.597}{.021} & \OmniMeanStd{0.633}{.006} & \OmniMeanStd{0.491}{.003} & \OmniMeanStd{0.413}{.015} & \OmniMeanStd{0.607}{.014} & \OmniMeanStd{0.523}{.016} & \OmniMeanStd{0.573}{.016} & \OmniMeanStd{69{,}699}{427} & \OmniMeanStd{259}{2.0} \\
\rowcolor{omniOurs}
\textbf{OmniManim} & \textbf{\OmniMeanStd{0.753}{.016}} & \textbf{\OmniMeanStd{0.903}{.023}} & \textbf{\OmniMeanStd{0.730}{.013}} & \textbf{\OmniMeanStd{0.702}{.010}} & \textbf{\OmniMeanStd{0.649}{.021}} & \textbf{\OmniMeanStd{0.753}{.001}} & \textbf{\OmniMeanStd{0.820}{.018}} & \textbf{\OmniMeanStd{0.730}{.003}} & \textbf{\OmniMeanStd{0.795}{.016}} & \OmniMeanStd{46{,}070}{1{,}332} & \OmniMeanStd{91}{2.2} \\
\bottomrule
\end{tabularx}
\end{table*}

\begin{figure*}[t]
\centering\includegraphics[width=\linewidth]{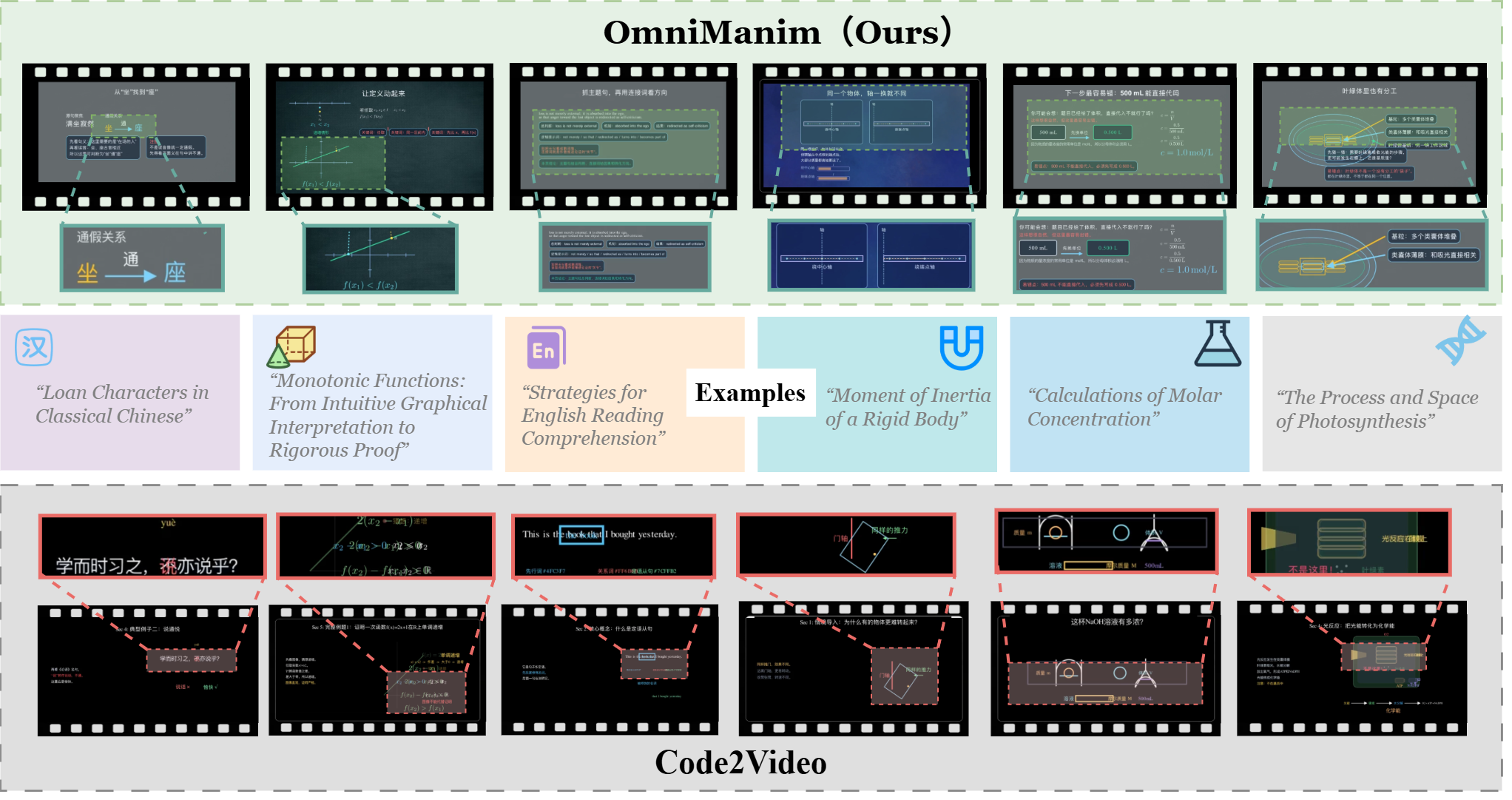}
\caption{Qualitative comparison of OmniManim with Code2Video. The examples illustrate differences in spatial organization, object placement, and visual clarity in the rendered educational animations.}\label{fig:contrast}
\end{figure*}
\subsection{Main Results}
Table~\ref{tab:main_results} shows improvements in executability and spatial quality. OmniManim with GPT-5.4 raises R@F from 0.923 to 0.986 over Code2Video and OV/LQ from 0.598/0.742 to 0.763/0.931. OmniManim with Qwen raises R@F from 0.698 to 0.947; OmniManim with Gemma and Llama improves it by 25.2 and 27.4 percentage points. OmniManim leads all nine execution and quality metrics within each open-model group. TheoremExplainAgent retains higher PC and VC in both API groups and higher AC with GPT-5.4. Figure~\ref{fig:contrast} illustrates spatial improvements.
\paragraph{Cross-dataset transfer.}
Across six dataset--backbone groups in Tables~\ref{tab:external_results} and~\ref{tab:visualedu_results}, OmniManim leads all nine execution and quality metrics. R@F reaches 0.853--0.919 on TheoremExplainBench and 0.903--0.937 on VisualEDU, versus Code2Video's 0.582--0.644 and 0.633--0.683. These results support transfer of the trained Qwen, Gemma, and Llama policies across both STEM explanation tasks and mathematical problem-solving video generation.
\paragraph{Efficiency.}
OmniManim takes 72/78\,s with GPT/Gemini versus Code2Video's 148/142\,s. OmniManim with Qwen uses 43{,}817 tokens and 86\,s versus 54{,}826 tokens and 157\,s: 20.1\% fewer tokens and 45.2\% less time. OmniManim with Gemma and Llama also improves efficiency over Code2Video. With GPT-5.4, OmniManim uses 73{,}279 tokens and 72\,s against Direct LLM's 56{,}328 tokens and 108\,s: scene programs are written and rendered concurrently, then concatenated, so wall-clock time undercuts one sequential generation and compilation. Direct LLM is cheaper in the open-model groups but substantially less reliable. Costs cover deployment under the shared repair budget; offline SGCS/FLRPO training costs are excluded.
\subsection{Module Ablation}\label{subsec:training_evaluation}
\begin{table*}[t]
\centering
\caption{Qwen3-14B module ablation on EduRequire-500. Checks mark enabled components; Base is the shared scaffold. Row 5 enables Base, Vision, SGCS, and FLRPO. Entries are mean$_{\pm\mathrm{SD}}$ over three seeds; higher is better.}
\label{tab:ablation}
\footnotesize\OmniTableSetup\setlength{\tabcolsep}{2.5pt}\setlength{\arrayrulewidth}{0.3pt}\renewcommand{\arraystretch}{1.08}
\begin{tabularx}{\textwidth}{@{}>{\centering\arraybackslash}p{0.035\textwidth}|*{4}{>{\centering\arraybackslash}p{0.06\textwidth}}|*{8}{>{\centering\arraybackslash}X}@{}}
\toprule
\rowcolor{omniHeader}
& \multicolumn{4}{c|}{\textbf{Components}}
& \multicolumn{5}{c}{\textbf{Generation and rendering} $\uparrow$}
& \multicolumn{3}{c}{\textbf{Repair} $\uparrow$}\\
\rowcolor{omniHeader}
\textbf{Idx} & \textbf{Base} & \textbf{Vision} & \textbf{SGCS} & \textbf{FLRPO} & R@1 & R@F & OV & LQ & AC & RS@1 & CPR & ELR\\
\specialrule{0.45pt}{0pt}{1pt}
\specialrule{0.3pt}{0pt}{2pt}
1 & \OmniCheck &  &  &  & \OmniMeanStd{0.427}{.019} & \OmniMeanStd{0.613}{.021} & \OmniMeanStd{0.315}{.024} & \OmniMeanStd{0.482}{.015} & \OmniMeanStd{0.456}{.031} & \OmniMeanStd{0.328}{.008} & \OmniMeanStd{0.824}{.012} & \OmniMeanStd{0.485}{.008} \\
\rowcolor{omniStripe}
2 & \OmniCheck &  & \OmniCheck & \OmniCheck & \OmniMeanStd{0.739}{.012} & \OmniMeanStd{0.871}{.016} & \OmniMeanStd{0.631}{.011} & \OmniMeanStd{0.724}{.007} & \OmniMeanStd{0.685}{.026} & \OmniMeanStd{0.708}{.015} & \OmniMeanStd{0.905}{.011} & \OmniMeanStd{0.828}{.009} \\
3 & \OmniCheck & \OmniCheck &  &  & \OmniMeanStd{0.714}{.033} & \OmniMeanStd{0.864}{.009} & \OmniMeanStd{0.695}{.013} & \OmniMeanStd{0.815}{.016} & \OmniMeanStd{0.717}{.010} & \OmniMeanStd{0.617}{.021} & \OmniMeanStd{0.937}{.008} & \OmniMeanStd{0.810}{.004} \\
\rowcolor{omniStripe}
4 & \OmniCheck & \OmniCheck & \OmniCheck &  & \OmniMeanStd{0.823}{.006} & \OmniMeanStd{0.916}{.029} & \OmniMeanStd{0.786}{.015} & \OmniMeanStd{0.873}{.008} & \OmniMeanStd{0.766}{.012} & \OmniMeanStd{0.665}{.013} & \OmniMeanStd{0.922}{.004} & \OmniMeanStd{0.857}{.002} \\
\rowcolor{omniOurs}
\textbf{5} & \OmniCheck & \OmniCheck & \OmniCheck & \OmniCheck & \textbf{\OmniMeanStd{0.846}{.017}} & \textbf{\OmniMeanStd{0.947}{.009}} & \textbf{\OmniMeanStd{0.804}{.017}} & \textbf{\OmniMeanStd{0.894}{.024}} & \textbf{\OmniMeanStd{0.793}{.008}} & \textbf{\OmniMeanStd{0.783}{.010}} & \textbf{\OmniMeanStd{0.957}{.003}} & \textbf{\OmniMeanStd{0.909}{.006}} \\
\bottomrule
\end{tabularx}
\end{table*}
In Table~\ref{tab:ablation}, we train Qwen3-14B variants under fixed data and deployment budgets. Row 5 enables Base, Vision, SGCS, and FLRPO. Row 2 removes Vision, and row 4 removes FLRPO. Row 3 is agent-only, with Base and Vision only. Removing Vision lowers OV/LQ from 0.804/0.894 to 0.631/0.724. Agent-only Qwen reaches R@F 0.864, against 0.947 with SGCS and FLRPO. Removing FLRPO lowers RS@1 from 0.783 to 0.665 and ELR from 0.909 to 0.857. The trained configuration leads every ablation metric. Supplementary Section~S5 compares SGCS with uniform SFT ($\omega=0$) and FLRPO with GRPO using global rewards alone.

\begin{figure*}[t]
\centering
\includegraphics[width=0.9\linewidth]{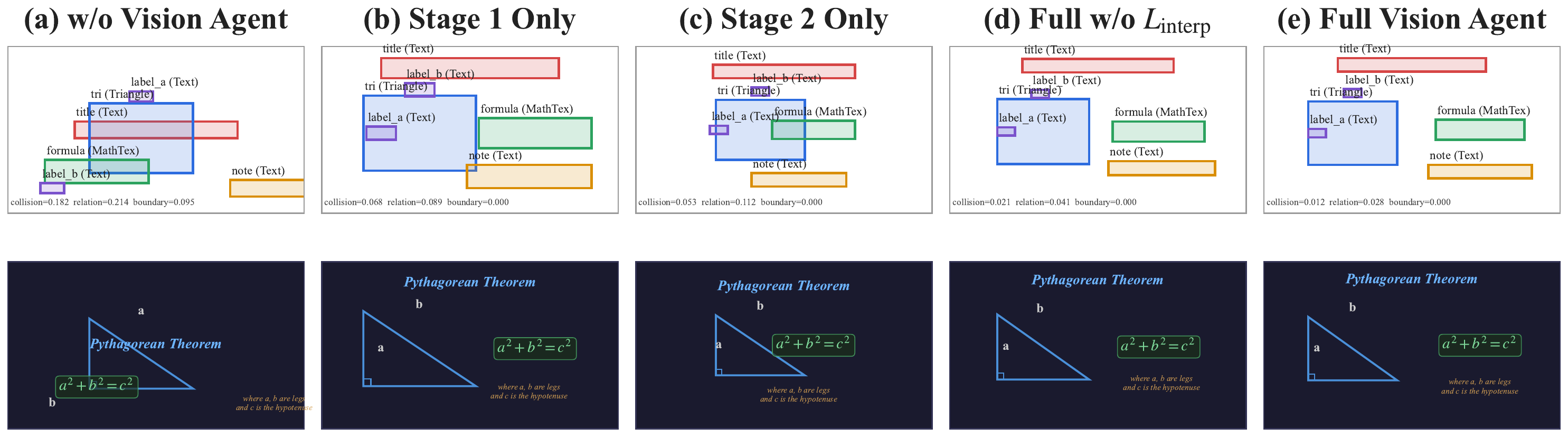}
\caption{Vision Agent ablation: layouts and diagnostics (top), rendered frames (bottom). Columns omit the planner, use only coarse priors or box denoising, omit interpolation loss, or use the full planner. Stage 1/2 refer to coarse-prior learning and box denoising.}
\label{fig:ablation_qualitative}
\end{figure*}

\paragraph{Vision Agent ablation.}
Figure~\ref{fig:ablation_qualitative} compares predicted boxes and rendered frames across five planner variants, isolating coarse-prior learning, box denoising, and interpolation constraints. These experiments isolate the visual planner design rather than SGCS/FLRPO policy training.

\paragraph{Repair evaluation.}\label{par:repair_protocol}
On a common held-out failure set, one-patch repair is measured by target correction (RS@1), preservation of unaffected constraints (CPR), and edit localization (ELR). The trained Qwen configuration in Table~\ref{tab:main_results} reaches 0.783/0.957/0.909, respectively. With SGCS, diagnostics, and patch actions fixed, global-reward GRPO reaches RS@1/ELR of 0.732/0.872 (Supplementary Table~S8), supporting FLRPO's local rendering reward and edit penalty. Definitions and auditing procedures appear in Supplementary Section~S5.
\Needspace{0.50\textheight}
\subsection{Human Evaluation}\label{sec:human_eval}
\begin{figure}[H]
\centering
\includegraphics[width=0.82\linewidth]{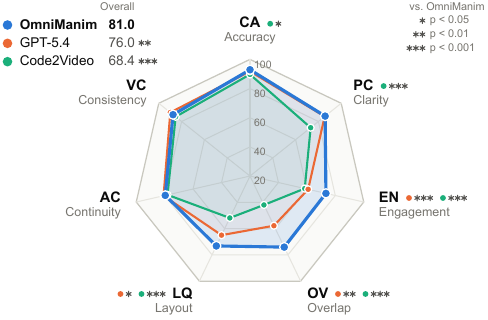}
\caption{Human ratings (0--100) on 60 EduRequire-500 tasks (three methods each). Asterisks: paired Wilcoxon tests (Supplementary Table~S9).}
\label{fig:human_eval}
\end{figure}
One hundred undergraduates rate 60 tasks (Figure~\ref{fig:human_eval}). Gains favor engagement and spatial presentation (overall $p=0.009$ vs.\ GPT-5.4, $p<10^{-5}$ vs.\ Code2Video); GPT-5.4 retains higher AC/VC. Ratings use agent-only GPT-5.4.

\section{Discussion and Conclusion}
\label{sec:discussion}

With GPT-5.4, final render success and layout quality reach 0.986 and 0.931 in 72\,s, versus 148\,s for Code2Video. Latency includes pipeline scheduling; equal-compute gains remain untested. Limitations include 2D Manim layouts, one repair with one candidate, closed vision-language review of flagged frames, and untested scaling to larger models. Code and datasets will be made publicly available for reproducible evaluation.

\section{LLM Usage Statement}
Generative models are used in the experiments, the code, the training-data construction, and limited language editing, as follows. GPT-5.4 and Gemini 3.1 Pro Preview are frozen agent-only generators. Qwen3-14B, Gemma-3-12B-IT, and Llama-3.1-8B-Instruct are trained with SGCS and FLRPO. These models write and locally patch Manim code. The Scene Agent parses teaching requirements with temperature-0 structured decoding. The Vision Agent is a trained box-denoising model, not a prompted language model. For the policy-training records in Supplementary Section~S4.3, GPT-5.4 drafts requirement and scene descriptions and candidate repair patches at temperature 0. The executable target remains the source-derived program adapted to the frozen planner's layouts. These drafts are accepted only after code execution and the prescribed visual checks, and no evaluation requirement is used. Claude Opus 4.6 is the vision-language judge for deployment review and for the offline content-accuracy, pedagogical-clarity, and engagement scores, using fixed prompts and deterministic decoding. Generative AI assists with English polishing of the expert-authored EduRequire-500 requirements and this manuscript, LaTeX formatting, and reference checking. The authors are responsible for verifying the scientific claims, references, and final submission.

\clearpage
\renewcommand{\OmniTableSetup}{%
  \setlength{\tabcolsep}{3pt}%
  \renewcommand{\arraystretch}{1.08}%
  \arrayrulecolor{black!70}%
}
\setlength{\textfloatsep}{10pt plus 2pt minus 2pt}
\setlength{\dbltextfloatsep}{10pt plus 2pt minus 2pt}
\setlength{\intextsep}{12pt plus 2pt minus 2pt}
\setcounter{dbltopnumber}{2}
\clubpenalty=150
\widowpenalty=150
\displaywidowpenalty=50
\finalhyphendemerits=5000
\setlength{\parfillskip}{0pt plus 1fil}
\maketitlesupplementary

\setcounter{section}{0}
\setcounter{table}{0}
\setcounter{figure}{0}
\renewcommand{\thesection}{S\arabic{section}}
\renewcommand{\thetable}{S\arabic{table}}
\renewcommand{\thefigure}{S\arabic{figure}}
\renewcommand{\theequation}{S\arabic{equation}}
\setcounter{equation}{0}
\renewcommand{\theHsection}{S\arabic{section}}
\renewcommand{\theHtable}{S\arabic{table}}
\renewcommand{\theHfigure}{S\arabic{figure}}
\renewcommand{\theHequation}{S\arabic{equation}}

\section{Metric Definitions}
\label{supp:metrics}

Table~\ref{tab:metrics} summarizes the eleven generation metrics and three repair-policy metrics. The generation metrics are evaluated on each full benchmark; repair metrics are conditional on a common frozen set of failed programs. Section~\ref{app:metric_validation} explains their rationale, available evidence, and validation limits. Detailed VLM and CV procedures are given in Sections~\ref{app:vlm} and~\ref{app:cv}.

\begin{table*}[!t]
\centering
\caption{Evaluation metrics for the main comparisons and module ablation. Rendering rates and quality scores are in $[0,1]$; tokens are counts and time is in seconds. Repair metrics use a common fixed set of failing programs. Arrows indicate the preferred direction.}
\label{tab:metrics}
\footnotesize
\OmniTableSetup
\setlength{\tabcolsep}{3pt}
\begin{tabularx}{\textwidth}{@{}>{\raggedright\arraybackslash}p{0.065\textwidth} >{\raggedright\arraybackslash}p{0.23\textwidth} >{\raggedright\arraybackslash}X@{}}
\toprule
\rowcolor{omniHeader}
\textbf{Abbrev.} & \textbf{Metric} & \textbf{What it measures}\\
\midrule
\OmniGroup{3}{Executability $\uparrow$}
\rowcolor{omniStripe}
R@1 & First-attempt render & Successful initial rendering.\\
R@F & Final render & Successful rendering after repair.\\
\midrule
\OmniGroup{3}{Instructional quality $\uparrow$}
\rowcolor{omniStripe}
CA & Content accuracy & Correctness against the teaching requirement.\\
PC & Pedagogical clarity & Clarity and instructional organization.\\
\rowcolor{omniStripe}
EN & Engagement & Viewer interest and presentation quality.\\
\midrule
\OmniGroup{3}{Visual quality $\uparrow$}
\rowcolor{omniStripe}
OV & Overlap avoidance & Spatial non-overlap; higher is cleaner.\\
LQ & Layout quality & Spatial organization and local crowding.\\
\rowcolor{omniStripe}
AC & Animation continuity & Smoothness and temporal coherence.\\
VC & Visual consistency & Stability of appearance and render state.\\
\midrule
\OmniGroup{3}{Efficiency $\downarrow$}
\rowcolor{omniStripe}
Tok & Token usage & Total generation token consumption.\\
Time & End-to-end time & Wall time for generation and repair.\\
\midrule
\OmniGroup{3}{Repair-policy metrics $\uparrow$}
\rowcolor{omniStripe}
RS@1 & Repair success@1 & Valid target-error correction with one patch.\\
CPR & Constraint preservation & Unaffected, initially valid constraints retained.\\
\rowcolor{omniStripe}
ELR & Edit localization ratio & Edits inside audited relevant code regions.\\
\bottomrule
\end{tabularx}
\end{table*}

\subsection{Evaluation Metric Rationale and Validation}
\label{app:metric_validation}

Our evaluation separates executability, visual quality, instructional presentation, repair behavior, and resource use. These dimensions answer different questions: a video may render successfully while containing an incorrect equation, and a localized edit may fail to repair its target. We therefore interpret each score within its stated scope rather than treating any single metric as a complete measure of educational quality.

\paragraph{Design rationale and related evidence.}
VBench evaluates video generation along distinct quality dimensions and checks their alignment with human judgments~\citep{huang2024vbench}. This motivates dimension-specific evaluation, but our OV/LQ/AC/VC formulas are task-specific proxies, not adopted VBench metrics. G-Eval studies rubric-based model scoring and its correlation with human judgments in text generation~\citep{liu2023geval}; work on LLM-as-a-judge also documents evaluator biases and tests agreement with human preferences~\citep{zheng2023judging}. These studies motivate explicit rubrics and independent validation. Their findings do not establish the validity of our formulas or of Claude's judgments on educational animations.

\paragraph{Executability and efficiency.}
R@1 and R@F measure successful rendering before and after the allowed repair, respectively. They establish execution outcomes, not factual correctness or fulfillment of every teaching requirement. Their measurement depends on consistent success, failure, and timeout handling, supported by execution logs and output checks. Tok and Time measure resource use under the reported deployment configuration. Token counts depend on model tokenization, and wall time depends on hardware, concurrency, API latency, and rendering settings. Equal repair-attempt budgets alone do not establish equal compute. Section~\ref{app:efficiency_accounting} specifies the deployment accounting boundary and distinguishes latency, token consumption, and offline training costs.

\paragraph{Visual quality.}
OV, LQ, AC, and VC target overlap, spatial organization, temporal continuity, and appearance consistency. Their definitions in Section~\ref{app:cv} use color components, foreground occupancy, motion, and palette changes as observable proxies. Color segmentation can merge colliding objects of the same color; edge occupancy cannot identify a required object that is already entirely outside the viewport; sparse sampling can miss brief failures. Intentional containment, geometric intersections, and scene transitions can also resemble defects. Consequently, these scores support comparisons under a fixed measurement procedure, but are not exact object-level defect counts or guarantees of collision-free motion.

\paragraph{Instructional presentation.}
The VLM rubric assesses content accuracy (CA), pedagogical clarity (PC), and engagement (EN) from sampled frames and the teaching requirement (Section~\ref{app:vlm}). The human study provides complementary judgments from full videos (Section~\ref{app:human_eval}). However, its CA anchors measure topical relevance, whereas VLM CA includes factual and mathematical correctness; the two are not interchangeable measures of accuracy. Human PC/EN describe perceived clarity and engagement, not demonstrated learning gains. Expert factual checks are needed to validate CA, while matched human ratings provide the relevant reference for PC/EN.

\paragraph{Repair quality.}
RS@1 measures target-error correction, CPR measures retention of initially satisfied non-target constraints, and ELR measures the fraction of edits within audited relevant code regions (Section~\ref{app:repair_protocol}). These measures distinguish successful correction, preservation, and localization. ELR alone does not establish semantic correctness, and CPR covers only the enumerated constraints. Independent review of before/after renders and relevant code regions is therefore needed to assess errors in repair verification and region annotation, rather than relying solely on the checks used to generate repair feedback.

\paragraph{Available evidence and its limits.}
The repeated VLM scoring check in Section~\ref{app:vlm} assesses run-to-run consistency. The human study measures perceived differences between methods. Neither establishes defect-detection precision/recall or per-video agreement between automatic and human scores. Deployment review and offline scoring have separate prompts and roles, but share Claude as evaluator, so procedural separation does not eliminate shared evaluator preferences. We do not report an independently annotated defect benchmark, expert factual-accuracy validation, or a second-evaluator comparison in the present results. The literature and existing checks support the evaluation design, but do not establish these forms of empirical validity.

\paragraph{Criteria for further validation.}
A direct validation would use held-out videos stratified by method, task, and failure type, with annotators blinded to method identity and automatic scores. Annotation would distinguish genuine collisions, partial and complete overflow, and transient failures from intentional overlap, recording affected regions and time intervals. Threshold calibration and final evaluation must use disjoint tasks, with thresholds frozen before testing. For a prespecified detection unit and matching rule, defect-wise precision, recall, and F1 would quantify false alarms and missed defects; a separate challenge subset could expose rare edge cases without being interpreted as the natural defect prevalence.

For continuous quality scores, per-video Spearman correlation with the corresponding human dimension and within-task method-order agreement would assess human alignment. Binary CA instead requires expert correctness labels and classification agreement. Confidence intervals must account for shared tasks and, where applicable, shared raters; annotator agreement should be reported before adjudication. Evaluating the same clips at 2 fps and denser sampling, up to native frame rate, would quantify missed transient events and additional cost. A second evaluator would test sensitivity to judge choice, while expert checks remain the reference for factual accuracy. These are validation criteria, not experiments claimed as completed here.

\section{Agent-Level Details}
\label{app:agent_details}

OmniManim integrates the Scene Agent, Vision Agent, Code Agent, and Repair Agent through a shared structured scene state. The Vision Agent supplies learned layouts and remains frozen during code-policy training and deployment.

\begin{table*}[!t]
\centering
\caption{Agent-level roles in OmniManim.}
\label{tab:agent_details}
\footnotesize
\OmniTableSetup
\begin{tabularx}{\textwidth}{@{}>{\raggedright\arraybackslash}p{0.105\textwidth} >{\raggedright\arraybackslash}p{0.265\textwidth} >{\raggedright\arraybackslash}p{0.265\textwidth} >{\raggedright\arraybackslash}X@{}}
\toprule
\rowcolor{omniHeader}
\textbf{Agent} & \textbf{Reads} & \textbf{Generates} & \textbf{Role} \\
\midrule
\rowcolor{omniStripe}
Scene Agent
& Teaching requirement, task metadata, and previous scene state
& Object set, relation set, pedagogical roles, and coarse temporal plan
& Converts the natural-language requirement into a structured scene specification.\\
Vision Agent
& Scene schema, object attributes, relations, previous layouts, and temporal cues
& Sparse keyframe bounding boxes and layout constraints
& Predicts object-level spatial layouts before code synthesis to reduce overlap, overflow, and clutter.\\
\rowcolor{omniStripe}
Code Agent
& Scene state, predicted layouts, object attributes, and temporal plan
& Executable Manim script, object-level code blocks, and animation schedule
& Converts the structured scene state into executable animation code.\\
Repair Agent
& Rendered frames, execution logs, CV diagnostics, and current scene state
& Local code patch with an identified edit region
& Repairs localized failures while preserving required content and unrelated constraints.\\
\bottomrule
\end{tabularx}
\end{table*}

\paragraph{Scene Agent.}
The Scene Agent parses the input teaching requirement into a structured scene specification and partitions it into scenes. It extracts the instructional topic, required visual objects, semantic relations among objects, pedagogical roles, and coarse temporal ordering. Separate model calls then write the scene programs concurrently; those scenes are rendered concurrently and concatenated into one video. Its output provides a unified semantic basis for the subsequent Vision Agent and Code Agent. The fixed prompt is given in Section~\ref{app:scene_prompt} and is decoded at temperature $0$.

\paragraph{Vision Agent.}
The Vision Agent predicts sparse keyframe layouts from the structured scene state. It reads object attributes, object relations, previous layouts, and temporal cues, and outputs normalized bounding boxes for objects at each sparse keyframe. Its purpose is to explicitly determine spatial layouts before code synthesis, thereby reducing element overlap, boundary overflow, and interpolation-time conflicts.

\paragraph{Code Agent.}
The Code Agent translates the scene state and predicted layouts into executable Manim code. It does not freely decide object positions from scratch; instead, it preserves the layouts predicted by the Vision Agent. Each object bounding box is mapped to Manim coordinates and scale parameters, while the temporal plan is converted into object creation, transformation, emphasis, and removal operations. The fixed prompt is given in Section~\ref{app:code_prompt}.

\paragraph{Repair Agent.}
The Repair Agent uses compiler evidence and render diagnostics to predict a local code patch. Spatial and temporal failures are associated with object identities, time intervals, and candidate AST regions through event traces and object-to-code mappings. The patch modifies the implicated placement, transformation, or lifecycle operations while preserving required content and unrelated constraints. The scene specification and frozen visual planner provide conditioning; repair does not update model parameters or retrain the planner. Deployment permits one repair attempt with one patch candidate. The fixed prompt is given in Section~\ref{app:repair_prompt}.

\section{Vision Agent Details}
\label{app:vision_details}

The Vision Agent is trained through coarse-prior learning followed by coordinate denoising. These two stages learn the visual layout model. SGCS and FLRPO subsequently optimize the code-and-repair policy while keeping the Vision Agent frozen. The visual-planner stages and the code-policy stages are described separately below.

\paragraph{Object representation.}
Each scene object is represented by its type, textual content, pedagogical role, temporal status, previous-keyframe box, size prior, and relation summary. We encode object content, retaining formula text, with the frozen \nolinkurl{sentence-transformers/all-MiniLM-L6-v2} checkpoint. Mean pooling over non-padding text tokens yields a 384-dimensional vector, followed by a trainable $384\rightarrow256$ projection. Object type, role, and temporal status use learnable embeddings. Two-layer MLPs encode the previous center--size box ($4\rightarrow256\rightarrow256$) and width--height prior ($2\rightarrow256\rightarrow256$). Missing geometry is zero-filled with an explicit validity indicator, so missing values are distinguishable from genuine zeros. Relation features aggregate relation-type embeddings and neighboring object features. Concatenated fields are projected to 256 dimensions and normalized with LayerNorm. The trainable scene encoder comprises these projections, embeddings, and fusion layers; it does not include updates to the pretrained text backbone.

\begin{table}[!t]
\centering
\caption{Implementation configuration of the Vision Agent.}
\label{tab:vision_arch}
\footnotesize
\OmniTableSetup
\begin{tabularx}{\columnwidth}{@{}>{\raggedright\arraybackslash}p{0.35\columnwidth} >{\raggedright\arraybackslash}X@{}}
\toprule
\rowcolor{omniHeader}
\textbf{Component} & \textbf{Configuration} \\
\midrule
\rowcolor{omniStripe}
Object token dimension & $256$\\
Categorical fields & Learnable embeddings for object type, pedagogical role, and temporal status\\
\rowcolor{omniStripe}
Text encoding & Frozen MiniLM-L6-v2; masked mean pooling; $384\rightarrow256$ projection\\
Geometry encoding & Box MLP $4\rightarrow256\rightarrow256$; size MLP $2\rightarrow256\rightarrow256$; validity indicators\\
\rowcolor{omniStripe}
Relation encoding & Aggregation over relation-type embeddings and neighboring objects\\
Spatial prior branch & Four $32\times32$ sigmoid heatmaps; Gaussian targets and MSE supervision\\
\rowcolor{omniStripe}
Coordinate denoiser & 4 blocks; hidden size 256; 8 attention heads; FFN size 1024\\
\rowcolor{omniStripe}
Block configuration & Pre-LayerNorm; GELU; dropout 0.1; object self-attention and prior cross-attention\\
Time conditioning & Sinusoidal flow-time embedding followed by an MLP to 256 dimensions\\
\rowcolor{omniStripe}
Prior conditioning & Spatial tokens with 2D positional encoding, used as cross-attention memory\\
Euler steps & $T_d=16$\\
\rowcolor{omniStripe}
Output head & Unbounded linear $256\rightarrow4$ velocity per object; no sigmoid\\
\bottomrule
\end{tabularx}
\end{table}

\paragraph{Denoiser architecture and conditioning.}
The denoiser has four pre-normalized Transformer blocks, each with eight attention heads, a 1024-dimensional feed-forward layer, GELU activation, and dropout 0.1. A box MLP embeds the current noisy coordinates into 256 dimensions. The scalar flow time $\tau=t/T_d$ receives sinusoidal encoding followed by an MLP to the same dimension. The initial object features are
\[
H_i=O_i+\mathrm{MLP}_{b}(b_i^\tau)
       +\mathrm{MLP}_{t}(\mathrm{PE}(\tau)).
\]
Self-attention exchanges information among object tokens; cross-attention conditions them on spatial tokens obtained from the coarse heatmaps with two-dimensional positional encoding. A linear $256\rightarrow4$ head predicts signed coordinate velocities without a sigmoid. Padding tokens are masked in attention and losses. Flow time is distinct from animation time: the former controls denoising, while object status and previous-keyframe geometry encode animation progression.

\paragraph{Coarse-prior supervision.}
Stage 1 predicts four semantic-region heatmaps for title, figure, equation, and annotation. Object roles determine the target channel. At a normalized location $(x,y)$ on the $32\times32$ grid, a visible object's ground-truth center--size box defines
\begin{equation}\begin{aligned}
G_i(x,y)&=\exp\!\left[-\frac{(x-x_i)^2}{2\sigma_{x,i}^2}
                         -\frac{(y-y_i)^2}{2\sigma_{y,i}^2}\right],\\
\sigma_{x,i}&=\max(w_i/2,1/32),\\
\sigma_{y,i}&=\max(h_i/2,1/32).
\end{aligned}\end{equation}
For channel $c$, $P_c^\star(x,y)$ is the pointwise maximum of $G_i(x,y)$ over visible objects assigned to $c$; an empty channel is zero. Channels are independent and may overlap. The prior head uses a separate sigmoid for each channel, rather than a mutually exclusive softmax, and minimizes
\begin{equation}
\mathcal L_{\mathrm{prior}}
=\frac{\|\mathbf P-\mathbf P^\star\|_F^2}{4\cdot32\cdot32}.
\end{equation}
This is the Stage~1 objective, separate from the Stage~2 loss in the main paper. Stage~2 keeps the learned prior-conditioning path fixed while updating the denoiser-side scene encoder and coordinate denoiser; the pretrained text encoder remains frozen throughout.

\paragraph{Flow matching and projected sampling.}
Stage~2 constructs \(\mathbf b^\tau=(1-\tau)\mathbf b^\star+\tau\boldsymbol\epsilon\), with \(\tau\in(0,1]\) and \(\boldsymbol\epsilon\sim\mathcal N(0,\mathbf I)\). These training interpolants are not clamped. For each sampled pair, the velocity target is \(\boldsymbol\epsilon-\mathbf b^\star\). Inference initializes $\mathbf b^{T_d}=\boldsymbol\epsilon$ without initial clipping and applies $T_d=16$ Euler updates in decreasing time, clamping each updated coordinate to $[0,1]$. The output velocity is signed and unbounded; clamping applies to coordinates, not velocities. The box-regression term uses the final prediction after the full 16-step rollout, rather than a one-step estimate of the clean box.

Without clipping, an oracle returning the same pair-specific velocity at every step would remove one sixteenth of the displacement per step, yielding $\mathbf b^0=\boldsymbol\epsilon-(\boldsymbol\epsilon-\mathbf b^\star)=\mathbf b^\star$. This identity describes an ideal trajectory only. Active projection changes that trajectory, and the learned conditional vector field need not remain equal to a single training pair's velocity along inference. Thus, 16 projected steps do not guarantee exact target recovery. Clamping center--size coordinates does not ensure full-box containment, so boundary penalties remain necessary.

\paragraph{Object lifecycle and loss masks.}
Let $a_i^{(k)}$ indicate that object $i$ is scheduled to exist at keyframe $k$. The states \texttt{new}, \texttt{keep}, and \texttt{move} receive current-frame box supervision. A new object has an invalid previous-box indicator; continuing objects retain their identity and previous geometry, without forcing their current box to equal the previous one. At a keyframe where an object has disappeared, its box target and geometric losses are masked out. Absence follows the event schedule, not a model's omission of a required object. Padding masks and lifecycle masks are distinct: padding is excluded from attention, while the lifecycle mask determines which real objects receive geometric supervision.

For $\|z\|_a^2=\sum_i a_i\|z_i\|_2^2/\max(1,\sum_i a_i)$, the velocity and final-box losses use
\begin{equation}\begin{aligned}
\mathcal L_{\mathrm{diff}}&=\mathbb E\|\hat v^t-(\epsilon-b^\star)\|_a^2,\\
\mathcal L_{\mathrm{box}}&=\mathbb E\|b^0-b^\star\|_a^2.
\end{aligned}\end{equation}
Only visible target objects contribute to prior heatmaps. Relation and boundary checks use the same scheduled visibility; removing a required object from a generated program does not remove its deployment visibility check.

\paragraph{Permitted overlap and containment.}
Let $m_{ij}^{(k)}=1$ when objects $i\ne j$ must avoid overlap, and zero when an explicit scene relation permits containment, intersection, or layering at that keyframe. Eligible ordered pairs satisfy
\[
\mathcal E_k=\{(i,j):i\ne j,\ a_i^{(k)}a_j^{(k)}m_{ij}^{(k)}=1\}.
\]
Permissions come from scene relations and their active intervals, not from observing a predicted overlap. An \texttt{inside} relation removes the generic collision penalty for that pair but retains containment checks: the child's left/right and bottom/top edges must remain within the parent's corresponding edges. Directional relations retain their own checks. Boundary constraints test $x_c-w/2\ge0$, $x_c+w/2\le1$, and the analogous vertical inequalities. These checks explain why valid coordinate ranges alone are insufficient.

\paragraph{Interpolation constraint.}
For adjacent keyframes $k$ and $k+1$, we approximate intermediate object trajectories by linear interpolation in normalized bounding-box space:
$$
\tilde b_i^{(k\rightarrow k+1)}(u)
=
(1-u)b_i^{(k)} + u b_i^{(k+1)}.
$$
We use $\mathcal U=\{0.25,0.50,0.75\}$. A trajectory is eligible for full-interval interpolation only when its object identity is matched and present at both endpoint keyframes. The set $\mathcal E_k(u)$ further requires both objects to be active at the sampled time and excludes overlap permitted by a relation active at that time. Birth/death transitions without two valid endpoint boxes are excluded from this interpolation loss and remain subject to lifecycle checks. Thus, the loss covers sampled linear box trajectories, not all rendered motion or lifecycle transitions. Curved paths, rotations, transformations, and failures between sampled times are outside any collision-free guarantee.

\paragraph{Loss weights.}
The fixed loss weights are:
$$
\lambda_1=2,\qquad
\lambda_2=5,\qquad
\lambda_3=3,\qquad
\lambda_4=3,\qquad
\lambda_5=2.
$$
The five terms correspond to bounding-box regression, collision penalty, relation constraint, boundary constraint, and interpolation-path collision penalty, respectively. The weights are selected on the validation split and kept fixed during testing.

\paragraph{Bounding-box extraction.}
For each source Manim animation, we render the script into frame sequences and sample sparse keyframes according to object creation, transformation, and removal events. Foreground regions are extracted using alpha masks when available and RGB thresholding otherwise. Candidate object regions are obtained by connected-component analysis and filtered by area and aspect ratio. Components across adjacent keyframes are matched using IoU and center-distance matching. Each retained component is converted into a normalized bounding box $(x,y,w,h)\in[0,1]^4$.

\begin{figure*}[!t]
\centering
\includegraphics[width=\linewidth]{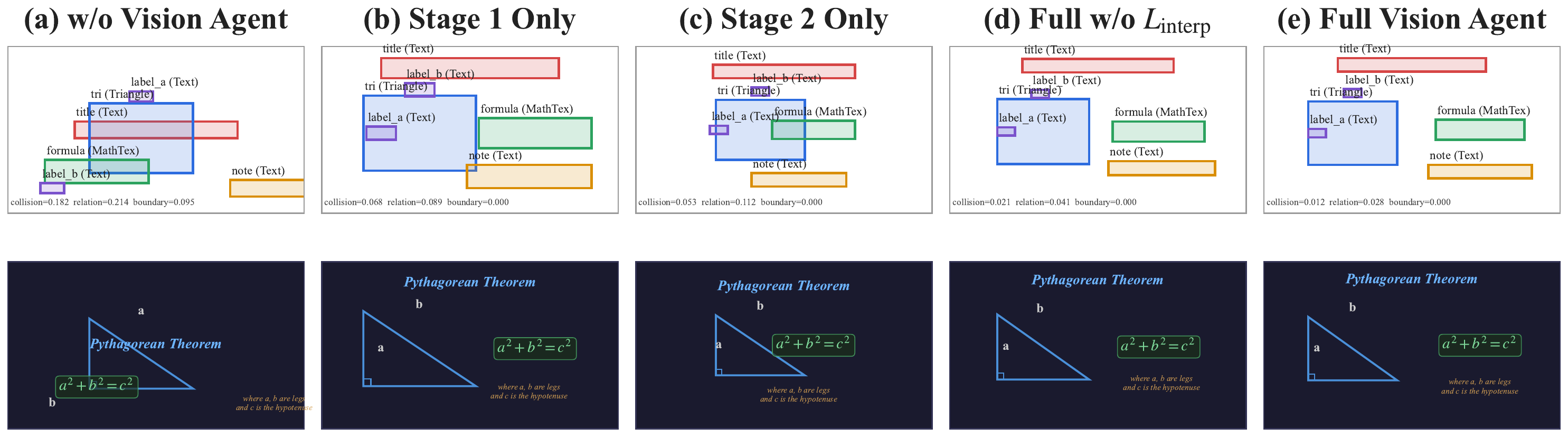}
\caption{Qualitative ablation within the Vision Agent. Top: object-layout predictions and spatial diagnostics. Bottom: corresponding rendered frames. Columns compare removal of the Vision Agent, coarse-prior-only planning, box-denoising-only planning, removal of the interpolation loss, and the complete visual planner. Stage 1/Stage 2 in this figure refer to the planner's coarse-prior and box-denoising stages, respectively, rather than SGCS/FLRPO policy training.}
\label{fig:ablation_qualitative_supp}
\end{figure*}

Figure~\ref{fig:ablation_qualitative_supp} compares variants of the visual planner. Its Stage 1/Stage 2 labels denote coarse-prior learning and box denoising, rather than SGCS/FLRPO code-policy training.

\section{Dataset Construction}
\label{app:dataset}

\subsection{ManimLayout-1K}
\label{app:training_data}

Training the Vision Agent requires paired examples of scene specifications and ground-truth object layouts. ManimLayout-1K contains 1{,}000 source educational Manim animations, from which we automatically construct keyframe-level structured records. Each extracted sample contains:

\begin{itemize}
    \item \textbf{Scene specification}: a scene-level text description (\texttt{scene\_text}), a list of objects with attributes (\texttt{id}, \texttt{type}, \texttt{content}, \texttt{role}, \texttt{status}), explicit inter-object relations (e.g., \texttt{above}, \texttt{left\_of}), and the previous keyframe layout (\texttt{prev\_layout}) for temporally continuing objects.
    \item \textbf{Ground-truth layout}: per-object bounding boxes $\{(\text{id}_i, x_i, y_i, w_i, h_i)\}_{i=1}^{N}$, where all coordinates are normalized to $[0,1]$. Object ordering is consistent with the scene specification.
\end{itemize}

Training data is constructed from high-quality open-source Manim code repositories collected from GitHub and community tutorial sites. We curate 1{,}000 educational animation scripts covering diverse topics, filter them for rendering correctness and visual quality, and then extract per-keyframe object bounding boxes via automated scene parsing. This process expands the source animations into 22{,}579 keyframe-level (scene, layout) samples grouped under 2{,}968 extracted scene identifiers. For each keyframe, we record the scene graph (objects, types, roles, relations) and the rendered spatial layout. Object status labels (\texttt{new}, \texttt{keep}, \texttt{move}, \texttt{disappear}) are derived by comparing consecutive keyframes. The representative subset prepared for release contains 500 records sampled uniformly in the original order. In that subset the observed statuses are \texttt{new} and \texttt{keep}, and parser-residual type names (\texttt{self}, \texttt{np}, \texttt{dict}, and \texttt{list}) are stored as \texttt{CustomMobject}. Its explicit relations are \texttt{inside}, \texttt{above}, and \texttt{left\_of}, plus 11 \texttt{centered\_on} and 4 \texttt{aligned\_with} links. The two training stages consume this data differently: Stage 1 converts visible ground-truth boxes into the four Gaussian target maps defined in Section~\ref{app:vision_details}, using channel-wise maxima and MSE supervision. Stage 2 uses clean boxes as flow-matching endpoints, with scheduled visibility and permitted-overlap masks applied to the geometric losses.

\paragraph{Dataset statistics.}
ManimLayout-1K contains 22{,}579 line-delimited JSON records. Each record contains at least one visual object and one ground-truth bounding box. The number of objects per keyframe ranges from 1 to 159, with a mean of 12.74 and a median of 8. The number of explicit relations per keyframe ranges from 0 to 133, with a mean of 2.28 and a median of 0, reflecting that many layout constraints are implicit in object roles and previous-frame layouts. Across all object instances, the most frequent pedagogical roles are geometric shape, group, equation, title, vector, annotation, container, main figure, coordinate system, and curve. All bounding-box coordinates are normalized to $[0,1]$.

\paragraph{Provenance and Licensing.}
For ManimLayout-1K, we only retain source animations from repositories or tutorial materials whose licenses or terms allow research use. During collection, we record the source URL, access date, repository commit when available, license identifier, and whether redistribution of the original source code is permitted. Sources with unclear, missing, or incompatible licensing are excluded. To reduce leakage risk, source animations are split by source-animation group, with repository/script identifiers retained for provenance, and EduRequire-500 requirements are independently authored rather than copied or paraphrased from the training sources. To avoid redistributing third-party code unnecessarily, our planned release focuses on derived scene--layout metadata, normalized bounding boxes, split files, and preprocessing scripts rather than republishing original Manim source files. The source-provenance table is maintained for auditability and will be released to the extent permitted by the corresponding licenses and terms of use.

\paragraph{Vision-Agent training and inference.}
To ensure reproducibility, we construct the training, validation, and test sets for the Vision Agent using a deterministic grouped split at the source-animation level, so that temporally adjacent keyframes from the same source animation are not assigned to different subsets. After keyframe decomposition, the resulting corpus contains 22{,}579 samples, of which 17{,}773, 2{,}681, and 2{,}125 are used for training, validation, and testing, respectively. The Vision Agent is trained in two stages. In both stages, we use AdamW with batch size $4$, learning rate $1\times 10^{-4}$, weight decay $1\times 10^{-4}$, and $20$ epochs. The hidden dimension is fixed to $256$, and the coordinate flow-matching module uses $T_d=16$ Euler steps. During requirement parsing, the Scene Agent uses the prompt in Section~\ref{app:scene_prompt} at temperature $0.0$. The relation vocabulary is limited to \texttt{left\_of}, \texttt{above}, and \texttt{inside}, so that it remains consistent with the subsequent layout optimization and repair process. In the second training stage, the prior-conditioning path and pretrained text encoder remain fixed; only the denoiser-side scene encoder and flow-matching denoiser are updated as specified in Section~\ref{app:vision_details}. At inference time, at most one round of repair is allowed. A layout is accepted as valid only when
$$
L_{\mathrm{coll}} \le 0.02,\qquad
L_{\mathrm{rel}} \le 0.15,\qquad
L_{\mathrm{bound}} \le 0.01.
$$
These thresholds flag candidate spatial defects for VLM review. Only a confirmed failure triggers the single local code-repair attempt; intentional layering is preserved. The same geometric checks are used for post-repair validation.

\paragraph{Compute resources.}
\label{app:vision_compute}
We train the Vision Agent on 4 NVIDIA GeForce RTX 3090 GPUs with 24 GiB memory each, using an internal server equipped with 8 GPUs in total. The server has an Intel Xeon Gold 6240C CPU at 2.60GHz with 72 CPU threads and 251 GiB RAM. The two-stage training takes approximately 43.5 hours in total, including 1.6 hours for Stage 1 and 41.9 hours for Stage 2. The Vision-Agent training environment uses Python~3.7.6, PyTorch~1.13.0+cu117, and NumPy~1.18.1; end-to-end generation and rendering use the Python~3.12 and Manim~0.19.0 environment described in the main evaluation protocol.

Using allocated GPU-hours (four devices multiplied by elapsed training hours), these durations correspond to approximately 6.4 GPU-hours for Stage~1 and 167.6 GPU-hours for Stage~2, or 174.0 GPU-hours in total. This accounting measures allocated device time, not utilization-weighted computation or energy consumption. The server inventory and planner-training allocation do not specify the resources assigned to an individual deployment request.

\subsection{EduRequire-500}
\label{app:eval_benchmark}

Each evaluation instance is a structured record containing:

\begin{itemize}
    \item \textbf{ID}: A unique identifier for reproducible referencing and cross-system comparison.
    \item \textbf{Requirement}: A natural language description specifying the desired visual content, including the target concept, expected visual elements, layout constraints, and pedagogical intent (e.g., ``Animate the derivation of the quadratic formula with step-by-step equation transformations'').
    \item \textbf{Task type}: A categorical label describing the pedagogical operation required by the prompt, such as concept explanation, problem solving, mathematical derivation, proof, algorithm design, data analysis, or system analysis.
    \item \textbf{Discipline}: A categorical label indicating the subject domain (e.g., algebra, geometry, calculus, physics, statistics, engineering, and literature), enabling fine-grained cross-domain analysis.
\end{itemize}

All requirements are authored by domain experts from the corresponding subjects, ensuring realistic instructional intent, appropriate difficulty calibration, and subject-specific terminology. The benchmark contains 500 tasks spanning a diverse range of complexity levels, from elementary visual demonstrations to multi-step mathematical derivations, with broad coverage across core school subjects and specialized university-level topics.

\paragraph{Dataset statistics.}
EduRequire-500 contains 500 JSON records with unique integer IDs from 1 to 500. The benchmark covers 120 subject labels and 16 task types. The most frequent task types are problem solving (129 tasks), concept explanation (118), mathematical derivation (68), algorithm design (65), system analysis (29), system design (19), data analysis (18), proof (14), and literature analysis (13), with the remaining 27 tasks covering translation, reading comprehension, reaction mechanisms, grammar correction, syntax rewriting, protocol analysis, and pathway analysis. Requirement descriptions contain 21.66 words on average, with a median of 22 and a range from 12 to 33 words.

\paragraph{Task examples and difficulty.}
Table~\ref{tab:edurequire_examples} shows representative requirements sampled from EduRequire-500. The expected visual elements are descriptive annotations used to clarify what a Manim animation would naturally need to instantiate; they are not additional input fields given to the generation systems. Difficulty labels are assigned according to a fixed rubric before system evaluation, based on task type, subject level, number of reasoning steps, and expected visual/narrative complexity. They are used only for descriptive analysis and benchmark breakdowns; they are not provided to any generation system during evaluation. Under this rubric, EduRequire-500 contains 132 easy tasks (26.4\%), 246 medium tasks (49.2\%), and 122 hard tasks (24.4\%).

\begin{table*}[!t]
\centering
\caption{Representative EduRequire-500 task examples. Requirements are taken from the benchmark; visual elements and difficulty labels summarize the expected animation structure for descriptive analysis.}
\label{tab:edurequire_examples}
\footnotesize
\OmniTableSetup
\begin{tabularx}{\textwidth}{@{}>{\raggedright\arraybackslash}p{0.14\textwidth} >{\raggedright\arraybackslash}X >{\raggedright\arraybackslash}p{0.26\textwidth} >{\raggedright\arraybackslash}p{0.09\textwidth}@{}}
\toprule
\rowcolor{omniHeader}
\textbf{Task type} & \textbf{Example requirement} & \textbf{Expected visual elements} & \textbf{Difficulty} \\
\midrule
\rowcolor{omniStripe}
Concept explanation &
State Newton's Third Law of Motion and identify three distinct macroscopic phenomena that physically demonstrate action-reaction pairs. &
paired force arrows, interacting objects, short labels &
Easy\\
Problem solving &
Calculate the total flight time, horizontal displacement, and final velocity vector for a projectile launched horizontally from a 1.8m platform at 3m/s, neglecting air resistance (g=10m/s\textsuperscript{2}). &
coordinate axes, parabolic trajectory, velocity vectors, equations &
Medium\\
\rowcolor{omniStripe}
Mathematical derivation &
Derive the differential form of the continuity equation for a compressible fluid using the principle of mass conservation within a control volume. &
control volume, flux arrows, density terms, equation transformations &
Medium\\
Proof &
Formulate a rigorous proof of the Law of Cosines using vector dot products within a two-dimensional Cartesian coordinate framework. &
triangle, coordinate axes, vectors, angle labels, algebraic steps &
Medium\\
\rowcolor{omniStripe}
Data analysis &
Analyze an acid-base titration curve for a weak acid and strong base, identifying the buffer region, equivalence point, and appropriate indicator selection. &
titration curve, highlighted regions, labeled points, indicator marker &
Medium\\
Literature analysis &
Analyze the strategic use of environmental description to reflect the protagonist's psychological isolation and societal critique in Lu Xun's \emph{A Madman's Diary}. &
scene panels, character-state labels, environmental motifs, theme arrows &
Medium\\
\rowcolor{omniStripe}
System analysis &
Calculate the inverse Jacobian matrix for a serial manipulator and evaluate the manipulability ellipsoid to identify and resolve kinematic singularities. &
robot arm, joint frames, matrix blocks, ellipsoid, singular state &
Hard\\
\bottomrule
\end{tabularx}
\end{table*}

\subsection{Policy-Training Data Construction and Accounting}
\label{app:policy_data_accounting}

\paragraph{Source collection and split inheritance.}
The policy-training data are derived from the training sources of ManimLayout-1K. We collect 3,246 candidate animation scripts from GitHub repositories and community tutorial materials permitted for research use. The exclusion counts comprise 1,418 rendering failures, 527 non-educational scripts, and 301 visually inadequate scripts, leaving 1,000 source animations. These are non-overlapping exclusions, so $3{,}246-1{,}418-527-301=1{,}000$. Source animations are assigned to splits before deriving keyframes, code targets, failures, or patches. Versions and near-duplicate copies of the same animation are kept in the same source group. Every derived record inherits its source split; temporally adjacent frames and related repair examples cannot cross splits. Repository identifiers support provenance tracking, while the partition unit is the source-animation group. Only the training partition supplies optimization examples. Validation sources support model selection and reward settings; held-out sources remain excluded from training.

\paragraph{Generation supervision.}
The code-policy supervision unit is an independently renderable scene program. Multiple keyframe records from the same scene are aggregated into a layout sequence rather than counted as separate full-program examples. Teaching requirements and scene descriptions are derived from the source animation's content and accompanying textual descriptions, not from evaluation requirements. A generation example stores $(z,c^\star)$, where $z$ contains the requirement, scene structure, and layouts predicted by the frozen Vision Agent. Execution traces and AST mappings align object identities with code variables, placement and scale operations, and animation events. The original source code is not assumed to satisfy the predicted layout: mismatches are corrected by code adaptation and verified by rendering, or the example is discarded. The retained target $c^\star$ must execute and implement the supplied scene and layout constraints.

\paragraph{Verified repair supervision.}
Repair examples are constructed from naturally occurring model-generation failures and controlled perturbations of training-source code. Perturbations concern placement, scale, timing, and object lifecycle. Each record stores $(z,c,d,p^\star)$: the scene conditions, failing program, diagnostic, and target patch. Diagnostics identify the affected objects, temporal interval, candidate code region, and error evidence. Teacher-assisted or manual patches are retained only after execution and rendering verify both correction of the target failure and preservation of unrelated constraints. Invalid patches are excluded from positive supervised targets. Failure cases with ambiguous object/time-to-code attribution are excluded from localized policy training. All variants of a source program retain its source-group assignment.

\paragraph{Failure pool for policy optimization.}
FLRPO samples from a fixed pool of failure states constructed from training sources. The pool contains failed programs and their diagnostics, rather than a dataset of preselected successful patches. Candidate patches are generated online by the current behavior policy. Within each failure state, all candidates are independently applied to the same unmodified program, with object identities, checking times, and target constraints fixed before sampling. Invalid candidates receive the prescribed penalty; their execution attempts still count toward the rollout budget. The SFT/GRPO comparison uses the same supervision corpus and failure pool across the corresponding controlled variants. EduRequire-500, TheoremExplainBench, and VisualEDU evaluation tasks are excluded from supervision, failure mining, and reward tuning.

\begin{table*}[!tbp]
\centering
\caption{Data provenance and training-budget accounting. Corpus cardinalities, unique policy examples, and repeated RL sampling events are reported separately.}
\label{tab:policy_data_accounting}
\footnotesize
\OmniTableSetup
\begin{tabularx}{\textwidth}{@{}>{\raggedright\arraybackslash}p{0.34\textwidth} >{\raggedleft\arraybackslash}p{0.18\textwidth} >{\raggedright\arraybackslash}X@{}}
\toprule
\rowcolor{omniHeader}
\textbf{Quantity} & \textbf{Count} & \textbf{Unit and scope}\\
\midrule
\OmniGroup{3}{Source corpus and layout records}
Collected candidates & 3,246 & Source animation scripts before filtering\\
\rowcolor{omniStripe}
Excluded: rendering failure & 1,418 & Scripts; disjoint exclusion count\\
Excluded: non-educational & 527 & Scripts; disjoint exclusion count\\
\rowcolor{omniStripe}
Excluded: inadequate visual quality & 301 & Scripts; disjoint exclusion count\\
Retained source animations & 1,000 & Entire corpus, before split-specific derivation\\
\rowcolor{omniStripe}
Extracted scenes & 2,968 & Entire corpus; not the training-scene count\\
Layout records, total & 22,579 & Keyframe-level scene--layout records\\
\rowcolor{omniStripe}
Layout records, train / val / test & 17,773 / 2,681 / 2,125 & Records after grouping by source animation\\
\midrule
\OmniGroup{3}{Source-group partition}
Source animations, train / val / test & 800 / 100 / 100 & Disjoint source-animation groups\\
\rowcolor{omniStripe}
Scenes, train / val / test & 2,336 / 352 / 280 & Sum to 2,968 scenes\\
\midrule
\OmniGroup{3}{Policy-data cardinalities}
Generation pairs & 2,300 & Retained scene-level $(z,c^\star)$ records\\
\rowcolor{omniStripe}
Repair examples & 6,900 & Verified repair records after filtering and deduplication\\
Total supervised records & 9,200 & Sum of generation and repair records; not unique scenes\\
\rowcolor{omniStripe}
RL failure-state pool & 4,600 & Unique failure states before repeated sampling\\
Held-out repair programs & 500 & Unique programs in the common held-out failure set\\
\midrule
\OmniGroup{3}{Configured RL budget, per model and training configuration}
Optimizer updates per seed & 500 & One update per rollout batch\\
\rowcolor{omniStripe}
Failure-state draws per seed & 4,000 & $500\times8$ draws; states may repeat\\
Candidate execution attempts per seed & 16,000 & $500\times8\times4$ patch candidates\\
\rowcolor{omniStripe}
Candidate attempts across three seeds & 48,000 & $3\times16{,}000$ for completed runs\\
\bottomrule
\end{tabularx}
\end{table*}

\paragraph{Counting conventions.}
Table~\ref{tab:policy_data_accounting} separates corpus cardinalities from configured training budgets. In particular, 17,773 is the number of training keyframe records, not the number of supervised code programs. The 2,968 scenes span the entire corpus, not just its training partition. A generation pair, a failed program, a failure state, and a verified repair patch are distinct counting units: one program may have several diagnostics or repair targets. Sampling a state again does not create a new unique state, and sampling four candidate patches does not guarantee four successful executions or four valid repairs. The 1:1 generation/repair sampling ratio specifies the supervised training mixture, not equal corpus sizes. Generation pairs and repair records are counted after verification and deduplication. Unique RL states are counted before repeated sampling.

\paragraph{Requirement construction and patch collection.}
Training requirements are derived from source titles, comments, and scene content using a fixed text template. GPT-5.4 drafts the requirement/scene descriptions and candidate repair patches at temperature 0; the executable target remains the source-derived program adapted to the frozen planner's layouts. Teacher outputs are accepted only after code execution and the prescribed visual checks. No evaluation requirement is used in this construction. Object identities, temporal events, and AST alignments are retained with each record.

\paragraph{Filtering and data reuse.}
Table~\ref{tab:policy_filter_counts} details construction of the generation and repair corpora. The generation stage starts with 2,421 candidate scene-program records, including alternative targets for the same scene, so candidate counts need not equal unique source-scene counts. Of these, 1,764 already satisfy the planner-conditioned constraints and 657 require adaptation; 536 adaptations pass, yielding 2,300 generation pairs. Failure mining constructs 9,200 distinct states from 2,764 natural generation failures and 6,436 controlled perturbations. Removing 318 ambiguously localized states leaves 8,882 eligible states. Teacher patch proposals are filtered sequentially by executability, target correction, preservation, and deduplication. Multiple proposals may be generated for a single state. The resulting supervised corpus contains 6,900 repair records. The fixed RL pool contains 4,600 selected eligible states, comprising 1,673 natural failures and 2,927 perturbed states. Supervised repair and RL may reuse training-source states; their counts are not disjoint source counts.

\paragraph{Held-out repair set.}
The common repair set contains 500 distinct failed program variants derived exclusively from held-out source animations. Multiple variants may share a source animation, but none shares a source-animation group with training. Its primary error labels comprise 121 overlap cases, 107 boundary/scale cases, 94 spatial-relation cases, 103 timing/lifecycle cases, and 75 compilation/API cases. Each variant receives one mutually exclusive primary label for reporting, although additional defects may coexist. Every configuration receives the same 500 programs and diagnostics and produces one candidate patch per program. RS@1 uses all 500 cases without excluding failed patch executions.

\begin{table*}[!tbp]
\centering
\caption{Policy-data construction and filtering statistics. Patch-verification counts are sequential; pool-origin and evaluation-error counts are disjoint within their respective groups.}
\label{tab:policy_filter_counts}
\footnotesize\OmniTableSetup
\begin{tabularx}{\textwidth}{@{}>{\raggedright\arraybackslash}p{0.49\textwidth}r>{\raggedright\arraybackslash}X@{}}
\toprule
\rowcolor{omniHeader}
\textbf{Construction stage} & \textbf{Count} & \textbf{Unit / relationship}\\
\midrule
\OmniGroup{3}{Generation corpus}
Candidate scene-program records & 2,421 & May include alternative targets per scene\\
Directly aligned / requiring adaptation & 1,764 / 657 & Disjoint candidate subsets\\
Adaptations accepted / rejected & 536 / 121 & Sum to 657\\
Retained generation pairs & 2,300 & $1{,}764+536$\\
\midrule
\OmniGroup{3}{Failure mining and supervised patch filtering}
Natural / perturbed failure states & 2,764 / 6,436 & Total 9,200 distinct candidate states\\
Ambiguous / eligible states & 318 / 8,882 & Disjoint filtering outcome\\
Candidate patch proposals & 10,284 & Multiple proposals per state permitted\\
Executable proposals & 8,193 & 2,091 rejected at execution stage\\
Target-correcting proposals & 7,286 & 907 additional rejections\\
Constraint-preserving proposals & 7,041 & 245 additional rejections\\
Deduplicated repair records & 6,900 & 141 duplicate records removed\\
\midrule
\OmniGroup{3}{RL pool and repair evaluation}
RL pool: natural / perturbed & 1,673 / 2,927 & Total 4,600 unique training states\\
Held-out repair programs & 500 & Distinct variants; grouped by source animation\\
Overlap / boundary and scale / relation & 121 / 107 / 94 & Primary error counts\\
Timing and lifecycle / compilation and API & 103 / 75 & Primary error counts; all five sum to 500\\
\bottomrule
\end{tabularx}
\end{table*}

\subsection{Code and Data Release}
\label{app:release}

We document the data format, annotation conventions, provenance policy, and evaluation protocol in the supplementary material. A representative data package, containing the 500-record ManimLayout subset described above and the full \textbf{EduRequire-500} benchmark together with README and schema documentation, will be released publicly. The ManimLayout-1K subset contains derived scene--layout JSONL records and normalized bounding boxes, without redistributing original third-party Manim source code or source identifiers. The EduRequire-500 file contains evaluation requirements and task metadata used by our protocol. These files are intended to make the record format, task diversity, and annotation conventions inspectable.

To support reproducibility beyond the paper text, we intend to release the main research artifacts subject to the provenance and licensing constraints described above. Specifically, the release is planned to include: (1) the full evaluation toolkit, including the CV-based metric implementation, VLM judging prompts, and result aggregation scripts; (2) the training code for the Vision Agent and the end-to-end OmniManim pipeline; and (3) both datasets introduced in this work, namely \textbf{ManimLayout-1K} and \textbf{EduRequire-500}, together with data documentation and preprocessing instructions. Dataset release will follow the provenance and licensing policy described above, and we will provide environment specifications and usage instructions so that the reported experiments can be reproduced from the released assets.

\section{Experimental Protocol}
\label{app:experimental_protocol}

This section describes the implementation, training, and evaluation of OmniManim. Baselines, module ablations, and alternative training strategies are identified explicitly.

\subsection{Implementation and Scoring}
\label{app:implementation}
\paragraph{Baseline budgets.}
Self-Refine uses one self-feedback revision; the other systems retain their planning and code-generation roles under the shared deployment budget. Methods receive the same requirements and diagnostic services, with decoding/output limits matched within each model group.

\paragraph{Model configurations.}
The main and transfer comparisons evaluate agent-only OmniManim with GPT-5.4 and Gemini, and SGCS/FLRPO-trained OmniManim with Qwen3-14B, Gemma, and Llama. Every OmniManim entry uses scene planning, learned visual layouts, code generation, and render-feedback-aware repair. Agent-only means that the code-and-repair backbone receives no SGCS or FLRPO updates; it still uses the trained Vision Agent. Planner inference is part of deployment, while planner training is an offline cost. Only the trained open-model entries update the code policy with SGCS and FLRPO. Main-paper Table~4 ablates the trained Qwen3-14B configuration. Row 5 enables Base, Vision, SGCS, and FLRPO. Row 3 is agent-only Qwen. Human evaluation uses agent-only GPT-5.4.

Controlled runs use Python~3.12, Manim Community~0.19.0~\citep{manim2025}, and seeds $\{17,23,42\}$. Open checkpoints are \nolinkurl{Qwen/Qwen3-14B}~\citep{yang2025qwen3}, \nolinkurl{google/gemma-3-12b-it}~\citep{google2025gemma3card}, and \nolinkurl{meta-llama/Llama-3.1-8B-Instruct}~\citep{meta2024llama31card}. Each uses its official chat template with text-only scene plans and structured diagnostics; Gemma receives no pixels, and Qwen uses non-thinking mode (\texttt{enable\_thinking=False}). The Vision Agent and Gemma visual encoder remain frozen. Methods share task inputs, diagnostic services, decoding/output limits within each model group, and one repair attempt with one candidate. All automated generation comparisons report the mean and sample standard deviation over three seeds, displayed as mean$_{\pm\mathrm{SD}}$. For configurations involving policy optimization, policies are fitted independently for each seed. Configurations with fixed policy parameters repeat generation under the same seed protocol. The same deployment repair budget applies to all reported configurations.

\paragraph{Scoring and aggregation.}
Table~\ref{tab:metrics} defines eleven generation metrics covering rendering, instruction, visual quality, and efficiency. Frozen CV code scores full-length renders at 2 fps after scaling each frame to a width of 512 pixels; tasks are averaged within each run, without pooling datasets. Following LLM-as-a-judge protocols~\citep{liu2023llava, zheng2023judging, liu2023geval}, Claude Opus 4.6 uses deterministic decoding and fixed prompts on requirements, teaching plans, and sampled keyframes: CA is binary, while PC/EN are scored on 0--100 and normalized. Render rates use the full task set, and token/time measurements cover generation and repair under the shared deployment budget. The external benchmarks are evaluated with the same eleven metrics. Offline judgments are independent of training rewards and deployment stopping; VLM procedures and CV algorithms are given in Sections~\ref{app:vlm} and~\ref{app:cv}.

\paragraph{Deployment timing boundary.}
\label{app:efficiency_accounting}
Time measures per-task elapsed wall-clock time from the start of requirement processing until the complete video has been written, or the run terminates in failure or timeout. It includes scene parsing, visual planning, code generation, initial execution and rendering, deployment diagnostics, any triggered repair and re-rendering, and final concatenation and output. API waiting, retry backoff, and failed execution attempts within the task are included. An API transport retry is distinct from the single permitted code-repair attempt. Failed and timed-out tasks retain their elapsed durations rather than being assigned zero time or excluded solely because they failed.

\paragraph{Deployment versus offline evaluation.}
CV checks, frame extraction, and VLM calls used to decide whether a patch is needed belong to deployment and are included in its timing. Offline CA/PC/EN judging, benchmark CV scoring, and human evaluation are separate assessment activities and are excluded from deployment time. Thus, Claude calls used for deployment review and calls used for offline scoring have different accounting roles even though they use the same evaluator. The 2 fps sampling rate and 512-pixel frame width describe the offline CV scoring procedure, not the native frame rate or resolution of generated videos.

\paragraph{Parallelism and interpretation of latency.}
OmniManim writes and renders independent scene programs concurrently, then concatenates their outputs. Wall-clock time follows the execution's critical path and is not the sum of overlapping request or process durations. Shared requirements, diagnostic services, and a one-patch repair budget control task access and repair opportunities; they do not by themselves establish equal concurrency or compute consumption. The reported times therefore compare the deployed systems, including their scheduling, rather than isolate an algorithmic reduction in computational work.

For example, the GPT-5.4 comparison on EduRequire-500 reports 72\,s for OmniManim and 148\,s for Code2Video, a 51.4\% latency reduction or approximately $2.06\times$ speedup. The corresponding token counts are 73{,}279 and 69{,}532, respectively: OmniManim uses 5.4\% more reported tokens in this comparison. These results support lower deployment latency despite greater token consumption; they do not establish lower API fees, energy use, or total compute.

\paragraph{Token and training-cost scope.}
Tok retains the generation-token accounting used in Table~\ref{tab:metrics} and the main result tables. It is distinct from elapsed time and monetary cost: tokenizer conventions and provider pricing prevent direct conversion from aggregate token counts to a hardware-independent compute measure or a common currency cost. Latency and token counts are interpreted alongside execution success and video quality.

Offline Vision-Agent, SGCS, and FLRPO training are excluded from per-task deployment time. Section~\ref{app:vision_compute} reports planner-training resources and elapsed time; this shared offline cost is separate from each task's planner inference. For each code backbone and seed, SGCS uses 1,296 optimizer updates across three epochs, while FLRPO uses 500 updates with eight failure states and four patches per state, totaling 16,000 candidate execution attempts. These are training-workload counts, not measured GPU-hours. Candidate generation, execution, rendering, and reward computation are part of the FLRPO training workload, rather than deployment evaluation. Planner training is a shared offline cost, separate from per-task inference.

\paragraph{Across-run mean and sample standard deviation.}
For each dataset, method, and metric $m$, we first compute a benchmark-level score separately for each seed $s\in\mathcal S=\{17,23,42\}$. Let $\mathcal V_{s,m}$ be the set of evaluation cases, $n_{s,m}=|\mathcal V_{s,m}|$, and $x_{s,i,m}$ the per-case score. The run mean and the reported summary are
\begin{equation}
\begin{aligned}
\mu_{s,m}&=\frac{1}{n_{s,m}}\sum_{i\in\mathcal V_{s,m}}x_{s,i,m},\\
\bar\mu_m&=\frac{1}{K}\sum_{s\in\mathcal S}\mu_{s,m},\qquad K=3,\\
\mathrm{SD}_m&=\sqrt{\frac{1}{K-1}\sum_{s\in\mathcal S}
(\mu_{s,m}-\bar\mu_m)^2}.
\end{aligned}
\label{eq:supp_run_aggregation}
\end{equation}
Thus, SD is the sample standard deviation of the three run means (\texttt{ddof=1}), not the standard deviation across tasks or the standard error of the mean. Each run receives equal weight. Repair metrics follow the rules in Section~\ref{app:repair_protocol} and use the same across-run aggregation.

\subsection{Code-Policy Training Configuration}
\label{app:code_policy_training}

Table~\ref{tab:code_policy_training} specifies the SGCS and FLRPO optimization settings for the Qwen3-14B, Gemma, and Llama code policies. The complete training recipe applies SGCS followed by FLRPO. The Qwen ablation disables only the stages indicated in main-paper Table~4. LoRA adapters are applied to attention and MLP projection layers of the language model; the underlying pretrained weights, Vision Agent, and Gemma visual encoder remain frozen. SGCS learns generation and repair through task-specific markers, and FLRPO continues from the resulting adapter parameters. The ablation follows the stage-removal rules in Section~\ref{app:ablation_protocol}.

\begin{table*}[!t]
\centering
\caption{Code-policy training configuration. Settings apply to the Qwen3-14B, Gemma, and Llama code policies; ablations retain the settings of each enabled stage. Batch sizes are effective sizes across devices and gradient accumulation.}
\label{tab:code_policy_training}
\footnotesize
\OmniTableSetup
\begin{tabularx}{\textwidth}{@{}p{0.28\textwidth}XX@{}}
\toprule
\rowcolor{omniHeader}
\textbf{Setting} & \textbf{SGCS} & \textbf{FLRPO}\\
\midrule
Adaptation & \multicolumn{2}{l}{LoRA on language-model attention and MLP projections}\\
\rowcolor{omniStripe}
LoRA rank / scaling / dropout & \multicolumn{2}{l}{$r=32$, $\alpha_{\mathrm{LoRA}}=64$, dropout $=0.05$}\\
Precision & \multicolumn{2}{l}{BF16}\\
\rowcolor{omniStripe}
Optimizer & \multicolumn{2}{l}{AdamW; $(\beta_1,\beta_2)=(0.9,0.999)$; weight decay $=0.01$}\\
Maximum gradient norm & \multicolumn{2}{l}{$1.0$}\\
\rowcolor{omniStripe}
Learning-rate schedule & \multicolumn{2}{l}{3\% warmup followed by cosine decay}\\
Random seeds & \multicolumn{2}{l}{$\{17,23,42\}$}\\
\midrule
\rowcolor{omniStripe}
Learning rate & $1\times10^{-4}$ & $5\times10^{-6}$\\
Training duration & 3 epochs & 500 optimizer updates\\
\rowcolor{omniStripe}
Effective batch size & 32 supervised examples & 8 failure states; 4 patches per state\\
Generation / repair sampling & $1:1$ & Repair patches only\\
\rowcolor{omniStripe}
Spatial token weighting & $\omega=1$; marked tokens have weight 2 & Not used in the policy-gradient loss\\
Repair loss coefficient & $\lambda_{\mathrm{repair}}=1$ & Not applicable\\
\rowcolor{omniStripe}
Candidates per failure state & Not applicable & $M=4$\\
Candidate sampling & Not applicable & Temperature $=0.8$; top-$p=0.95$\\
\rowcolor{omniStripe}
Updates per rollout batch & Not applicable & 1\\
GRPO clipping parameter & Not applicable & $\epsilon_c=0.2$\\
\rowcolor{omniStripe}
KL coefficient & Not applicable & $\eta_{\mathrm{KL}}=0.02$\\
KL reference & Not applicable & Fixed SGCS policy\\
\rowcolor{omniStripe}
Input token limit & 8192 & 8192\\
Repair output token limit & 2048 & 2048\\
\bottomrule
\end{tabularx}
\end{table*}

\paragraph{Supervised optimization.}
The SGCS batch samples generation and verified repair examples in equal proportions. Each task loss is averaged over its own examples, and the two losses are combined with $\lambda_{\mathrm{repair}}=1$. Prompt tokens are excluded from supervision. Aligned placement, timing, and object-reference tokens receive weight $1+\omega=2$; other output tokens receive weight 1. The input and repair-output limits in Table~\ref{tab:code_policy_training} are token budgets, distinct from the end-to-end token consumption reported in the main comparisons.

\paragraph{Epoch definition and sampling.}
Each effective supervised batch contains 16 generation and 16 repair records. An epoch shuffles and visits all 6,900 repair records once, pads its last batch with 12 resampled repair records, and samples the smaller generation corpus with replacement. This gives 432 updates per epoch and 1,296 updates over three epochs. Each task contributes 20,736 sampled examples over training, including repeated examples; this exposure count is distinct from the 2,300 and 6,900 unique-record counts. All supervised controls use the same sampler and number of updates. RL samples eight states uniformly with replacement from the fixed failure pool at each update, so its 4,000 state draws do not imply complete coverage of the pool.

\paragraph{Policy optimization and rollout budget.}
Each FLRPO update uses eight failure states and four independently sampled candidate patches per state. Every candidate is applied to the same unmodified starting program for that state, and the rollout batch is used for one optimizer update. The SGCS reference policy remains fixed throughout FLRPO. The configured budget is therefore $500\times8\times4=16{,}000$ candidate execution attempts per model and seed. This is a rollout budget, not a count of successful renders or unique failure states; compilation failures and repeated states do not constitute new successful renderings. Training-time group sampling does not change the deployment budget of one repair attempt with one candidate.

\subsection{FLRPO Reward Definition}
\label{app:flrpo_reward_definition}

\paragraph{Constraint satisfaction scores.}
For a failure diagnostic $d$, let $\mathcal C_{\mathrm{local}}(d)$ denote the checks associated with the implicated objects and time interval. Let $\mathcal C_{\mathrm{global}}(z)$ denote the checks over the full animation specified by scene state $z$. A check includes its relevant objects, temporal scope, and acceptance condition. Its binary outcome $s_a(c)\in\{0,1\}$ indicates whether program $c$ satisfies that condition after execution and rendering. For nonempty check sets, quality is the unweighted satisfaction rate:
\begin{equation}
Q_{\mathrm{local}}(c;d)
=\frac{\sum_{a\in\mathcal C_{\mathrm{local}}(d)}s_a(c)}
{|\mathcal C_{\mathrm{local}}(d)|},
\label{eq:supp_local_quality}
\end{equation}
\begin{equation}
Q_{\mathrm{global}}(c;z)
=\frac{\sum_{a\in\mathcal C_{\mathrm{global}}(z)}s_a(c)}
{|\mathcal C_{\mathrm{global}}(z)|}.
\label{eq:supp_global_quality}
\end{equation}
Local checks cover unintended overlap, viewport violations, specified spatial relations, and target-object visibility. Global checks cover spatial relations, object lifecycles, and event order across the animation. Semantically permitted containment and intentional overlap are excluded from collision violations. Object identities, check sets, and sampling times are fixed before candidate generation: deleting an object does not remove its checks, and a missing required object fails its visibility check. A non-renderable starting program receives the failure-floor values $Q_{\mathrm{local}}=Q_{\mathrm{global}}=0$.

\paragraph{Normalized edit cost.}
Let $N_{\mathrm{edited}}$ be the number of edited AST nodes, $N_{\mathrm{original}}$ the number of nodes in the starting program, and $N_{\mathrm{outside}}$ the number of edited nodes outside the permitted region $\Omega$ and its necessary dependencies. The edit cost is
\begin{equation}
\begin{aligned}
C_{\mathrm{edit}}(p;c,d)
={}&\frac12\min\!\left(1,\frac{N_{\mathrm{edited}}}{\max(1,N_{\mathrm{original}})}\right)\\
&+\frac12\frac{N_{\mathrm{outside}}}{\max(1,N_{\mathrm{edited}})}.
\end{aligned}
\label{eq:supp_edit_cost}
\end{equation}
The first term measures normalized patch size, and the second measures edits outside the allowed scope. Both are bounded in $[0,1]$, as is their average. This training cost uses the diagnostic's permitted edit region; the held-out ELR metric in Section~\ref{app:repair_protocol} uses independently audited relevant regions.

\paragraph{Reward and validity gate.}
For candidate program $c_j'$, define $\Delta Q_{\mathrm{local},j}=Q_{\mathrm{local}}(c_j';d)-Q_{\mathrm{local}}(c;d)$ and $\Delta Q_{\mathrm{global},j}=Q_{\mathrm{global}}(c_j';z)-Q_{\mathrm{global}}(c;z)$. The gate $G(c_j')$ is 1 only when executability, required content, and instructional order checks pass; otherwise it is 0. We use
\begin{equation}
\begin{aligned}
R_j={}&G(c_j')\big[\Delta Q_{\mathrm{local},j}
+0.25\Delta Q_{\mathrm{global},j}\big]\\
&-0.05C_{\mathrm{edit}}(p_j;c,d)-2\big[1-G(c_j')\big].
\end{aligned}
\label{eq:supp_flrpo_reward}
\end{equation}
Thus, $\alpha=1$, $\beta=0.25$, $\lambda_{\mathrm{edit}}=0.05$, and $\lambda_{\mathrm{fail}}=2$. Local correction provides the principal positive signal, global quality changes account for effects elsewhere in the animation, and edit cost discourages unnecessarily broad patches. Invalid candidates receive the failure penalty without quality credit; quality gains need not be evaluated for such candidates. These satisfaction rates are distinct from the aggregate OV/LQ scores defined in Section~\ref{app:cv}.

\subsection{FLRPO Training Procedure}
\label{app:flrpo_training_procedure}

\noindent\textbf{Inputs:} an SGCS-initialized code-and-repair policy, its fixed reference copy, training failure states $(z,c,d)$, the frozen Vision Agent, and the Manim execution and checking pipeline.
\begin{enumerate}
\item Initialize trainable adapter parameters from SGCS. Keep the reference policy fixed and use the hyperparameters in Table~\ref{tab:code_policy_training}.
\item For each of 500 updates, select eight failure states. For each state, fix the diagnostic objects, time interval, permitted edit region, and constraint checks before sampling patches. Compute the starting quality scores, using the failure floor for a non-renderable program.
\item Using the current behavior policy, sample $M=4$ patches independently for each state with temperature 0.8 and top-$p$ 0.95. Record their behavior-policy token log probabilities.
\item Apply each patch independently to the state's original program. Execute the candidate, evaluate the validity gate, and compute the edit cost. For valid candidates, render and evaluate local and global checks to obtain the quality gains.
\item Compute each candidate reward using Eq.~\ref{eq:supp_flrpo_reward}. Normalize rewards within each four-candidate group using the mean and standard deviation, with the numerical stabilizer in the main-paper objective. Equal-reward groups have zero relative advantage.
\item Take one GRPO optimizer update over the 32 candidates, using token-averaged sequence losses, clipping parameter 0.2, and KL coefficient 0.02 against the fixed reference. Discard the rollout batch and repeat from step 2 with the updated policy.
\end{enumerate}
After training, all parameters are fixed for deployment. The trained configuration deploys the FLRPO checkpoint with the frozen Vision Agent. The agent-only variant does not run SGCS or FLRPO and deploys the original backbone with the Vision Agent.

\subsection{Controlled Ablation}
\label{app:ablation_protocol}

Main-paper Table~4 compares four reduced variants of the trained Qwen3-14B configuration. Row 5 contains Base, Vision, SGCS, and FLRPO. Row 1 retains only the shared Base framework. Row 2 removes Vision while retaining SGCS and FLRPO. Row 3 is agent-only: Base and Vision, without SGCS or FLRPO. Row 4 removes FLRPO and deploys the SGCS checkpoint with Vision. Removing Vision omits layout conditioning during training and deployment but retains scene inputs and render diagnostics. Enabled stages retain their data and training budgets, with no budget reallocation. Tasks, seeds, diagnostics, and deployment repair budgets are fixed across all five configurations.

\subsection{Comparison with Standard SFT and GRPO}
\label{app:training_strategy_comparison}

The module ablation measures the contribution of enabling or removing training stages. Table~\ref{tab:sft_grpo_comparison} reports controlled comparisons against standard supervised fine-tuning (SFT) and GRPO with a global rendering reward. SGCS is a task-specific supervised objective, whereas FLRPO uses GRPO as its optimizer and specifies the failure-conditioned patch task and rendering reward. Thus, the GRPO baseline below controls the optimizer and patch interface while changing the reward formulation. Configurations A, C, and G correspond to rows 3, 4, and 5 of the main-paper module ablation, respectively. A removes both policy-training stages, C removes FLRPO, and G is the SGCS/FLRPO-trained configuration. Configurations B, D, E, and F replace the indicated training objectives with the matched SFT or GRPO-G controls.

\begin{table*}[!tbp]
\centering
\caption{Training-strategy comparison on EduRequire-500 with Qwen3-14B. All rows retain the shared Base framework and frozen Vision Agent; G is the SGCS/FLRPO-trained configuration, and A--F are controlled training variants. Entries are mean$_{\pm\mathrm{SD}}$ over three seeds. GRPO-G uses the global-reward control defined below. All metrics are higher-is-better; repair metrics use the common held-out failure set.}
\label{tab:sft_grpo_comparison}
\begingroup
\fontsize{7.5pt}{8.5pt}\selectfont
\OmniTableSetup
\setlength{\tabcolsep}{2pt}
\renewcommand{\arraystretch}{1.14}
\newcommand{\TSResult}[2]{\mbox{#1\ensuremath{_{\scriptscriptstyle\pm#2}}}}
\begin{tabularx}{\textwidth}{@{}l l l *{8}{>{\centering\arraybackslash}X}@{}}
\toprule
\rowcolor{omniHeader}
& \multicolumn{2}{c}{\textbf{Training}}
& \multicolumn{5}{c}{\textbf{Generation and rendering} $\uparrow$}
& \multicolumn{3}{c}{\textbf{Repair} $\uparrow$}\\
\rowcolor{omniHeader}
\textbf{ID} & \textbf{Supervised} & \textbf{RL} & R@1 & R@F & OV & LQ & AC & RS@1 & CPR & ELR\\
\midrule
A & None & None & \TSResult{0.714}{.033} & \TSResult{0.864}{.009} & \TSResult{0.695}{.013} & \TSResult{0.815}{.016} & \TSResult{0.717}{.010} & \TSResult{0.617}{.021} & \TSResult{0.937}{.008} & \TSResult{0.810}{.004}\\
\rowcolor{omniStripe}
B & SFT & None & \TSResult{0.798}{.014} & \TSResult{0.897}{.018} & \TSResult{0.742}{.016} & \TSResult{0.851}{.011} & \TSResult{0.749}{.013} & \TSResult{0.604}{.019} & \TSResult{0.929}{.007} & \TSResult{0.791}{.012}\\
C & SGCS & None & \TSResult{0.823}{.006} & \TSResult{0.916}{.029} & \TSResult{0.786}{.015} & \TSResult{0.873}{.008} & \TSResult{0.766}{.012} & \TSResult{0.665}{.013} & \TSResult{0.922}{.004} & \TSResult{0.857}{.002}\\
\rowcolor{omniStripe}
D & SFT & GRPO-G & \TSResult{0.815}{.019} & \TSResult{0.919}{.012} & \TSResult{0.761}{.018} & \TSResult{0.867}{.014} & \TSResult{0.763}{.009} & \TSResult{0.691}{.016} & \TSResult{0.938}{.006} & \TSResult{0.844}{.013}\\
E & SGCS & GRPO-G & \TSResult{0.837}{.011} & \TSResult{0.933}{.014} & \TSResult{0.789}{.012} & \TSResult{0.881}{.009} & \TSResult{0.780}{.015} & \TSResult{0.732}{.018} & \TSResult{0.946}{.005} & \TSResult{0.872}{.009}\\
\rowcolor{omniStripe}
F & SFT & FLRPO & \TSResult{0.826}{.016} & \TSResult{0.929}{.011} & \TSResult{0.779}{.014} & \TSResult{0.879}{.013} & \TSResult{0.773}{.011} & \TSResult{0.751}{.014} & \TSResult{0.949}{.004} & \TSResult{0.887}{.007}\\
\rowcolor{omniOurs}
G & SGCS & FLRPO & \TSResult{0.846}{.017} & \TSResult{0.947}{.009} & \TSResult{0.804}{.017} & \TSResult{0.894}{.024} & \TSResult{0.793}{.008} & \TSResult{0.783}{.010} & \TSResult{0.957}{.003} & \TSResult{0.909}{.006}\\
\bottomrule
\end{tabularx}
\endgroup
\end{table*}

\paragraph{Standard SFT control.}
The SFT control uses the same layout-conditioned generation examples, verified repair examples, task markers, and equal generation/repair sampling as SGCS. It sets $\omega=0$, assigning uniform weight to all supervised output tokens while retaining $\lambda_{\mathrm{repair}}=1$. Prompt tokens remain excluded. Architecture, adapter configuration, optimizer, training steps, and token limits follow Table~\ref{tab:code_policy_training}. Consequently, B versus C isolates spatial token weighting under matched supervision; it does not isolate the benefit of collecting repair examples or providing layouts.

\paragraph{Global-reward GRPO control.}
GRPO-G starts from the supervised checkpoint named in each row and uses that same checkpoint as its fixed KL reference. It retains the failure-state inputs, object/time diagnostics, patch output format, and candidate application procedure used by FLRPO. Its reward uses the full-animation constraint satisfaction gain from Eq.~\ref{eq:supp_global_quality}:
\begin{equation}
R_j^{\mathrm{G}}=G(c_j')\Delta Q_{\mathrm{global},j}
-2\big[1-G(c_j')\big].
\label{eq:supp_grpo_global_control}
\end{equation}
The validity gate and failure penalty match FLRPO. This control omits the diagnostic-local quality term and explicit edit-cost penalty, while global checks still cover required content, visibility, relations, and event order. GRPO-G uses the same GRPO optimizer as FLRPO with this global reward. E versus G therefore tests the FLRPO reward formulation with failure conditioning and patch actions held fixed; a separate diagnostic/action ablation would be needed to isolate those components.

\paragraph{Matched budgets and interpretation.}
All training-strategy comparisons use seeds $\{17,23,42\}$, identical source-level splits, and the same training failure pool. Each RL run uses 500 updates, eight failure states per update, and four candidate patches per state, totaling 16,000 candidate execution attempts per seed. Sampling settings, clipping, KL coefficient, and optimizer settings follow Table~\ref{tab:code_policy_training}; equal candidate-attempt budgets control rollout counts but do not imply identical rendering time. Deployment allows one repair attempt with one candidate for every row. D versus G compares the two complete training recipes; F versus G tests SGCS under FLRPO, and D versus E tests SGCS under GRPO-G. B versus D and C versus E measure the incremental effect of global-reward RL. Generation metrics use EduRequire-500, and repair metrics use the common failure set in Section~\ref{app:repair_protocol}. 

\subsection{Repair Evaluation}
\label{app:repair_protocol}

A common frozen set of held-out failing programs supplies identical code and diagnostics to each configuration; one predicted patch is verified by rendering. RS@1 measures target-error correction, and CPR averages per-program retention of initially satisfied non-target constraints. For edited AST nodes $\mathcal E(p)$ and independently audited relevant regions/dependencies $\Omega^\star$,
\begin{equation}
\label{eq:supp_elr}
\operatorname{ELR}(p)=\frac{|\mathcal E(p)\cap\Omega^\star|}{\max(1,|\mathcal E(p)|)}.
\end{equation}
ELR is averaged over programs before aggregation across seeds. Insertions are anchored to source regions, and no-op patches score zero. These conditional repair metrics complement the full-test-set generation scores.

\section{VLM Evaluation}
\label{app:vlm}

Following common LLM-as-a-judge protocols~\citep{liu2023llava, zheng2023judging, liu2023geval}, we use Claude Opus 4.6 as the vision-language evaluator with deterministic decoding. The following review and scoring tasks return structured JSON. Deployment review is separate from offline benchmark scoring:

\paragraph{Deployment Review.} For each CV-flagged suspicious segment, we send the start, middle, and end keyframes along with the CV diagnostics. The fixed prompt in Section~\ref{app:verify_prompt} classifies each segment as \texttt{PASS}, \texttt{FAIL}, or \texttt{INTENTIONAL}, and returns a confidence in $[0,1]$ and a severity in $\{0,1,2,3\}$.

\paragraph{Instructional Scores.} We send uniformly sampled, timestamped frames together with the teaching requirement and teaching plan. The fixed prompt in Section~\ref{app:instruction_prompt} returns CA in $\{0,1\}$ and PC/EN as integers from 0 to 100. Reported PC and EN divide those integers by 100. The requirement is the reference; the plan is not assumed correct.

All evaluation prompts are fixed across methods. We verified inter-run consistency by re-evaluating a random 10\% subset, observing a mean absolute deviation below 0.03. This check measures repeatability, not agreement with independent human or expert judgments; Section~\ref{app:metric_validation} distinguishes these forms of evidence.

\section{CV Metrics}
\label{app:cv}

The four visual scores, OV, LQ, AC, and VC, are reference-free video proxies computed from rendered frames. They do not use a vision-language model. The generated Manim program is inspected only to identify reliable animation boundaries. Components come from color segmentation, so the scores are not exact object-level collision or occlusion measurements. We uniformly sample $T$ frames at 2 fps and scale each frame to a width of 512 pixels. Let $[x]_0^1=\min(1,\max(0,x))$. All scores lie in $[0,1]$, and larger values indicate better visual quality.

\paragraph{Overlap (OV).}
Candidate components are color-based connected components. For two components $i$ and $j$ of different colors in frame $t$, the relative box overlap is
\begin{equation}
r_{ij}^{(t)}
=
\frac{|B_i^{(t)}\cap B_j^{(t)}|}
{\min(|B_i^{(t)}|,|B_j^{(t)}|)}.
\end{equation}
Only pairs with $r_{ij}^{(t)}\geq 0.08$ count as candidate overlaps. Let $A_t^{\mathrm{overlap}}$ be the fraction of foreground pixels covered by these candidate overlap regions, and let $R_t^{\mathrm{overlap}}$ be the maximum $r_{ij}^{(t)}$ in the frame, or zero when no pair qualifies. The overlap score is
\begin{equation}
S_{\mathrm{OV}}
=
\left[
1-\frac{1}{T}\sum_{t=1}^{T}
\left(
0.5A_t^{\mathrm{overlap}}
+
0.5R_t^{\mathrm{overlap}}
\right)
\right]_0^1.
\end{equation}
$A_t^{\mathrm{overlap}}$ measures how much of the foreground is affected, and $R_t^{\mathrm{overlap}}$ measures the most severe local overlap. Because the components are color segments, this is cross-color box overlap rather than the true occlusion area of the original Manim objects.

\paragraph{Layout quality (LQ).}
Each frame is partitioned into a $6\times 8$ grid. Let $\rho_{t,c}$ be the foreground occupancy of cell $c$ in frame $t$, and let $\mathcal C_t=\{c:\rho_{t,c}\geq 0.02\}$ be the active cells. Crowding is
\begin{equation}
D_t=
\frac{1}{|\mathcal C_t|}
\sum_{c\in\mathcal C_t}
\left[
\frac{\rho_{t,c}-0.35}{0.65}
\right]_0^1,
\end{equation}
and $D_t=1$ when $\mathcal C_t$ is empty, so a blank frame does not receive a perfect layout score. Let $F_t$ be the foreground and $\partial I_t$ the band within 2\% of the shorter image side of the frame boundary. The edge penalty is
\begin{equation}
E_t=
\left[
\frac{|F_t\cap \partial I_t|}
{0.1|F_t|}
\right]_0^1.
\end{equation}
The layout score is
\begin{equation}
S_{\mathrm{LQ}}
=
\left[
1-
\frac{1}{T}\sum_{t=1}^{T}
\left(
0.7D_t+0.3E_t
\right)
\right]_0^1.
\end{equation}
$D_t$ varies continuously with local density. $E_t$ measures contact with the edge band, not content that has already left the frame.

\paragraph{Animation continuity (AC).}
Let $m_t$ be the fraction of pixels that change substantially between frames $t-1$ and $t$, and let $\widetilde m$ be the median of $m_t$ over non-static frames. Motion-energy discontinuity and foreground-centroid acceleration are
\begin{equation}
\begin{aligned}
P_t^{\mathrm{energy}}
&=
\left[
\frac{|m_t-m_{t-1}|}
{\max(0.02,4\widetilde m)}
-0.25
\right]_0^1,\\
P_t^{\mathrm{centroid}}
&=
\left[
\frac{
\|\mathbf c_t-2\mathbf c_{t-1}+\mathbf c_{t-2}\|_2
}{25.6}
\right]_0^1,
\end{aligned}
\end{equation}
where $\mathbf c_t$ is the foreground centroid in this 512-pixel-wide frame, so the divisor $25.6$ is five percent of the evaluation width. Over the compared transitions $\mathcal T$, after excluding detected scene cuts and applicable animation boundaries,
\begin{equation}
P_{\mathrm{motion}}
=
\frac{1}{|\mathcal T|}
\sum_{t\in\mathcal T}
\max
\left(
P_t^{\mathrm{energy}},
P_t^{\mathrm{centroid}}
\right).
\end{equation}
$P_{\mathrm{transient}}$ is the number of significant box-overlap episodes that appear and then disappear during motion, divided by the number of motion frames and clipped to $[0,1]$. An episode is significant when its maximum relative box overlap is at least $0.35$. The continuity score is
\begin{equation}
S_{\mathrm{AC}}
=
\left[
1-
0.6P_{\mathrm{motion}}
-
0.4P_{\mathrm{transient}}
\right]_0^1.
\end{equation}
This penalizes abrupt motion and short-lived overlap during animation. It does not establish that two original object trajectories collided.

\paragraph{Visual consistency (VC).}
Let $H_t$ be the foreground HSV color histogram. Over adjacent comparisons $\mathcal T'$ that remain after excluding detected scene cuts,
\begin{equation}
P_{\mathrm{palette}}
=
\frac{1}{|\mathcal T'|}
\sum_{t\in\mathcal T'}
\mathbb{I}
\left[
\chi^2(H_t,H_{t-1})\geq 0.15
\right].
\end{equation}
Let $f_t$ be the foreground fraction and $n_t$ the number of color components. A foreground-state jump is flagged when either occupancy or component count changes sharply:
\begin{equation}
\begin{aligned}
P_{\mathrm{state}}
=
\frac{1}{|\mathcal T'|}
\sum_{t\in\mathcal T'}
\mathbb{I}\Big[
&\frac{|f_t-f_{t-1}|}
{\max(f_t,f_{t-1},0.02)}>0.40\\
&\lor\;
\frac{|n_t-n_{t-1}|}
{\max(n_t,n_{t-1},1)}>0.50
\Big].
\end{aligned}
\end{equation}
Blank$_t$ denotes an anomalous near-blank frame in the interior of the video, and Flash$_t$ denotes a brightness jump above $0.5$ between adjacent frames. The artifact rate is
\begin{equation}
P_{\mathrm{artifact}}
=
\frac{1}{T}
\sum_{t=1}^{T}
\mathbb{I}[\mathrm{Blank}_t\lor\mathrm{Flash}_t].
\end{equation}
The consistency score is
\begin{equation}
S_{\mathrm{VC}}
=
\left[
1-
0.4P_{\mathrm{palette}}
-
0.3P_{\mathrm{state}}
-
0.3P_{\mathrm{artifact}}
\right]_0^1.
\end{equation}

\paragraph{Edge-case conventions.}
Ratios are computed inside the per-frame loop. A blank frame, $|F_t|=0$, skips every sample whose denominator would be $|F_t|$: the component loop is not entered, and the implementation divides by $\max(1,|F_t|)$. Thus $E_t=0$, $A_t^{\mathrm{overlap}}=0$, and $R_t^{\mathrm{overlap}}=0$, with no $0/0$ value. The frame stays in the average and keeps $D_t=1$, so a blank frame is not treated as a perfect layout. If no frame has motion energy above $0.001$, the non-static median $\widetilde m$ is defined as $0$ and the energy denominator falls back to $0.02$. A missing foreground centroid sets $P_t^{\mathrm{centroid}}=0$. After scene cuts and animation boundaries are excluded, an empty $\mathcal T$ sets $P_{\mathrm{motion}}=0$; if no frame is in motion, $P_{\mathrm{transient}}=0$. An empty $\mathcal T'$ sets $P_{\mathrm{palette}}=P_{\mathrm{state}}=0$. An empty average returns $0$.

\subsection{Metric Scope and Limitations}
\label{app:metric_scope}

The CV metrics are intended to measure recurring render-level failures that are central to our evaluation setting: spatial collision, local overcrowding, unstable motion, and abrupt visual inconsistency. They complement the VLM and human evaluations, but they are not a substitute for expert judgment about pedagogy, factual correctness, or aesthetic style. Reported instructional-presentation and engagement gains should therefore be interpreted within EduRequire-500 and our rating protocol rather than as universal measures of instructional quality. OmniManim is designed around code-centric Manim animations; its behavior in open-domain video generation or highly stylized production settings may differ. The current Vision Agent operates at the object-bounding-box level, which directly supports layout repair but does not fully model typography, fine-grained graphic design, or cinematic effects. Longer-horizon temporal dependencies also remain challenging for extended animations.

\section{Human Evaluation}
\label{app:human_eval}

This section documents the human evaluation of agent-only OmniManim with GPT-5.4. Its code-and-repair backbone receives no SGCS or FLRPO updates. The study compares its rendered outputs against standalone GPT-5.4 and Code2Video on content, instructional presentation, and visual quality.

\subsection{Recruitment}
We recruited 100 undergraduate students enrolled in full-time programs at
multiple universities.
The participants collectively cover nine academic disciplines: theoretical and applied mechanics, environmental science,
automation, computer science, industrial design, cybersecurity,
materials science, electronic information engineering, and biological
science. Participants were recruited through open voluntary calls
within student communities. No participant had prior involvement with
the development of OmniManim or any of the baseline systems.

\subsection{Task Selection}
We selected 60 tasks from EduRequire-500 using stratified sampling across
the nine disciplines. Each task is assigned to 20 raters whose majors
are closest to the task's discipline whenever possible, so that
participants judge content within their domain of familiarity while
providing multiple independent assessments per task. Each participant
evaluates 12 tasks; in total, this yields 1{,}200 task-rater assignments.
For each selected task, each method generates one video. In each
task-rater assignment, the rater scores the three method outputs
(OmniManim, GPT-5.4, and Code2Video), yielding 1{,}200 individual ratings
per method per dimension and 20 independent ratings for each
task-method pair.

\subsection{Rubric}
For each task, participants are shown the educational requirement,
watch the three rendered videos in full, and assign a score in
$[0,100]$ on each of the following seven dimensions (translated from
the original Chinese instructions):

\begin{itemize}
    \item \textbf{CA} --- \textit{0 = severely off-topic; 100 = fully on-topic}.
    \item \textbf{PC} --- \textit{0 = not understandable; 100 = very clear}.
    \item \textbf{EN} --- \textit{0 = very dull; 100 = very engaging}.
    \item \textbf{OV} --- \textit{0 = severe overlap; 100 = almost no overlap}.
    \item \textbf{LQ} --- \textit{0 = chaotic; 100 = well-coordinated}.
    \item \textbf{AC} --- \textit{0 = very choppy; 100 = very smooth}.
    \item \textbf{VC} --- \textit{0 = highly inconsistent; 100 = fully consistent}.
\end{itemize}

All dimensions are phrased so that higher scores indicate better
quality. In this human rubric, CA measures topical relevance rather than
expert factual correctness; it should not be treated as a direct
validation of the VLM's factual-accuracy judgment. PC and EN measure
perceived presentation quality, not learning outcomes
(Section~\ref{app:metric_validation}).

\subsection{Presentation}
Evaluation sessions are conducted over Tencent Meeting with the
participant's screen visible to a supervisor. For each task the
supervisor shares the educational requirement, the participant watches
the three rendered videos in full without skipping, and then fills in
the scoring spreadsheet for all seven dimensions across the three
videos. The supervisor's role is limited to confirming that each video
is watched to completion before scoring; no feedback is provided on
scores. Participants may re-watch videos before finalizing their
scores.

\subsection{Blinding and Order}
Videos are presented with anonymized filenames
(\texttt{video\_A.mp4}, \texttt{video\_B.mp4}, \texttt{video\_C.mp4}),
and the filename-to-method mapping is independently randomized per
task. Participants are informed that three different systems produced
the videos but are not told which video corresponds to which system.

\subsection{Compensation}
Each participant received 100 RMB for participation. Compensation was
paid via mobile transfer after the session, irrespective of the scores
provided.

\subsection{Consent and Anonymity}
All participation was voluntary. Participants were informed of the
study purpose, approximate duration, compensation, and their right to
withdraw at any time without consequence. All participants gave
informed consent before taking part. Scoring records contain no
personally identifiable information; participants are referred to by
numerical IDs. The study collects no personal data beyond major and
institutional affiliation and poses no foreseeable risk; under our
institutional guidelines such minimal-risk anonymous studies do not
require formal IRB review. The study adheres to the conference code of
ethics.

\subsection{Data Handling}
All 100 participants submitted complete scores, yielding 1{,}200 task-rater
records and 20 independent ratings for each task-method pair. Three
cells in the VC column contained free-text comments appended to the
numerical score; comments are preserved for qualitative analysis, and
the numerical component is extracted for quantitative analysis.

\subsection{Statistical Analysis}
\begin{table}[!tbp]
\centering
\caption{Human evaluation on 60 EduRequire-500 tasks. Scores are means on 0--100. OmniManim is agent-only GPT-5.4. ICC is the one-way random-effects agreement of the 20 raters on each video, ICC$(1,20)$. $\Delta$ is OmniManim minus the baseline. $p$-values are paired Wilcoxon tests on 60 task means. $^{***}p<10^{-3}$, $^{**}p<10^{-2}$, $^{*}p<0.05$.}
\label{tab:human_eval}
\scriptsize\OmniTableSetup\setlength{\tabcolsep}{1.15pt}\renewcommand{\arraystretch}{1.05}
\begin{tabular}{@{}lcccccccccc@{}}
\toprule
& \multicolumn{2}{c}{\textbf{Omni}} & \multicolumn{4}{c}{\textbf{GPT-5.4}} & \multicolumn{4}{c}{\textbf{Code2Video}} \\
\cmidrule(lr){2-3}\cmidrule(lr){4-7}\cmidrule(lr){8-11}
& mean & ICC & mean & ICC & $\Delta$ & $p$ & mean & ICC & $\Delta$ & $p$ \\
\midrule
CA & \textbf{92.9} & 0.874 & 91.8 & 0.847 & $+1.1$ & $0.128$ & 89.7 & 0.853 & $+3.2$ & $0.032^{*}$ \\
PC & \textbf{85.9} & 0.855 & 85.5 & 0.835 & $+0.5$ & $0.134$ & 73.2 & 0.819 & $+12.7$ & $<10^{-4}{}^{***}$ \\
EN & \textbf{73.4} & 0.798 & 60.9 & 0.772 & $+12.5$ & $<10^{-3}{}^{***}$ & 58.5 & 0.779 & $+14.9$ & $<10^{-3}{}^{***}$ \\
OV & \textbf{74.1} & 0.865 & 57.8 & 0.804 & $+16.4$ & $0.003^{**}$ & 42.0 & 0.793 & $+32.1$ & $<10^{-5}{}^{***}$ \\
LQ & \textbf{73.2} & 0.859 & 65.0 & 0.797 & $+8.3$ & $0.041^{*}$ & 51.9 & 0.806 & $+21.4$ & $<10^{-5}{}^{***}$ \\
AC & 79.6 & 0.784 & \textbf{80.8} & 0.793 & $-1.2$ & $0.234$ & 78.0 & 0.787 & $+1.6$ & $0.352$ \\
VC & 87.5 & 0.850 & \textbf{89.9} & 0.854 & $-2.4$ & $0.172$ & 85.2 & 0.830 & $+2.3$ & $0.062$ \\
\midrule
All & \textbf{81.0} & 0.861 & 76.0 & 0.835 & $+5.0$ & $0.009^{**}$ & 68.4 & 0.833 & $+12.6$ & $<10^{-5}{}^{***}$ \\
\bottomrule
\end{tabular}
\end{table}
For each task $t$, method $m$, and dimension $d$, the 20 independent
rater scores are first averaged into a task-level method score:
\[
\bar{s}_{t,m,d}=\frac{1}{20}\sum_{r=1}^{20}s_{t,r,m,d}.
\]
Table~\ref{tab:human_eval} reports means over the 60 task-level scores
$\bar{s}_{t,m,d}$; Figure~5 of the main paper plots the same means. For pairwise comparisons between methods $A$ and
$B$, we compute task-level paired differences
\[
\Delta_{t,d}^{(A,B)}=\bar{s}_{t,A,d}-\bar{s}_{t,B,d},
\]
where $A$ denotes OmniManim in the comparisons reported in
Table~\ref{tab:human_eval}. Paired Wilcoxon signed-rank tests are then
computed over the 60 matched task-level differences, using the
\texttt{wilcox} zero-handling convention.
For each dimension $d$ and method $m$, we report the one-way
random-effects intraclass correlation ICC$(1,20)$ of the 20 raters on
each rendered video. Each coefficient is computed over the 60 videos
produced by method $m$, one per task, with each video contributing 20
scores on dimension $d$. The \emph{All} row applies ICC$(1,20)$ to
the per-rater average of the seven dimension scores on each video.

\subsection{Human Study Limits}
(i) Raters are undergraduate students rather than professional
educators; expert assessment of instructional presentation remains
future work. (ii) Although each task receives 20 independent ratings,
the study is still limited to 60 tasks and may be underpowered for small
effects.

\section{Additional Materials}
\label{app:materials}

\begin{figure*}[p]
\centering
\includegraphics[width=0.32\linewidth]{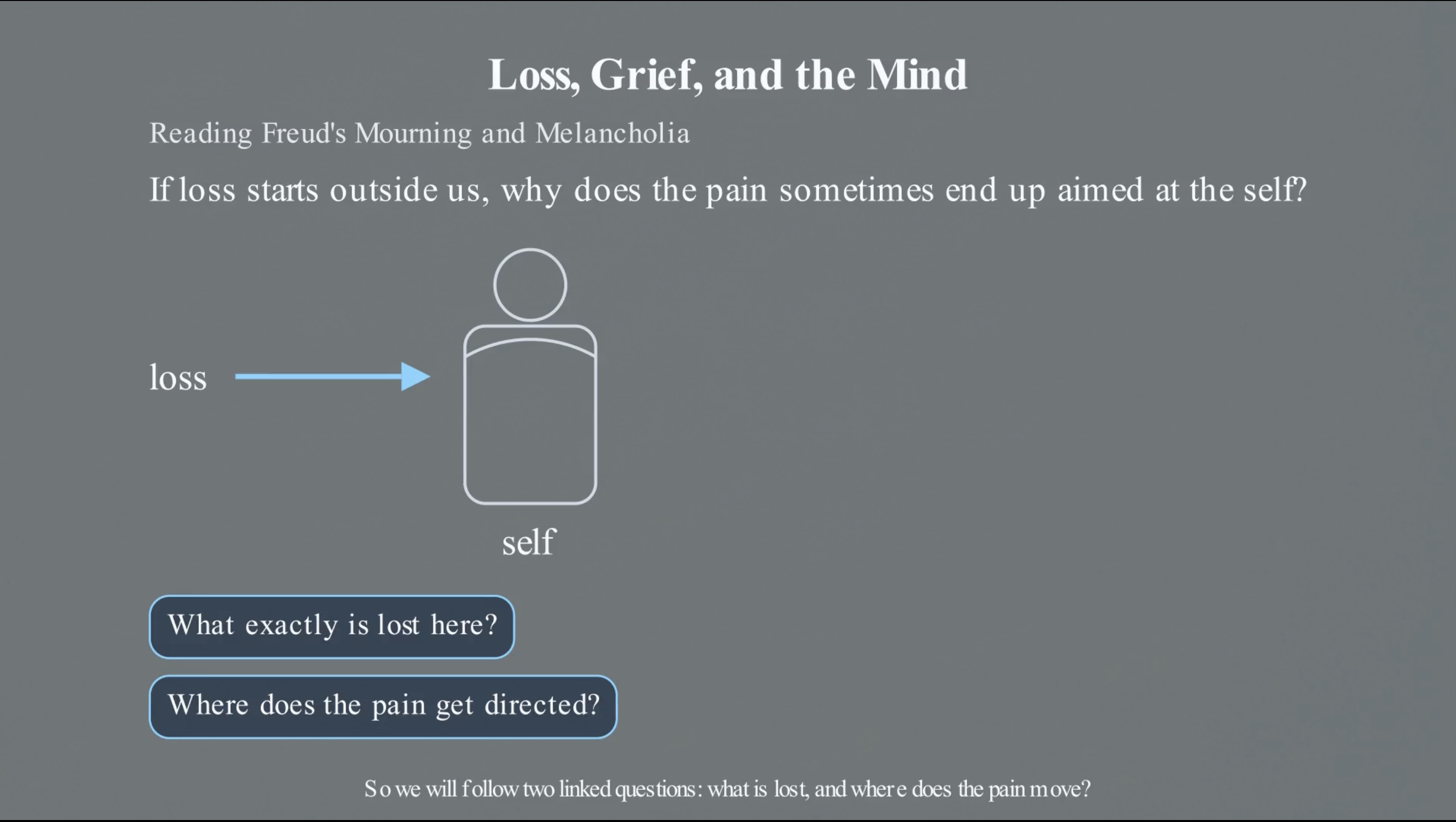}\hfill
\includegraphics[width=0.32\linewidth]{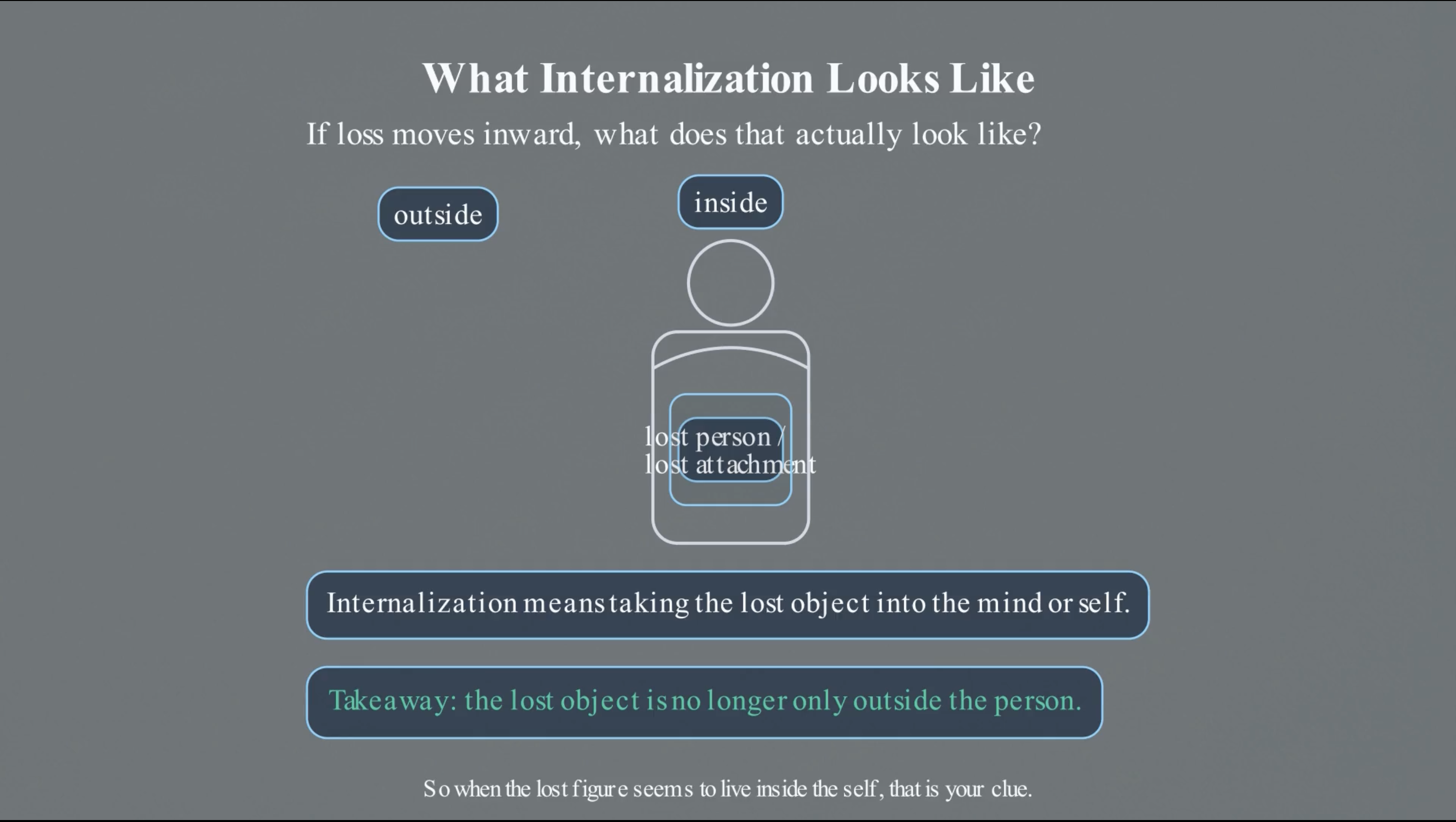}\hfill
\includegraphics[width=0.32\linewidth]{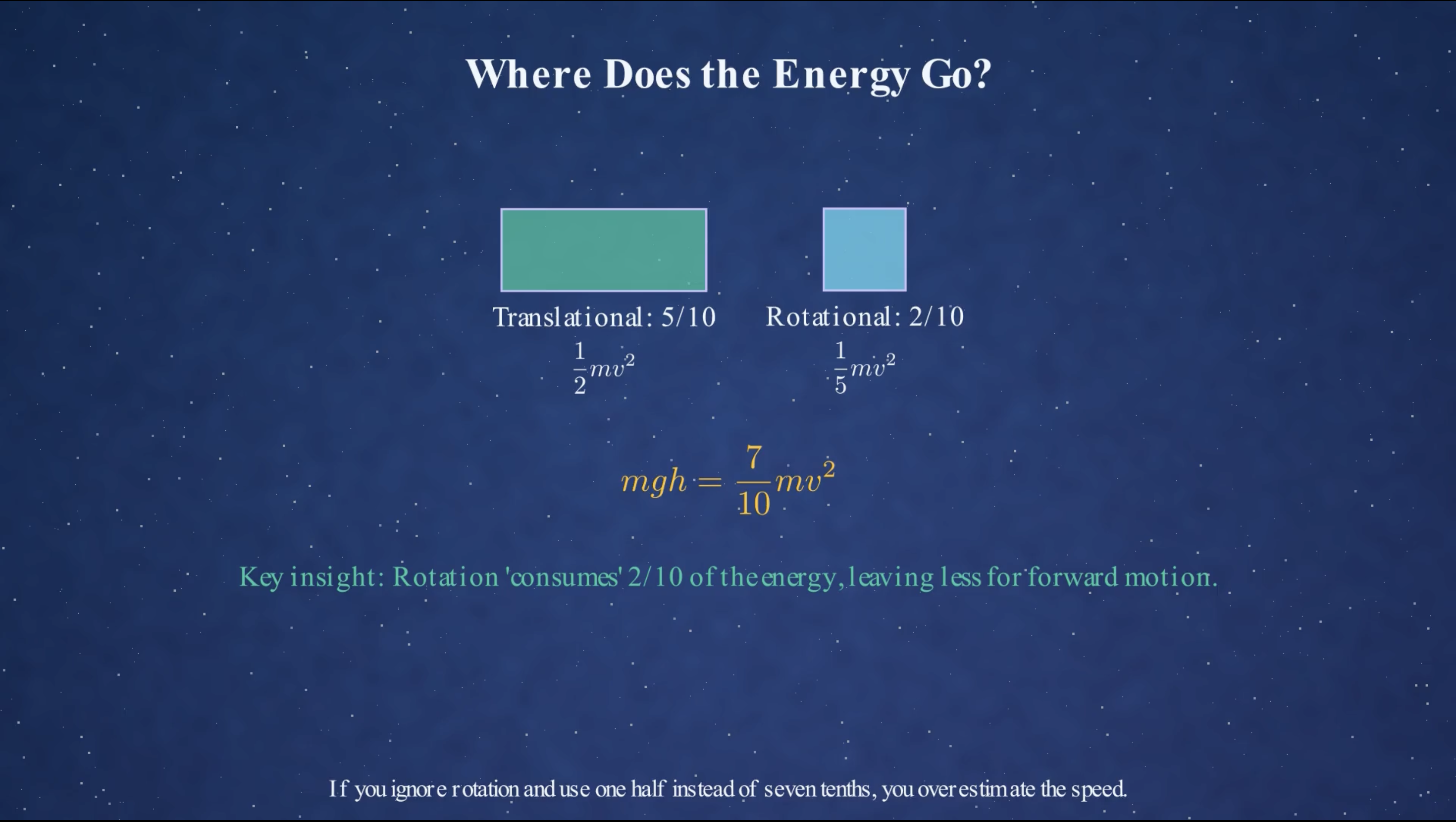}\\[0.35em]
\includegraphics[width=0.32\linewidth]{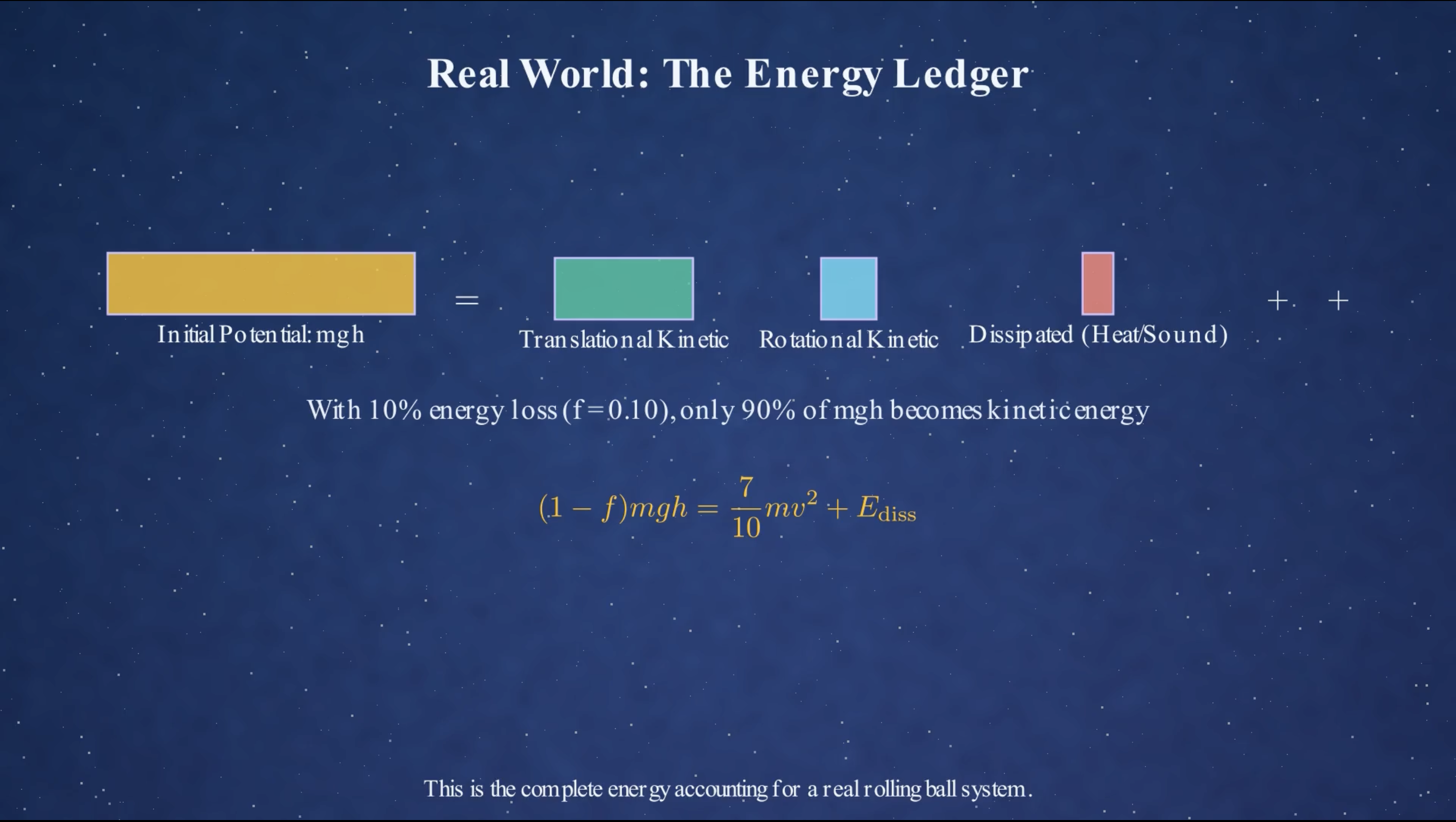}\hfill
\includegraphics[width=0.32\linewidth]{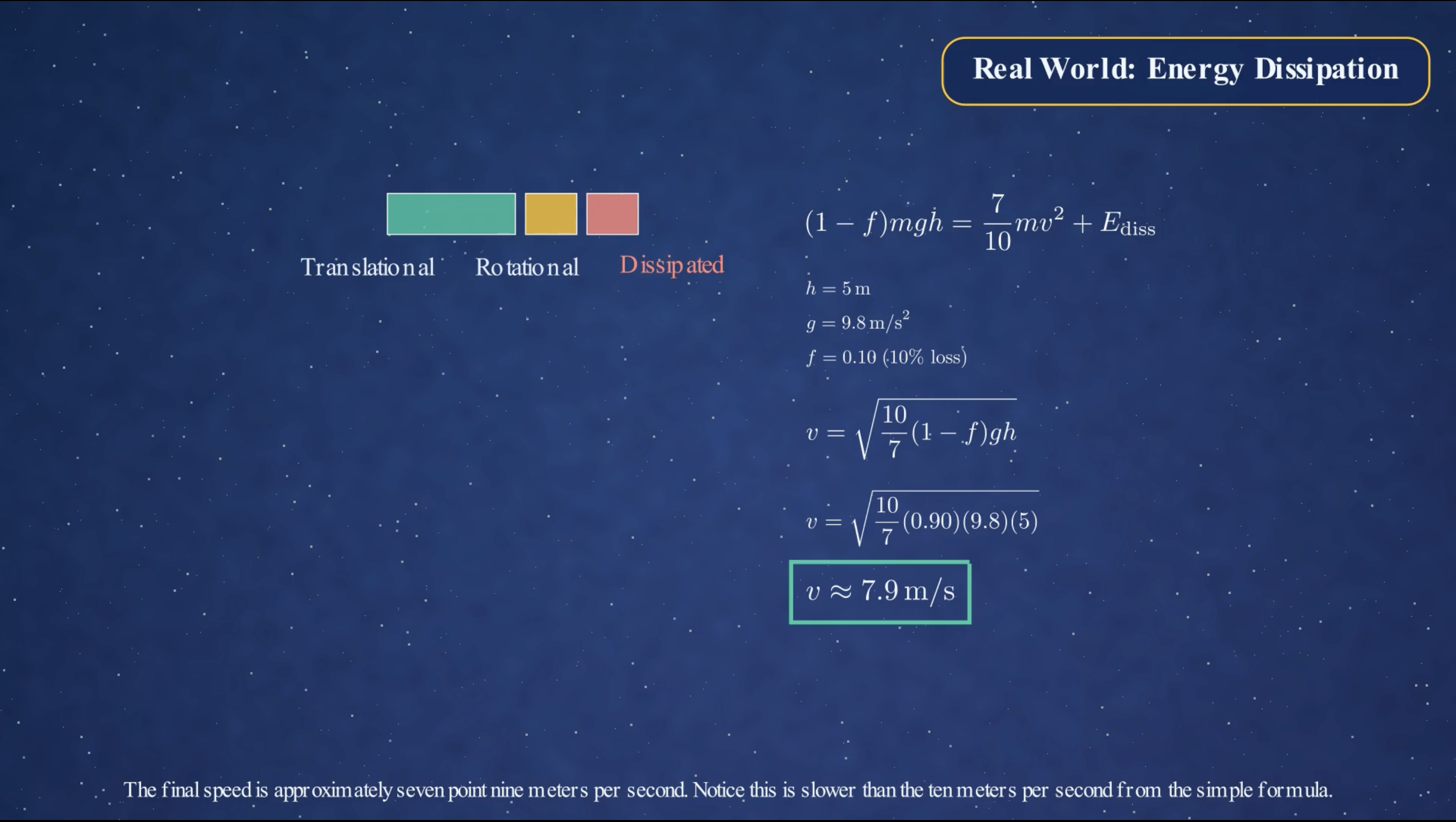}\hfill
\includegraphics[width=0.32\linewidth]{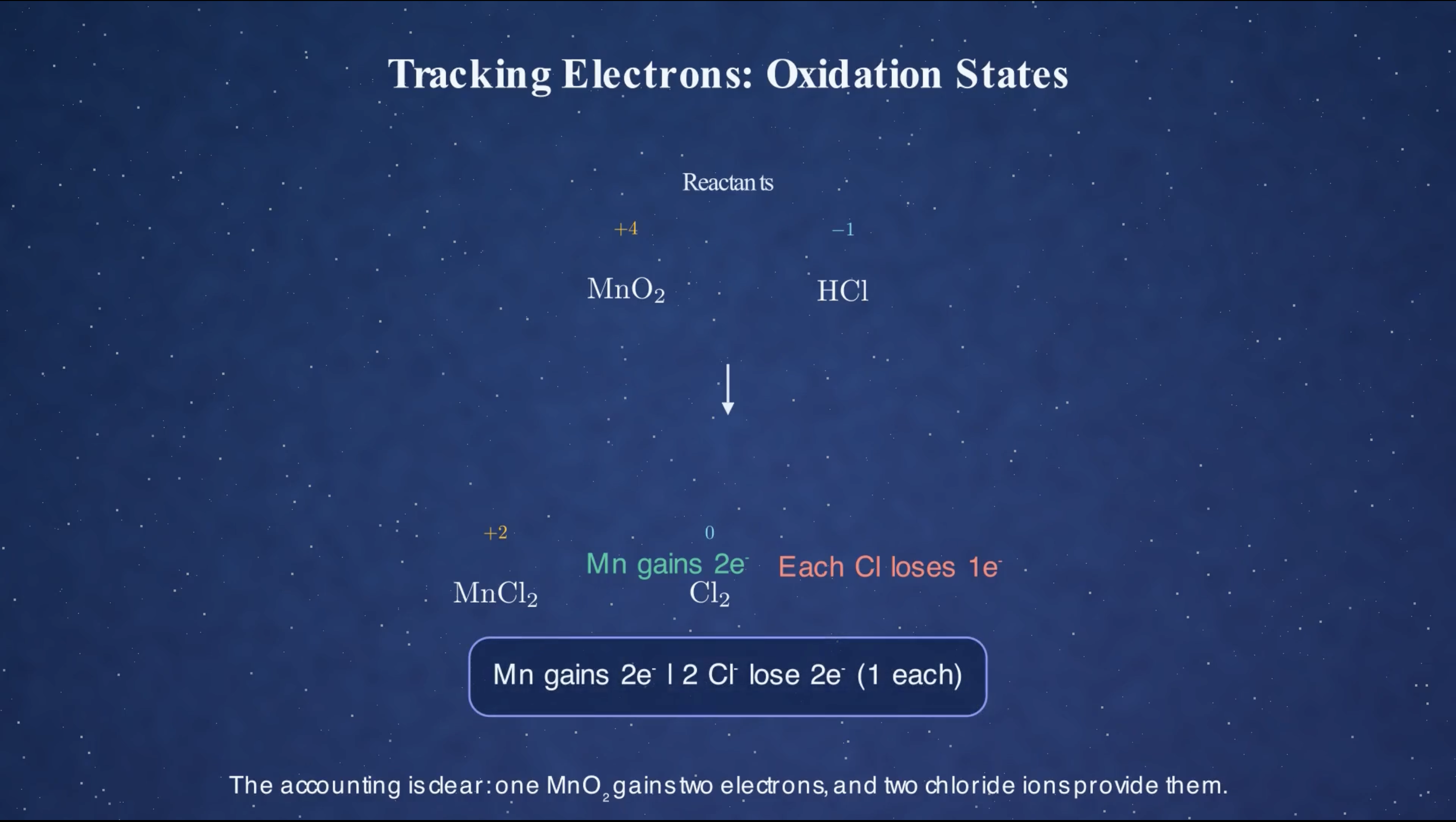}\\[0.35em]
\includegraphics[width=0.32\linewidth]{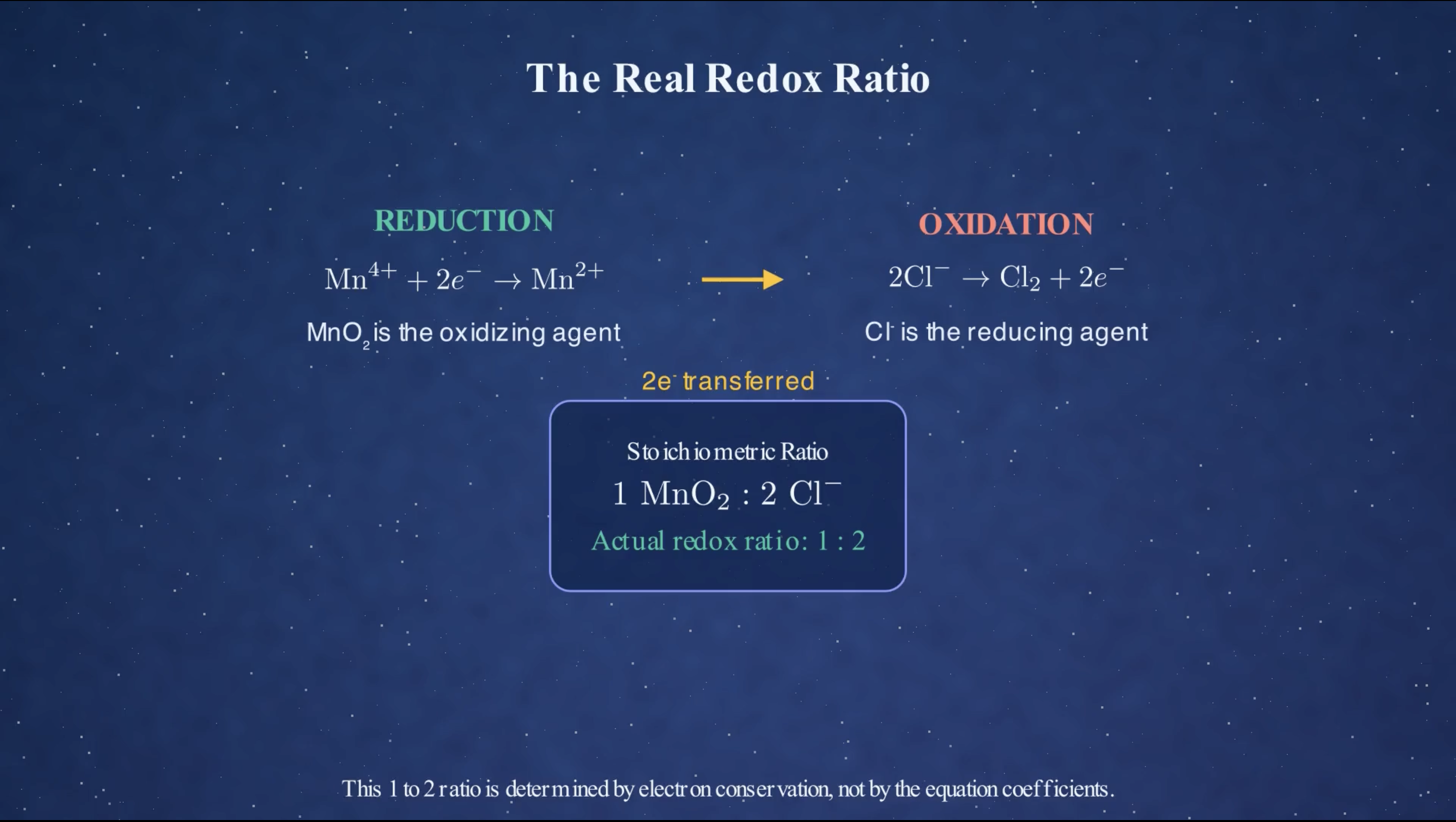}\hfill
\includegraphics[width=0.32\linewidth]{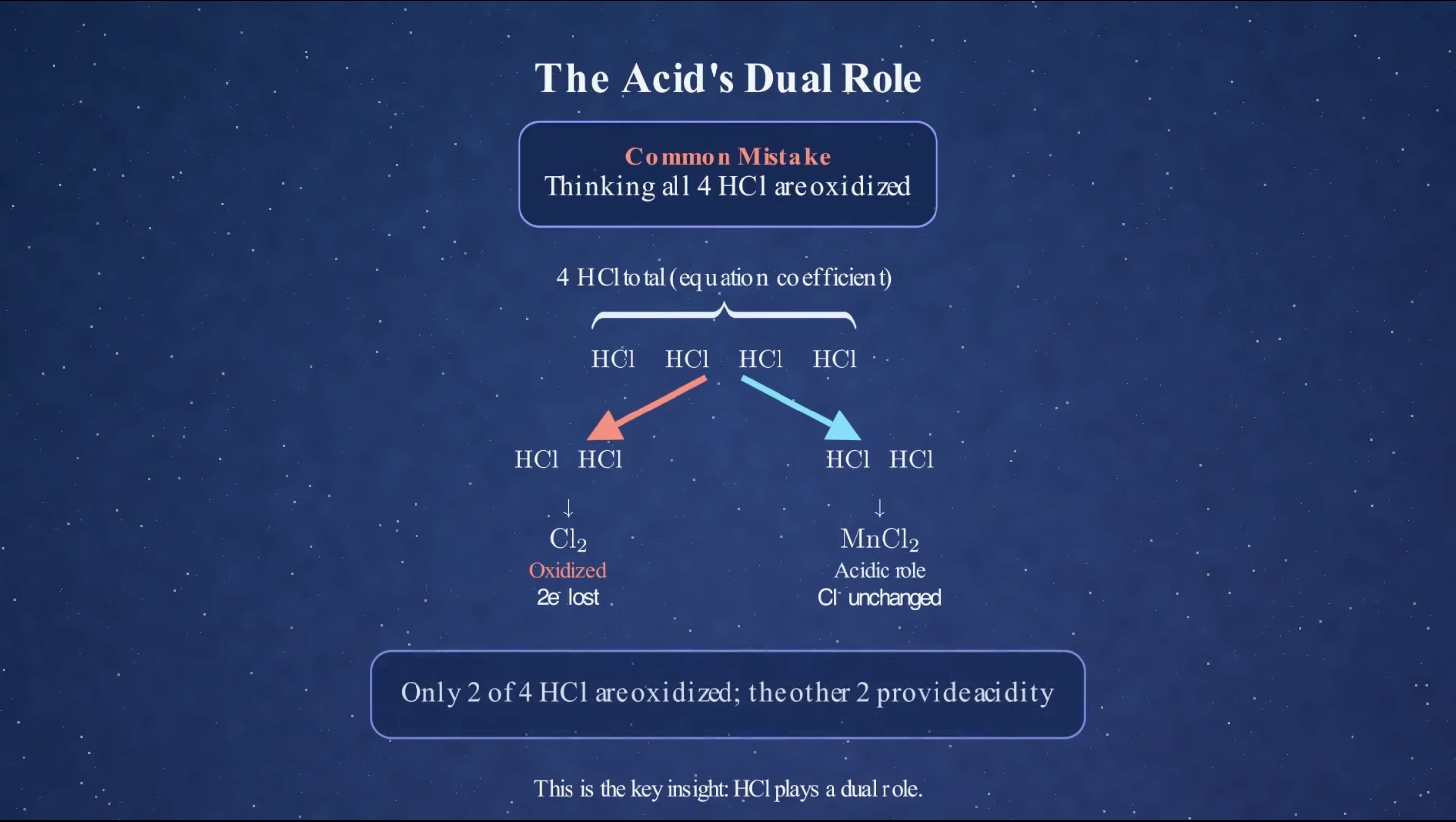}\hfill
\includegraphics[width=0.32\linewidth]{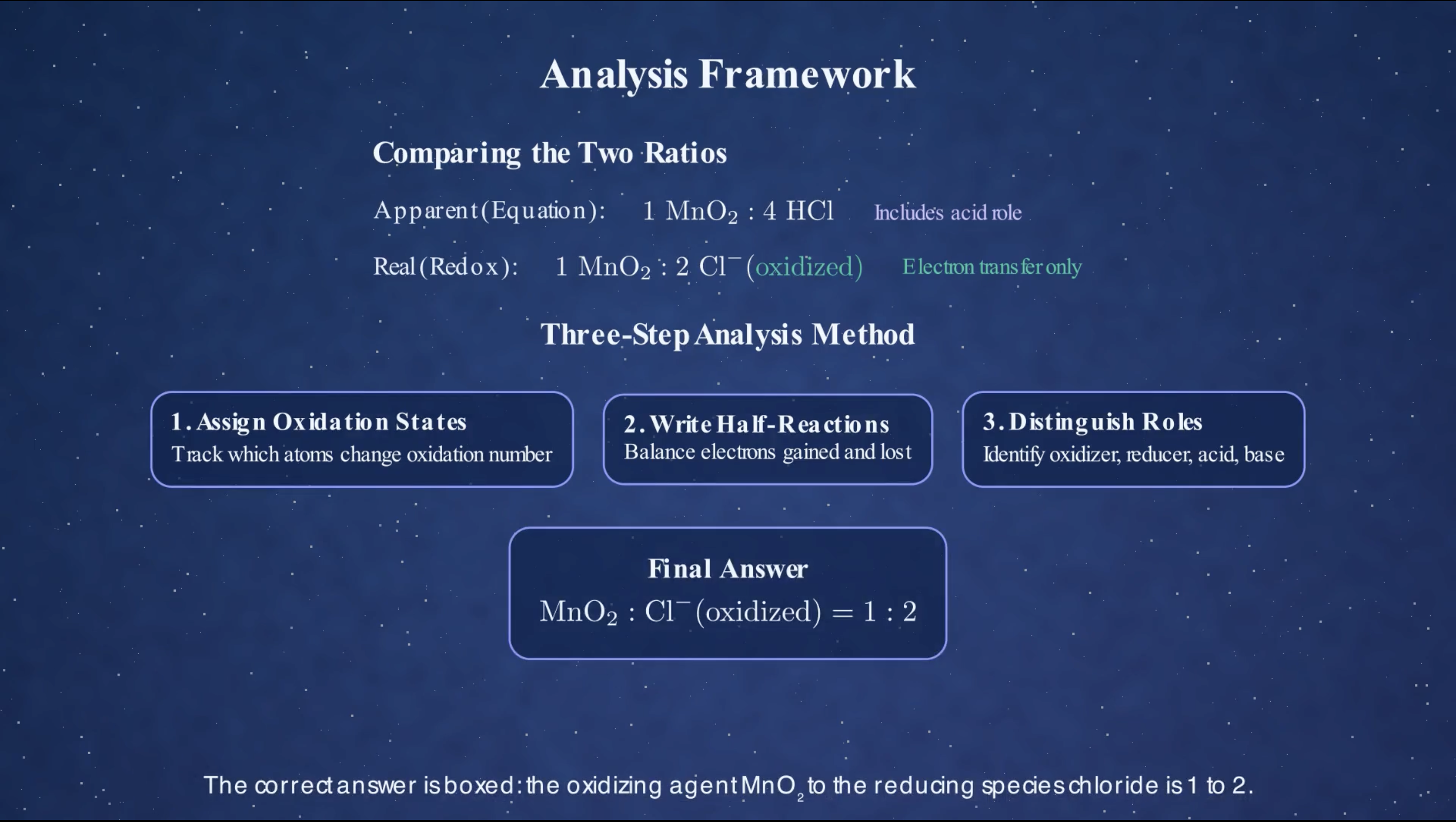}\\[0.35em]
\includegraphics[width=0.32\linewidth]{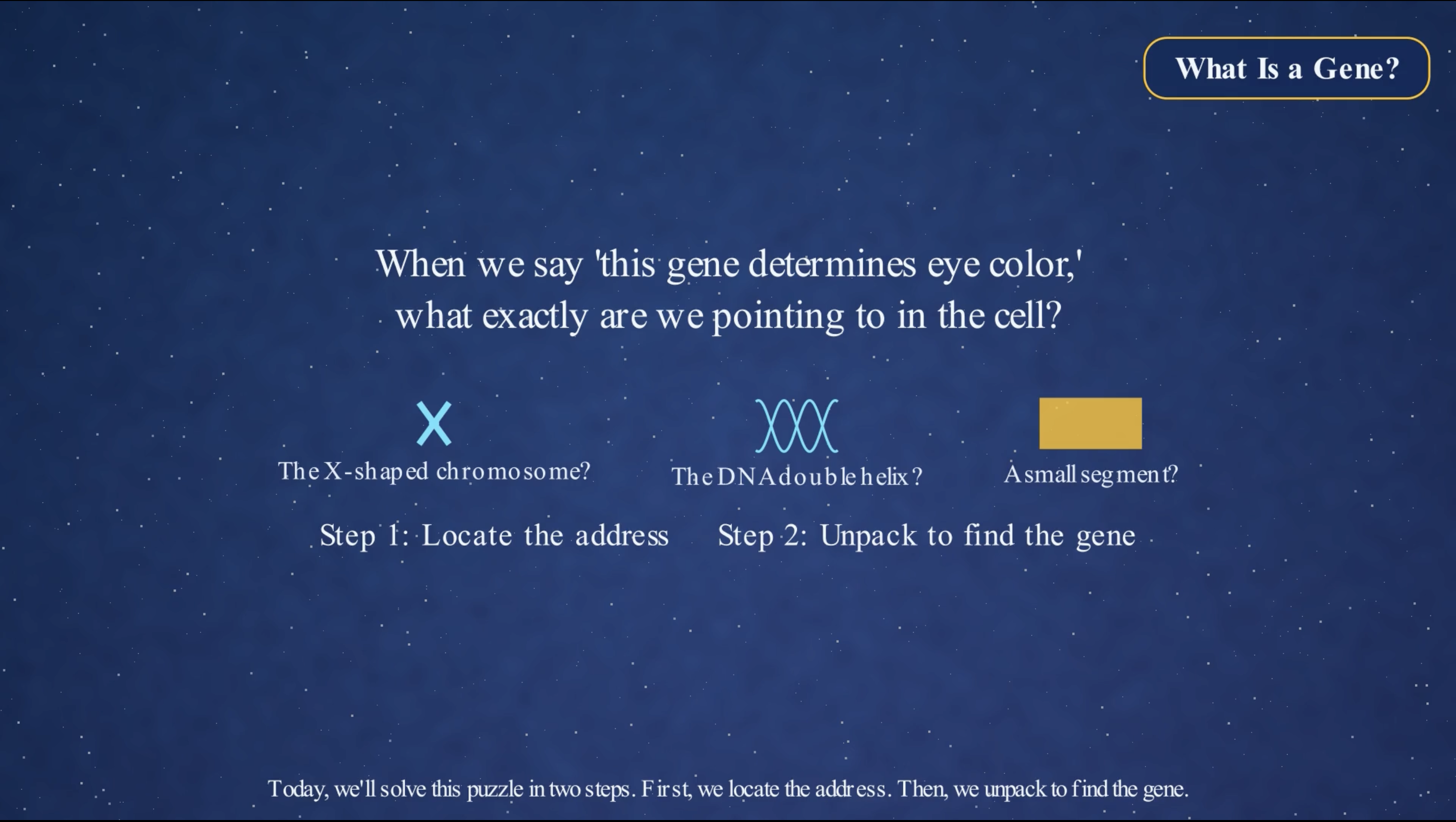}\hfill
\includegraphics[width=0.32\linewidth]{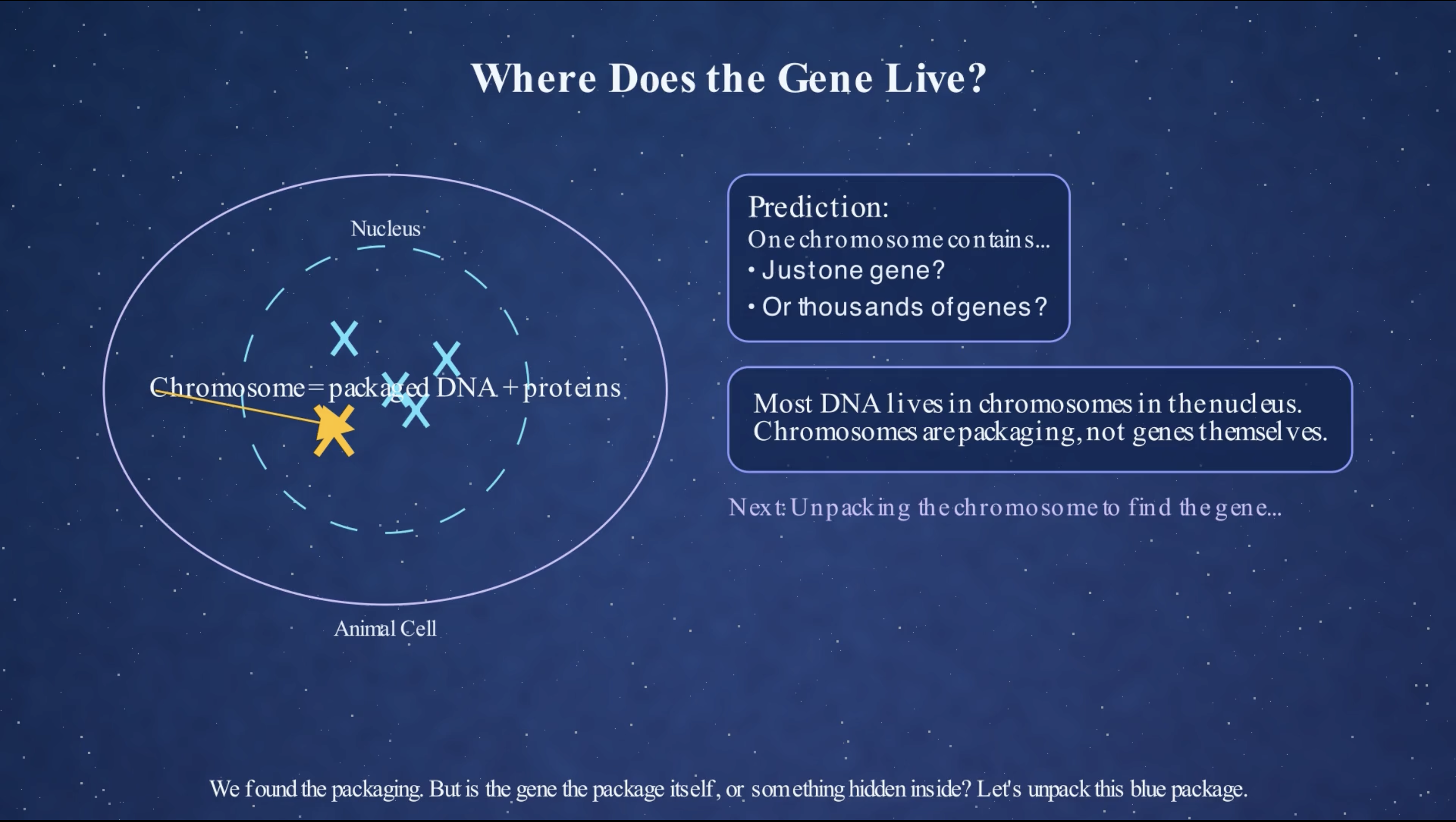}\hfill
\includegraphics[width=0.32\linewidth]{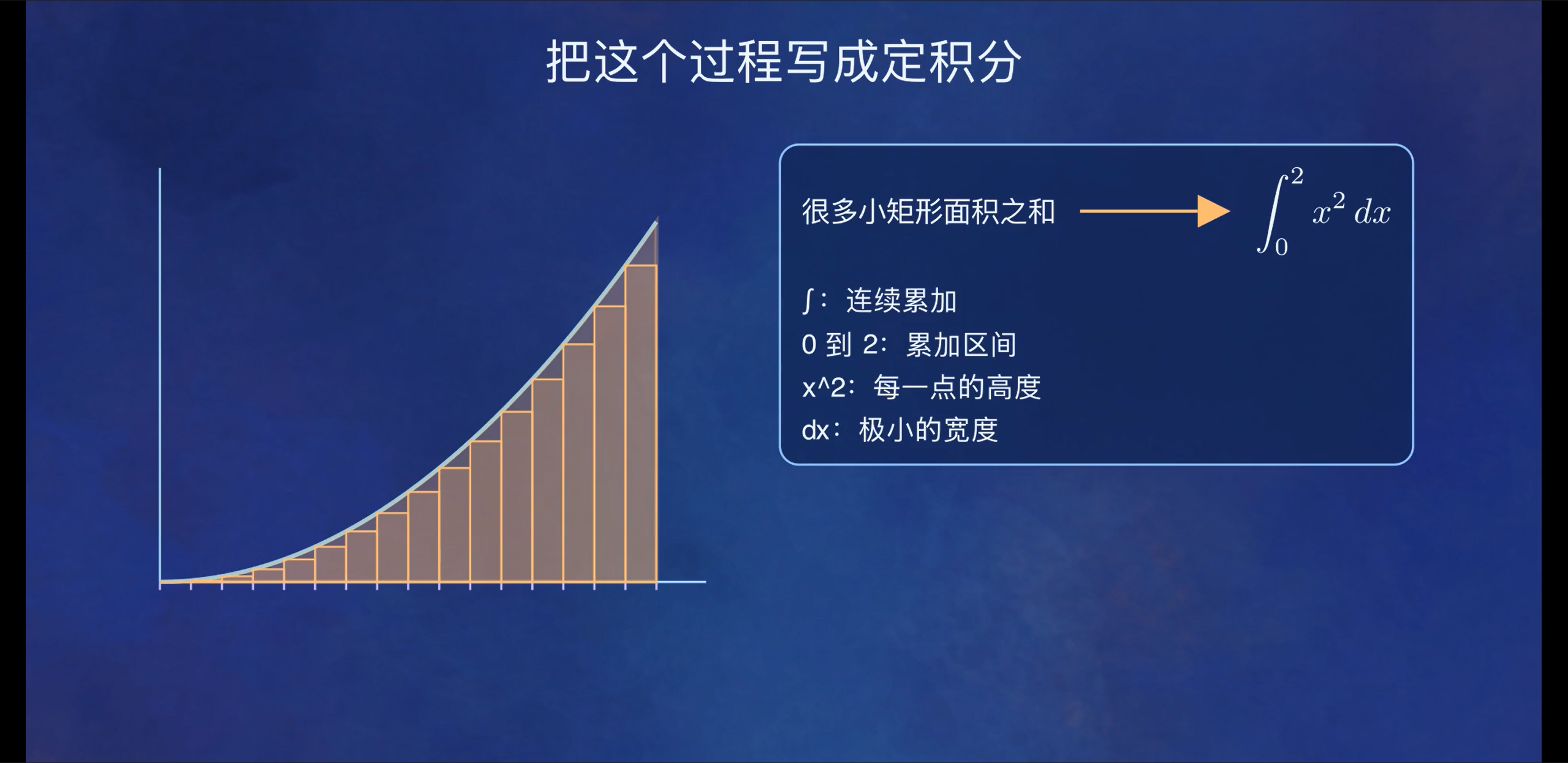}\\[0.35em]
\includegraphics[width=0.32\linewidth]{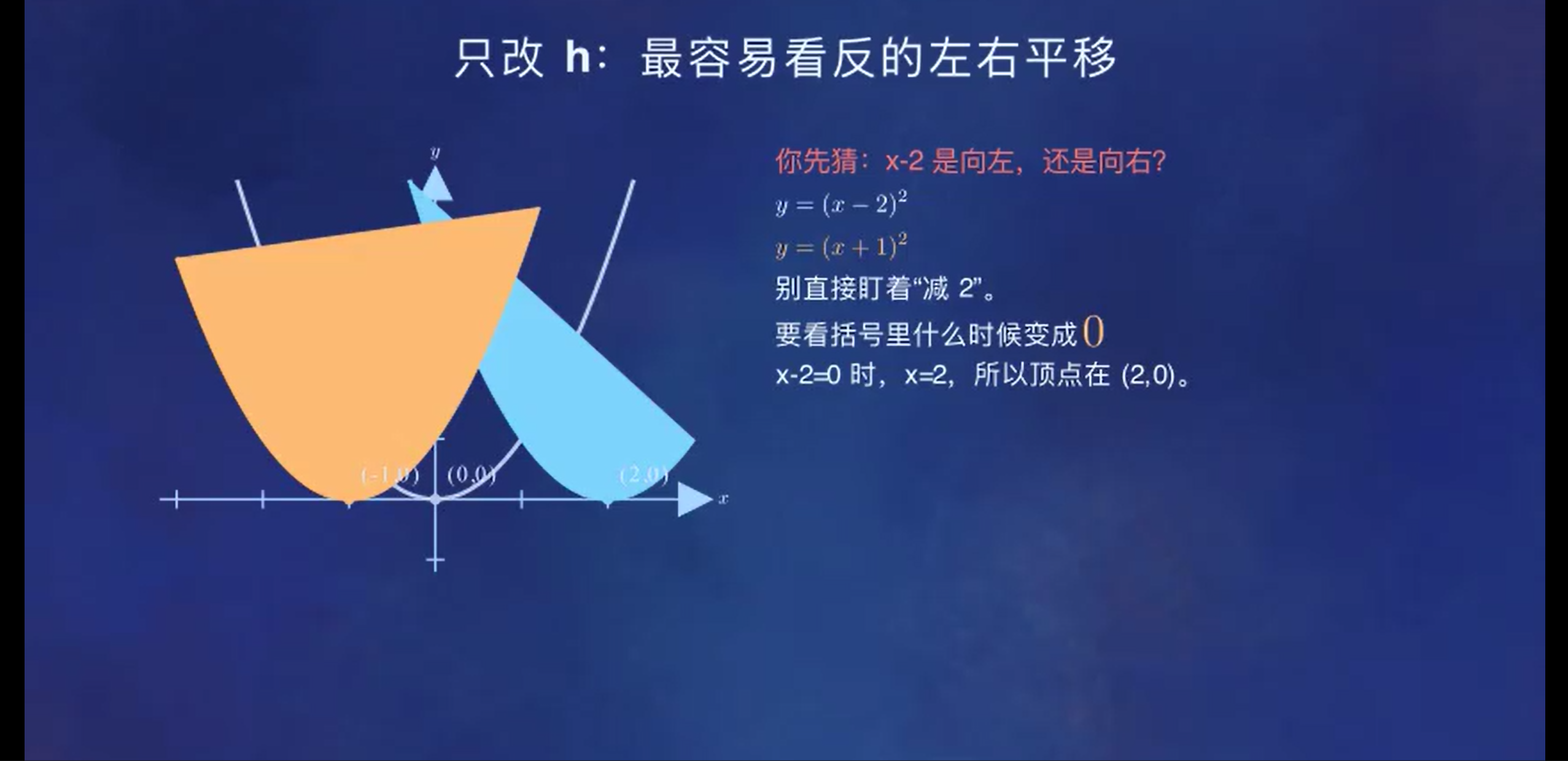}\hfill
\includegraphics[width=0.32\linewidth]{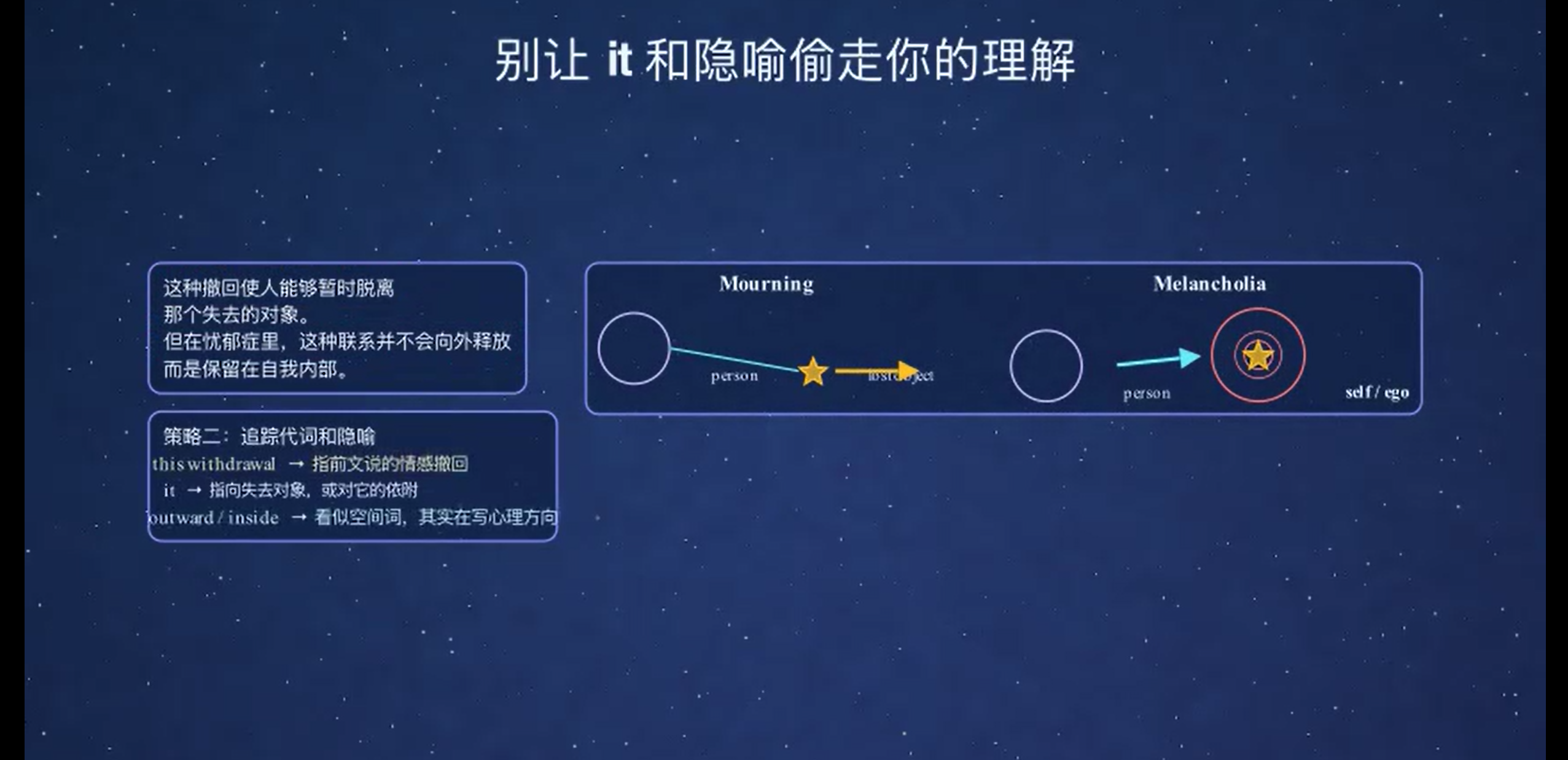}\hfill
\includegraphics[width=0.32\linewidth]{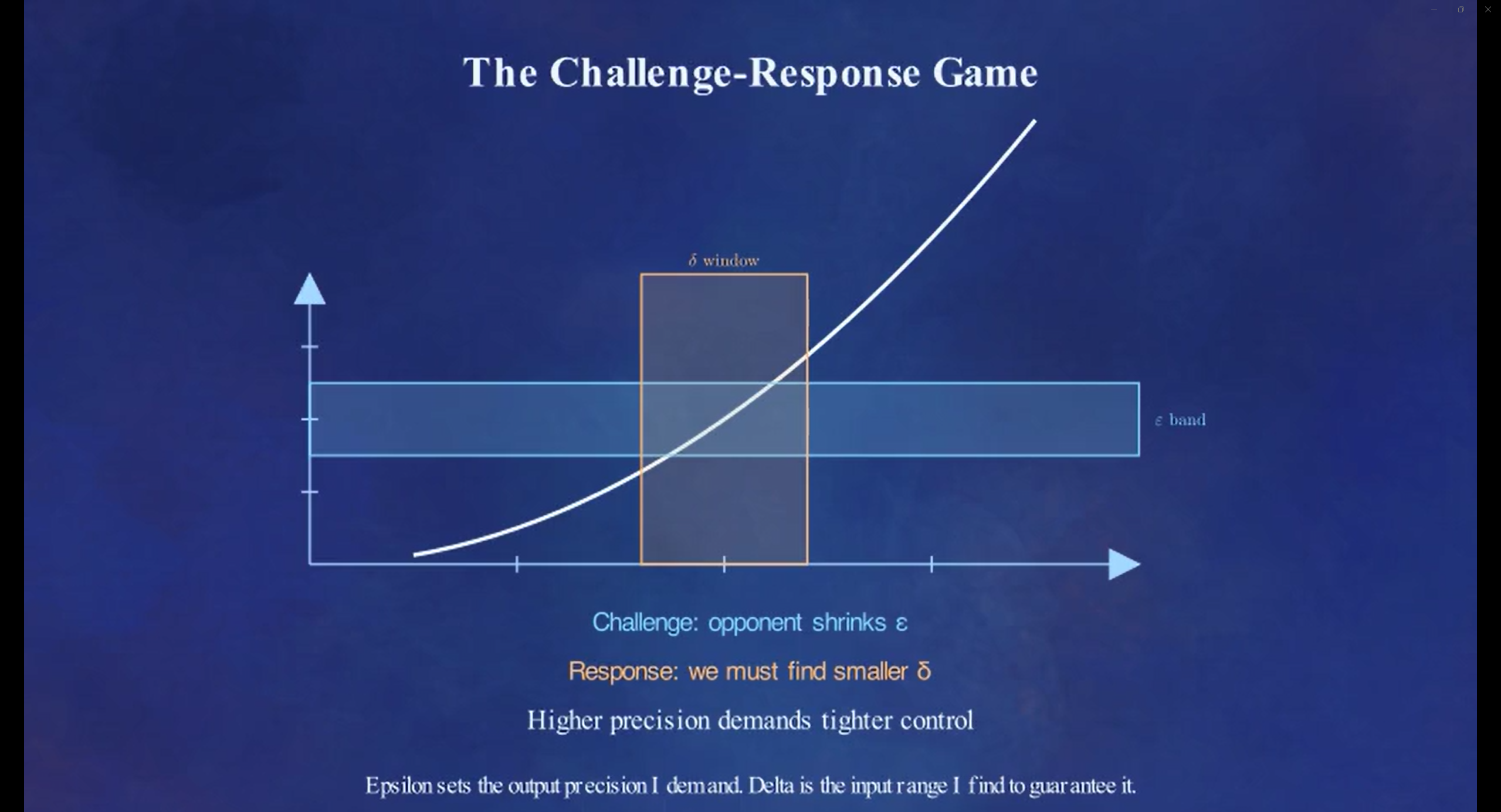}\\[0.35em]
\includegraphics[width=0.32\linewidth]{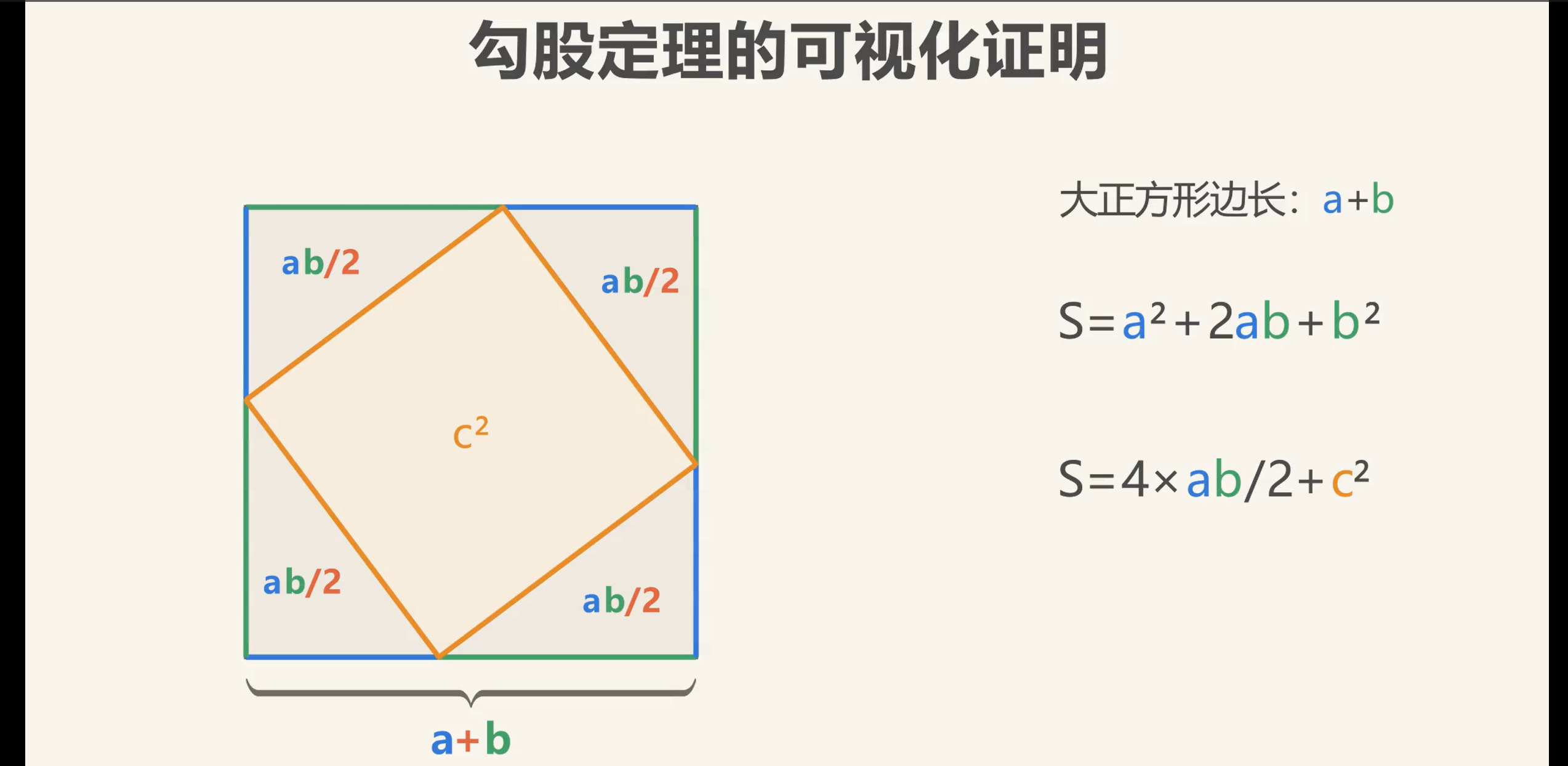}\hfill
\includegraphics[width=0.32\linewidth]{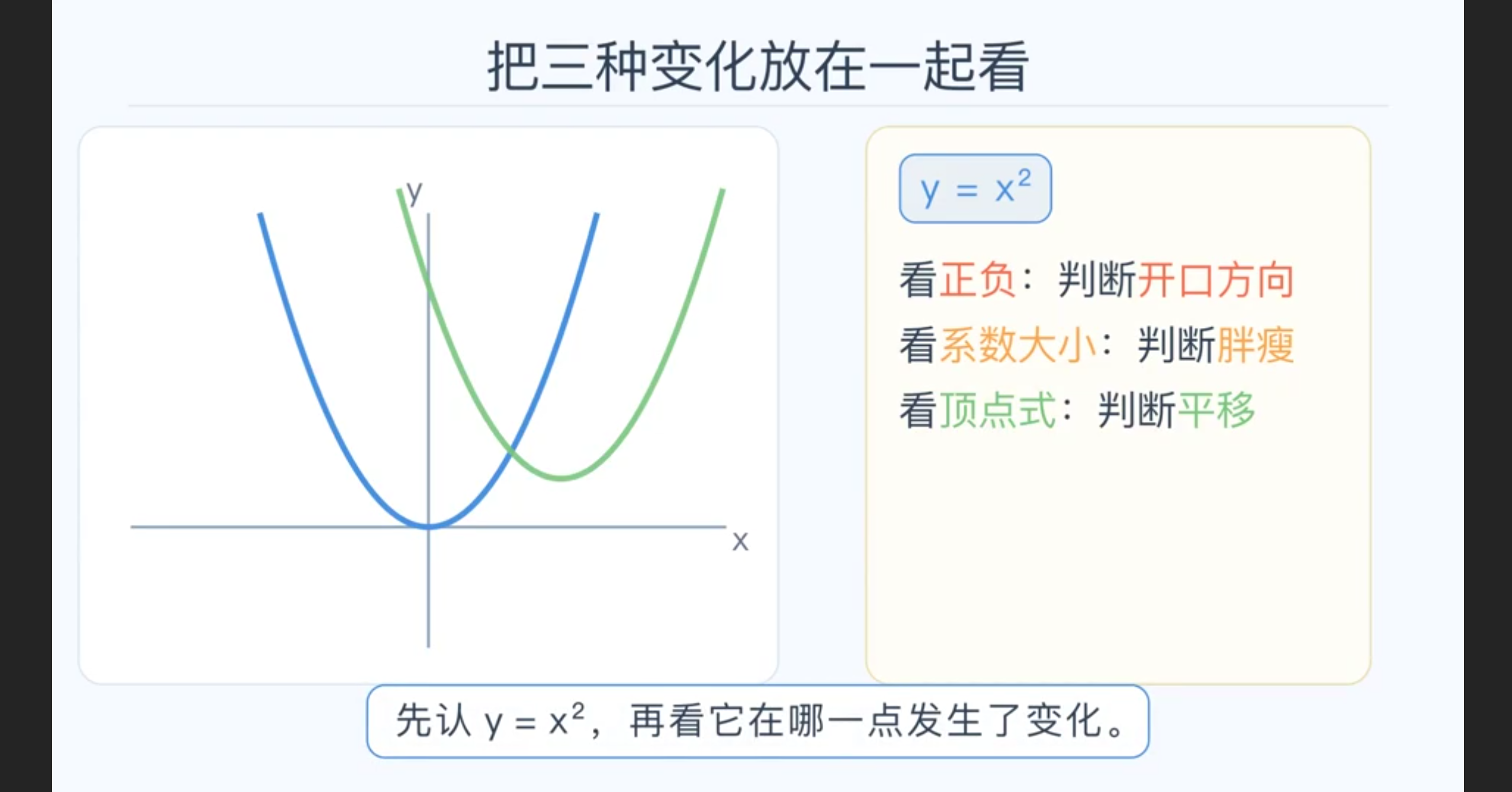}\hfill
\includegraphics[width=0.32\linewidth]{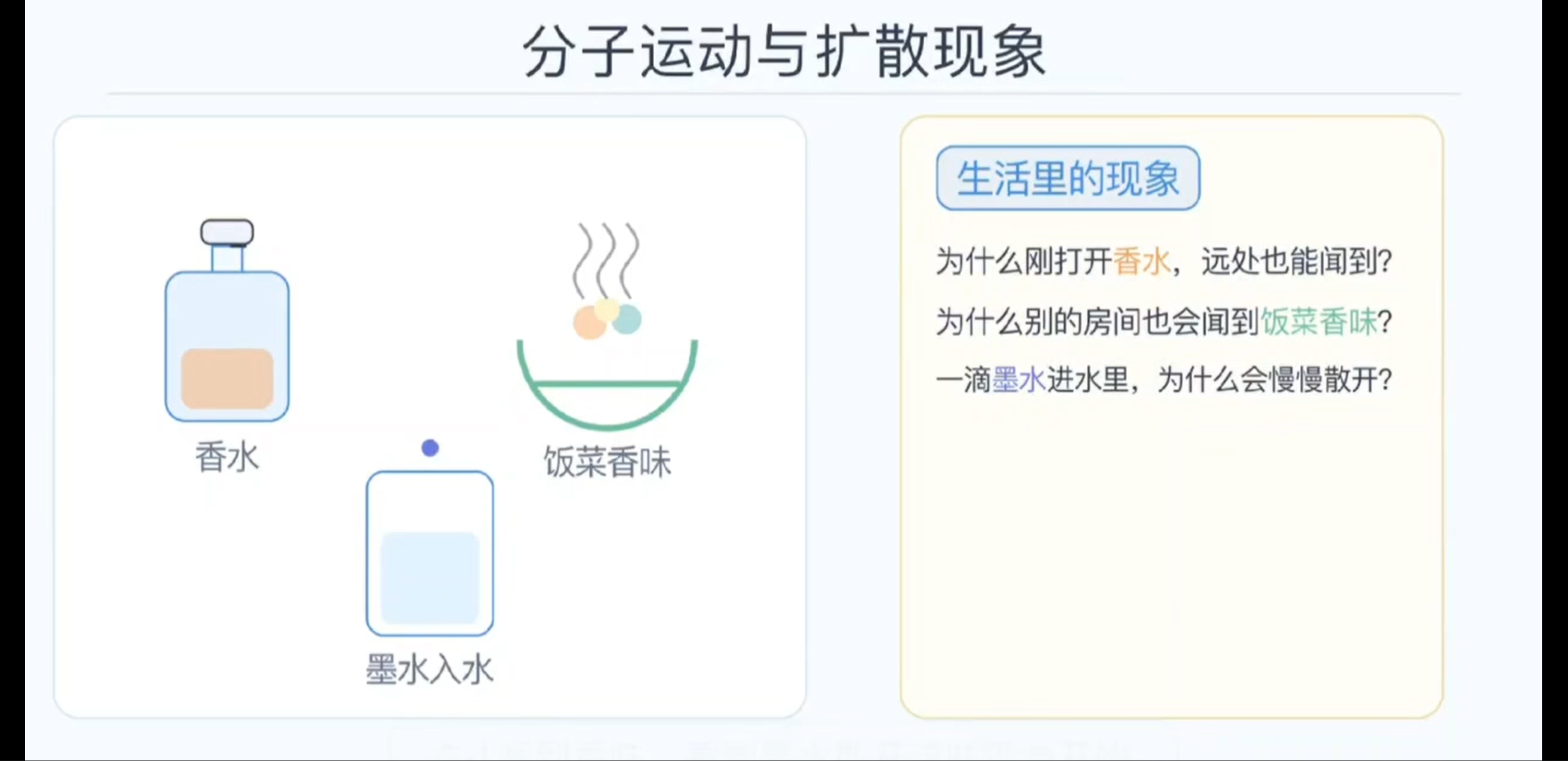}\\[0.35em]
\caption{Additional frames from animations generated by OmniManim with the SGCS/FLRPO-trained Qwen3-14B configuration, illustrating topic and visual diversity.}
\label{fig:materials}
\end{figure*}

\FloatBarrier

\onecolumn
\section{Prompts}
\label{app:prompts}
The fixed prompts are collected here. Each card is set in a single column. Fields in \texttt{\{\{braces\}\}} are filled at runtime.

\subsection{Scene Agent}
\label{app:scene_prompt}
{\small\itshape The Scene Agent receives the teaching requirement and, when present, the previous scene state. It returns JSON only. Relation predicates are restricted to \texttt{left\_of}, \texttt{above}, and \texttt{inside}. Exact coordinates are left to the Vision Agent.\par}
\vspace{3pt}
\begin{promptcard}
\noindent\textbf{Prompt.}
\par\vspace{0.28em}\noindent You are the Scene Agent of OmniManim, a framework for generating educational animations using Manim.
\par\vspace{0.28em}\noindent Your task is to convert a natural-language teaching requirement into a structured scene specification that can be used by a spatial layout planner and a code generation agent.
\par\vspace{0.28em}\noindent You must preserve the instructional meaning of the original requirement while explicitly identifying visual objects, spatial relations, pedagogical roles, and temporal dependencies.
\par\vspace{0.45em}\noindent\textbf{Follow these rules:}
\par\vspace{0.45em}\noindent\textbf{1. REQUIREMENT FAITHFULNESS}
\begin{promptlist}
\item Preserve all essential concepts, equations, facts, and instructional steps in the original requirement.
\item Do not invent additional facts or mathematical claims.
\item Do not omit necessary intermediate explanations.
\item If a requirement is ambiguous, record the ambiguity instead of inventing unsupported details.
\end{promptlist}
\par\vspace{0.45em}\noindent\textbf{2. SCENE DECOMPOSITION}
\begin{promptlist}
\item Divide the animation into logically coherent scenes only when necessary.
\item Each scene should have a clear instructional purpose.
\item Preserve the required order of explanations.
\end{promptlist}
\par\vspace{0.45em}\noindent\textbf{3. OBJECT EXTRACTION}
\begin{promptlist}
\item Identify every visual object needed to communicate the educational content.
\item Assign each object a unique and stable identifier.
\item Specify its type, content, and pedagogical role.
\item Reuse object identifiers across keyframes when the same object persists or transforms.
\end{promptlist}
\par\vspace{0.45em}\noindent\textbf{4. SPATIAL RELATIONS}
\begin{promptlist}
\item Represent explicit spatial relations using only: left\_of, above, inside.
\item Express inverse relationships by reversing object IDs.
\item Do not invent precise coordinates or bounding boxes.
\item Leave exact placement to the Vision Agent.
\end{promptlist}
\par\vspace{0.45em}\noindent\textbf{5. TEMPORAL PLANNING}
\begin{promptlist}
\item Describe when each object appears, persists, moves, or disappears.
\item Use the temporal states: new, keep, move, disappear.
\item Preserve dependencies between instructional steps.
\item Avoid unnecessary simultaneous display of unrelated content.
\end{promptlist}
\par\vspace{0.45em}\noindent\textbf{6. OUTPUT REQUIREMENTS}
\begin{promptlist}
\item Return valid JSON only.
\item Do not output Python code.
\item Do not include Markdown or explanatory text.
\item Use stable object and keyframe identifiers.
\item Include every required field in the schema.
\end{promptlist}
\par\vspace{0.55em}\noindent\textbf{User.}
\par\vspace{0.35em}\noindent\textbf{Teaching requirement:} \texttt{\{\{requirement\}\}}
\par\vspace{0.35em}\noindent\textbf{Previous scene state, if available:} \texttt{\{\{previous\_scene\_state\}\}}
\par\vspace{0.45em}\noindent\textbf{Produce a structured specification using this schema:}
\par\vspace{0.25em}{\ttfamily\small \hspace*{0.0em}\{\\\hspace*{1.0em}"scenes": [\\\hspace*{2.0em}\{\\\hspace*{3.0em}"scene\_id": "s01",\\\hspace*{3.0em}"scene\_text": "...",\\\hspace*{3.0em}"instructional\_goal": "...",\\\hspace*{3.0em}"required\_content": [\\\hspace*{4.0em}"..."\\\hspace*{3.0em}],\\\hspace*{3.0em}"objects": [\\\hspace*{4.0em}\{\\\hspace*{5.0em}"id": "o01",\\\hspace*{5.0em}"type": "...",\\\hspace*{5.0em}"content": "...",\\\hspace*{5.0em}"role": "...",\\\hspace*{5.0em}"size\_prior": null\\\hspace*{4.0em}\}\\\hspace*{3.0em}],\\\hspace*{3.0em}"relations": [\\\hspace*{4.0em}\{\\\hspace*{5.0em}"subject\_id": "o01",\\\hspace*{5.0em}"predicate": "above",\\\hspace*{5.0em}"object\_id": "o02",\\\hspace*{5.0em}"active\_keyframes": ["k01"]\\\hspace*{4.0em}\}\\\hspace*{3.0em}],\\\hspace*{3.0em}"keyframes": [\\\hspace*{4.0em}\{\\\hspace*{5.0em}"id": "k01",\\\hspace*{5.0em}"relative\_time": 0.0,\\\hspace*{5.0em}"object\_events": [\\\hspace*{6.0em}\{\\\hspace*{7.0em}"object\_id": "o01",\\\hspace*{7.0em}"status": "new"\\\hspace*{6.0em}\}\\\hspace*{5.0em}]\\\hspace*{4.0em}\}\\\hspace*{3.0em}],\\\hspace*{3.0em}"ambiguities": []\\\hspace*{2.0em}\}\\\hspace*{1.0em}]\\\hspace*{0.0em}\}\par}
\par\vspace{0.28em}\noindent Use only valid object types, pedagogical roles, relation predicates, and temporal states.
\par\vspace{0.28em}\noindent Relative keyframe times must be ordered within [0, 1].
\par\vspace{0.28em}\noindent Return JSON only.
\end{promptcard}

\subsection{Code Agent}
\label{app:code_prompt}
{\small\itshape The Code Agent receives the teaching requirement, the structured scene specification, the predicted keyframe layouts, and the target scene identifier. It returns Python only. A normalized box \((x_c,y_c,w_b,h_b)\) maps to \(X=(x_c-0.5)\times 14.22\) and \(Y=(0.5-y_c)\times 8.0\). Objects keep their aspect ratio inside the assigned box. The target runtime is Python~3.12 and Manim Community~v0.19.0.\par}
\vspace{3pt}
\begin{promptcard}
\noindent\textbf{Prompt.}
\par\vspace{0.28em}\noindent You are the Code Agent of OmniManim.
\par\vspace{0.28em}\noindent Your task is to generate executable Manim Community v0.19.0 Python code from a structured teaching plan and a sequence of predicted keyframe layouts.
\par\vspace{0.28em}\noindent The generated animation must faithfully implement the instructional content while respecting the provided spatial and temporal constraints.
\par\vspace{0.28em}\noindent The visual layout has already been planned. You must implement that plan rather than freely redesigning object positions.
\par\vspace{0.45em}\noindent\textbf{Follow these requirements:}
\par\vspace{0.45em}\noindent\textbf{1. CONTENT CORRECTNESS}
\begin{promptlist}
\item Preserve every required instructional concept, equation, and explanation.
\item Ensure mathematical expressions are correct.
\item Use appropriate Manim objects such as Text, MathTex, VGroup, Axes, and geometric primitives.
\item Do not invent additional mathematical claims.
\end{promptlist}
\par\vspace{0.45em}\noindent\textbf{2. LAYOUT COMPLIANCE}
\begin{promptlist}
\item Treat predicted bounding boxes as spatial constraints for the corresponding keyframes.
\item Each bounding box is normalized as:
\end{promptlist}
\par\vspace{0.2em}{\ttfamily\small (center\_x, center\_y, width, height).\par}
\begin{promptlist}
\item Convert normalized positions into Manim coordinates:
\end{promptlist}
\par\vspace{0.2em}{\ttfamily\small X = (center\_x - 0.5) * 14.22\\Y = (0.5 - center\_y) * 8.0\par}
\begin{promptlist}
\item Preserve aspect ratios when scaling objects.
\item Fit objects within their assigned bounding boxes.
\item Avoid unintended overlaps, viewport overflow, and visual clutter.
\item Preserve intentional geometric intersections and explicitly specified containment.
\end{promptlist}
\par\vspace{0.45em}\noindent\textbf{3. TEMPORAL CONSISTENCY}
\begin{promptlist}
\item Respect the order of instructional events.
\item Match object positions at the specified keyframes.
\item Preserve object identities across transformations.
\item Use appropriate animations for movement and transformation.
\item Remove objects when their scheduled lifecycle ends.
\item Avoid leaving unintended content from prior steps.
\end{promptlist}
\par\vspace{0.45em}\noindent\textbf{4. OBJECT TRACEABILITY}
\begin{promptlist}
\item Use stable Python variable names corresponding to object identifiers in the scene specification.
\item Keep object creation, transformation, and removal operations traceable to their associated objects.
\item Organize the code into clear, localized blocks so that later repairs can target specific objects and animation intervals.
\end{promptlist}
\par\vspace{0.45em}\noindent\textbf{5. EXECUTION REQUIREMENTS}
\begin{promptlist}
\item Generate valid Python 3.12 code compatible with Manim Community v0.19.0.
\item Include all required imports and scene definitions.
\item Do not require unavailable external assets.
\item Do not access the network or external services.
\item Ensure the scene can be rendered independently.
\item Do not suppress compilation or rendering errors.
\end{promptlist}
\par\vspace{0.45em}\noindent\textbf{6. OUTPUT FORMAT}
\begin{promptlist}
\item Return executable Python source code only.
\item Do not include Markdown code fences.
\item Do not include explanations outside the code.
\item Do not output incomplete functions or placeholders.
\end{promptlist}
\par\vspace{0.55em}\noindent\textbf{User.}
\par\vspace{0.35em}\noindent\textbf{Original teaching requirement:} \texttt{\{\{requirement\}\}}
\par\vspace{0.35em}\noindent\textbf{Structured scene specification:} \texttt{\{\{scene\_state\_json\}\}}
\par\vspace{0.35em}\noindent\textbf{Predicted keyframe layouts:} \texttt{\{\{keyframe\_layout\_json\}\}}
\par\vspace{0.35em}\noindent\textbf{Target scene identifier:} \texttt{\{\{scene\_id\}\}}
\par\vspace{0.28em}\noindent Generate a complete executable Manim scene program.
\par\vspace{0.45em}\noindent\textbf{The program must:}
\begin{promptlist}
\item 1. Implement all required content.
\item 2. Respect the predicted keyframe boxes.
\item 3. Follow the specified temporal schedule.
\item 4. Preserve stable object identities.
\item 5. Avoid unintended spatial and temporal violations.
\end{promptlist}
\par\vspace{0.28em}\noindent Return Python code only.
\end{promptcard}

\subsection{Repair Agent}
\label{app:repair_prompt}
{\small\itshape The Repair Agent receives the original program, the teaching requirement, the frozen keyframe layouts, a localized diagnostic, any rendered failure frames, and the allowed AST regions. It returns one JSON patch and does not regenerate the program. When no safe local edit exists, the status is \texttt{NO\_SAFE\_PATCH}.\par}
\vspace{3pt}
\begin{promptcard}
\noindent\textbf{Prompt.}
\par\vspace{0.28em}\noindent You are the Repair Agent of OmniManim.
\par\vspace{0.28em}\noindent Your task is to repair a specific compilation or visual failure in an existing Manim program.
\par\vspace{0.28em}\noindent You are given the original program, the teaching requirement, the frozen visual plan, and a localized failure diagnostic.
\par\vspace{0.28em}\noindent Your objective is to correct the identified failure using the smallest valid code modification while preserving all unrelated content and constraints.
\par\vspace{0.28em}\noindent You must perform FAILURE-LOCALIZED REPAIR, not unrestricted program regeneration.
\par\vspace{0.45em}\noindent\textbf{Follow these rules:}
\par\vspace{0.45em}\noindent\textbf{1. TARGETED CORRECTION}
\begin{promptlist}
\item Focus on the failure specified in the diagnostic.
\item Identify the objects, time interval, and AST regions associated with that failure.
\item Modify only the relevant code and its necessary dependencies.
\item Do not rewrite the entire program.
\end{promptlist}
\par\vspace{0.45em}\noindent\textbf{2. CONTENT PRESERVATION}
\begin{promptlist}
\item Preserve all required instructional content.
\item Do not delete required equations, labels, or objects.
\item Do not change the mathematical meaning of expressions.
\item Preserve the original instructional order.
\item Do not remove content merely to avoid overlap.
\end{promptlist}
\par\vspace{0.45em}\noindent\textbf{3. SPATIAL PRESERVATION}
\begin{promptlist}
\item Respect the frozen keyframe layouts.
\item Correct unintended overlaps, viewport violations, scaling errors, and spatial-relation violations.
\item Preserve intentional geometric intersections.
\item Do not introduce new spatial violations elsewhere.
\end{promptlist}
\par\vspace{0.45em}\noindent\textbf{4. TEMPORAL PRESERVATION}
\begin{promptlist}
\item Preserve object identities and lifecycle requirements.
\item Do not move a failure outside the inspected interval merely to evade detection.
\item Do not hide required objects through opacity, off-screen positioning, or premature removal.
\item Do not introduce new timing or lifecycle errors.
\end{promptlist}
\par\vspace{0.45em}\noindent\textbf{5. MINIMAL EDITING}
\begin{promptlist}
\item Prefer changing a small number of statements.
\item Avoid modifying unrelated AST nodes.
\item Preserve valid imports, functions, and scene definitions.
\item Maintain compatibility with Manim Community v0.19.0.
\end{promptlist}
\par\vspace{0.45em}\noindent\textbf{6. COMPILATION FAILURES}
\begin{promptlist}
\item If rendering failed, use the compiler traceback and associated AST regions as the primary evidence.
\item Do not assume visual evidence exists when rendering was unsuccessful.
\end{promptlist}
\par\vspace{0.45em}\noindent\textbf{7. SAFETY AND VALIDITY}
\begin{promptlist}
\item Produce syntactically valid Python modifications.
\item Do not suppress errors or bypass verification.
\item Do not change the evaluator or validation procedure.
\item If no safe localized repair is possible, report that a valid patch cannot be determined.
\end{promptlist}
\par\vspace{0.45em}\noindent\textbf{8. OUTPUT}
\begin{promptlist}
\item Return one structured patch candidate.
\item Do not return a completely regenerated program.
\item Do not include Markdown or additional commentary.
\end{promptlist}
\par\vspace{0.55em}\noindent\textbf{User.}
\par\vspace{0.35em}\noindent\textbf{Original teaching requirement:} \texttt{\{\{requirement\}\}}
\par\vspace{0.35em}\noindent\textbf{Structured scene state:} \texttt{\{\{scene\_state\_json\}\}}
\par\vspace{0.35em}\noindent\textbf{Frozen keyframe layouts:} \texttt{\{\{keyframe\_layout\_json\}\}}
\par\vspace{0.35em}\noindent\textbf{Original Manim program:} \texttt{\{\{original\_code\}\}}
\par\vspace{0.35em}\noindent\textbf{Failure diagnostic:} \texttt{\{\{diagnostic\_json\}\}}
\par\vspace{0.35em}\noindent\textbf{Relevant rendered frames, if available:} \texttt{\{\{failure\_frames\}\}}
\par\vspace{0.35em}\noindent\textbf{Allowed AST regions:} \texttt{\{\{allowed\_ast\_regions\}\}}
\par\vspace{0.28em}\noindent Generate one minimal patch that corrects the target failure while preserving all unrelated constraints.
\par\vspace{0.45em}\noindent\textbf{Return JSON using this schema:}
\par\vspace{0.25em}{\ttfamily\small \hspace*{0.0em}\{\\\hspace*{1.0em}"status": "PATCH",\\\hspace*{1.0em}"target\_failure\_id": "...",\\\hspace*{1.0em}"affected\_object\_ids": ["..."],\\\hspace*{1.0em}"edits": [\\\hspace*{2.0em}\{\\\hspace*{3.0em}"ast\_node\_id": "...",\\\hspace*{3.0em}"replacement\_code": "..."\\\hspace*{2.0em}\}\\\hspace*{1.0em}],\\\hspace*{1.0em}"reason": "Brief description of the correction"\\\hspace*{0.0em}\}\par}
\par\vspace{0.28em}\noindent If no valid localized repair can be determined, return:
\par\vspace{0.25em}{\ttfamily\small \hspace*{0.0em}\{\\\hspace*{1.0em}"status": "NO\_SAFE\_PATCH",\\\hspace*{1.0em}"target\_failure\_id": "...",\\\hspace*{1.0em}"affected\_object\_ids": [],\\\hspace*{1.0em}"edits": [],\\\hspace*{1.0em}"reason": "..."\\\hspace*{0.0em}\}\par}
\par\vspace{0.28em}\noindent Return JSON only.
\end{promptcard}

\subsection{Deployment Review}
\label{app:verify_prompt}
{\small\itshape Claude Opus 4.6 applies the following prompt with deterministic decoding. \texttt{FAIL} triggers one repair attempt. \texttt{PASS} and \texttt{INTENTIONAL} preserve the output.\par}
\vspace{3pt}
\begin{promptcard}
\noindent\textbf{Prompt.}
\par\vspace{0.28em}\noindent You are a visual verification agent for educational animations generated with Manim.
\par\vspace{0.28em}\noindent Your task is to determine whether a suspicious animation segment contains a genuine visual defect that requires code repair.
\par\vspace{0.45em}\noindent\textbf{You are given:}
\begin{promptlist}
\item The original teaching requirement.
\item The structured scene specification.
\item Three rendered keyframes from the suspicious interval: start, middle, and end.
\item Computer-vision diagnostic measurements.
\item The object identities associated with the warning.
\end{promptlist}
\par\vspace{0.28em}\noindent You must distinguish genuine rendering failures from intentional visual relationships.
\par\vspace{0.45em}\noindent\textbf{CLASSIFICATION LABELS:}
\par\vspace{0.45em}\noindent\textbf{PASS:}
\par\vspace{0.28em}\noindent The suspected issue is not a meaningful visual failure. The rendering is acceptable.
\par\vspace{0.45em}\noindent\textbf{FAIL:}
\par\vspace{0.28em}\noindent A genuine visual defect is clearly observable and should be corrected.
\par\vspace{0.45em}\noindent\textbf{INTENTIONAL:}
\par\vspace{0.28em}\noindent The apparent overlap or spatial interaction is deliberate and semantically appropriate for the educational content.
\par\vspace{0.45em}\noindent\textbf{REVIEW CRITERIA:}
\par\vspace{0.45em}\noindent\textbf{1. VIEWPORT BOUNDARIES}
\begin{promptlist}
\item Identify important text, equations, or objects extending beyond the visible frame.
\item Ignore harmless decorative effects.
\end{promptlist}
\par\vspace{0.45em}\noindent\textbf{2. OBJECT OVERLAP}
\begin{promptlist}
\item Identify unintended overlaps that obscure essential content or reduce readability.
\item Do not penalize meaningful geometric intersections or deliberate layering.
\end{promptlist}
\par\vspace{0.45em}\noindent\textbf{3. SCALE AND SPATIAL RELATIONS}
\begin{promptlist}
\item Check whether object sizes and placements create clear visual problems.
\item Identify violations of required spatial relationships.
\end{promptlist}
\par\vspace{0.45em}\noindent\textbf{4. TEMPORAL CONSISTENCY}
\begin{promptlist}
\item Compare the three supplied keyframes.
\item Identify visible remnants of objects that should have disappeared.
\item Identify obvious object-state inconsistencies.
\item Do not infer motion defects that cannot be established from the provided evidence.
\end{promptlist}
\par\vspace{0.45em}\noindent\textbf{5. SEMANTIC INTENT}
\begin{promptlist}
\item Interpret the frames in the context of the original teaching requirement.
\item An intersection that is necessary for a mathematical diagram is not automatically a visual defect.
\item A diagnostic threshold violation alone does not establish that repair is necessary.
\end{promptlist}
\par\vspace{0.45em}\noindent\textbf{DECISION RULES:}
\begin{promptlist}
\item Return FAIL only when a genuine defect is supported by visible evidence.
\item Return INTENTIONAL when the suspicious geometry is semantically justified.
\item Otherwise, return PASS.
\item Do not use aesthetic preference alone to trigger a repair.
\item Report the specific objects and time interval involved in any confirmed failure.
\end{promptlist}
\par\vspace{0.45em}\noindent\textbf{SEVERITY SCALE:}
\par\vspace{0.2em}{\ttfamily\small 0 = No meaningful defect\\1 = Minor defect\\2 = Moderate defect affecting presentation\\3 = Severe defect affecting required content\par}
\par\vspace{0.45em}\noindent\textbf{OUTPUT REQUIREMENTS:}
\par\vspace{0.28em}\noindent Return valid JSON only. Do not include Markdown or additional commentary.
\par\vspace{0.55em}\noindent\textbf{User.}
\par\vspace{0.35em}\noindent\textbf{Original teaching requirement:} \texttt{\{\{requirement\}\}}
\par\vspace{0.35em}\noindent\textbf{Structured scene specification:} \texttt{\{\{scene\_state\_json\}\}}
\par\vspace{0.35em}\noindent\textbf{Suspicious time interval:} \texttt{\{\{time\_interval\}\}}
\par\vspace{0.35em}\noindent\textbf{Affected object identifiers:} \texttt{\{\{affected\_object\_ids\}\}}
\par\vspace{0.35em}\noindent\textbf{Computer-vision diagnostics:} \texttt{\{\{cv\_diagnostics\_json\}\}}
\par\vspace{0.35em}\noindent\textbf{Start frame:} \texttt{\{\{start\_frame\}\}}
\par\vspace{0.35em}\noindent\textbf{Middle frame:} \texttt{\{\{middle\_frame\}\}}
\par\vspace{0.35em}\noindent\textbf{End frame:} \texttt{\{\{end\_frame\}\}}
\par\vspace{0.28em}\noindent Classify the suspicious segment.
\par\vspace{0.45em}\noindent\textbf{Return JSON:}
\par\vspace{0.25em}{\ttfamily\small \hspace*{0.0em}\{\\\hspace*{1.0em}"verdict": "PASS",\\\hspace*{1.0em}"confidence": 0.0,\\\hspace*{1.0em}"severity": 0,\\\hspace*{1.0em}"failure\_type": "none",\\\hspace*{1.0em}"affected\_object\_ids": [],\\\hspace*{1.0em}"evidence": "...",\\\hspace*{1.0em}"suggested\_constraint": "..."\\\hspace*{0.0em}\}\par}
\par\vspace{0.45em}\noindent\textbf{The verdict must be exactly one of:}
\par\vspace{0.28em}\noindent PASS, FAIL, INTENTIONAL.
\par\vspace{0.28em}\noindent Confidence must be between 0 and 1.
\par\vspace{0.45em}\noindent\textbf{Failure type must be one of:}
\par\vspace{0.28em}\noindent none, overlap, boundary, scale, spatial\_relation, temporal\_remnant, other\_visual\_error.
\par\vspace{0.28em}\noindent For PASS or INTENTIONAL, explain briefly why code repair is unnecessary.
\par\vspace{0.28em}\noindent Return JSON only.
\end{promptcard}

\subsection{Instructional Score}
\label{app:instruction_prompt}
{\small\itshape This offline prompt is separate from deployment review. Claude Opus 4.6 applies it with deterministic decoding. The returned scores are not used for training rewards or deployment stopping.\par}
\vspace{3pt}
\begin{promptcard}
\noindent\textbf{Prompt.}
\par\vspace{0.28em}\noindent You are an independent evaluator of educational animations.
\par\vspace{0.28em}\noindent Your task is to assess how effectively a rendered educational animation satisfies its original teaching requirement.
\par\vspace{0.28em}\noindent You must evaluate the presented animation objectively and independently of the system that generated it.
\par\vspace{0.45em}\noindent\textbf{You will receive:}
\begin{promptlist}
\item The original teaching requirement.
\item A structured teaching plan.
\item Uniformly sampled, timestamped frames from the rendered animation.
\end{promptlist}
\par\vspace{0.45em}\noindent\textbf{Evaluate three dimensions:}
\par\vspace{0.45em}\noindent\textbf{1. CONTENT ACCURACY (CA)}
\par\vspace{0.45em}\noindent\textbf{2. PEDAGOGICAL CLARITY (PC)}
\par\vspace{0.45em}\noindent\textbf{3. ENGAGEMENT (EN)}
\par\vspace{0.45em}\noindent\textbf{IMPORTANT EVALUATION RULES:}
\begin{promptlist}
\item The original teaching requirement is the primary reference.
\item Do not assume that the teaching plan is complete or correct.
\item Judge only the content supported by the supplied visual evidence.
\item Do not infer correctness from the identity or reputation of a generating model.
\item Do not reward verbosity or decorative complexity by itself.
\item Do not use information about compilation, repair history, or internal model behavior.
\item Apply the same criteria to every video.
\item Do not use previously assigned scores.
\end{promptlist}
\par\vspace{0.45em}\noindent\textbf{DIMENSION 1: CONTENT ACCURACY (CA)}
\par\vspace{0.28em}\noindent CA is a binary score in \{0, 1\}.
\par\vspace{0.28em}\noindent Evaluate whether the animation accurately communicates the essential knowledge required by the original teaching requirement.
\par\vspace{0.45em}\noindent\textbf{Check:}
\begin{promptlist}
\item Mathematical and scientific correctness.
\item Correctness of displayed equations.
\item Logical consistency of explanations.
\item Presence of essential concepts and steps.
\item Consistency between textual and visual claims.
\end{promptlist}
\par\vspace{0.45em}\noindent\textbf{Assign CA = 1 when:}
\par\vspace{0.28em}\noindent The essential instructional content is correct and sufficiently represented, with no material factual, mathematical, or logical errors.
\par\vspace{0.45em}\noindent\textbf{Assign CA = 0 when:}
\par\vspace{0.28em}\noindent The animation contains a material error, contradicts the requirement, omits essential instructional content, or lacks sufficient visible evidence to verify a core requirement.
\par\vspace{0.28em}\noindent Do not penalize harmless stylistic differences or minor wording variations.
\par\vspace{0.45em}\noindent\textbf{DIMENSION 2: PEDAGOGICAL CLARITY (PC)}
\par\vspace{0.28em}\noindent PC is an integer between 0 and 100.
\par\vspace{0.28em}\noindent Evaluate how clearly the animation communicates its educational content.
\par\vspace{0.45em}\noindent\textbf{Consider:}
\begin{promptlist}
\item Logical organization.
\item Clarity of the explanatory sequence.
\item Readability of essential visual elements.
\item Coherence between equations, labels, diagrams, and explanations.
\item Whether the presented information supports understanding rather than causing confusion.
\end{promptlist}
\par\vspace{0.45em}\noindent\textbf{Scoring anchors:}
\par\vspace{0.45em}\noindent\textbf{0-20:}
\par\vspace{0.28em}\noindent Very difficult to understand; major confusion or missing instructional structure.
\par\vspace{0.45em}\noindent\textbf{21-40:}
\par\vspace{0.28em}\noindent Weak organization; substantial ambiguity.
\par\vspace{0.45em}\noindent\textbf{41-60:}
\par\vspace{0.28em}\noindent Partially understandable, with noticeable clarity problems.
\par\vspace{0.45em}\noindent\textbf{61-80:}
\par\vspace{0.28em}\noindent Generally clear and logically organized, with minor weaknesses.
\par\vspace{0.45em}\noindent\textbf{81-100:}
\par\vspace{0.28em}\noindent Highly clear, coherent, and pedagogically effective presentation.
\par\vspace{0.28em}\noindent Do not automatically assign a low PC score solely because the subject matter is advanced.
\par\vspace{0.45em}\noindent\textbf{DIMENSION 3: ENGAGEMENT (EN)}
\par\vspace{0.28em}\noindent EN is an integer between 0 and 100.
\par\vspace{0.28em}\noindent Evaluate the extent to which the animation's visible presentation supports viewer attention and interest.
\par\vspace{0.45em}\noindent\textbf{Consider:}
\begin{promptlist}
\item Effective use of visual emphasis.
\item Meaningful visual variation.
\item Appropriateness of the visual presentation for the educational objective.
\item Whether the visual presentation helps maintain attention.
\item Avoidance of unnecessary visual clutter.
\end{promptlist}
\par\vspace{0.45em}\noindent\textbf{Scoring anchors:}
\par\vspace{0.45em}\noindent\textbf{0-20:}
\par\vspace{0.28em}\noindent Very poor presentation; little meaningful visual engagement.
\par\vspace{0.45em}\noindent\textbf{21-40:}
\par\vspace{0.28em}\noindent Limited visual engagement.
\par\vspace{0.45em}\noindent\textbf{41-60:}
\par\vspace{0.28em}\noindent Moderately engaging but visually basic.
\par\vspace{0.45em}\noindent\textbf{61-80:}
\par\vspace{0.28em}\noindent Engaging and purposeful visual presentation.
\par\vspace{0.45em}\noindent\textbf{81-100:}
\par\vspace{0.28em}\noindent Highly engaging, visually coherent, and well suited to the instructional content.
\par\vspace{0.28em}\noindent Do not reward unnecessary decoration, excessive animation, or visual complexity that does not support learning.
\par\vspace{0.45em}\noindent\textbf{EVIDENCE REQUIREMENTS:}
\begin{promptlist}
\item Base all judgments on the provided frames and teaching requirement.
\item Do not assume unseen frames contain missing instructional content.
\item If the visual sample is insufficient to establish content accuracy, flag this uncertainty in the output.
\item Keep the explanations brief and factual.
\end{promptlist}
\par\vspace{0.45em}\noindent\textbf{OUTPUT REQUIREMENTS:}
\par\vspace{0.28em}\noindent Return valid JSON only.
\par\vspace{0.28em}\noindent CA must be 0 or 1. PC and EN must be integers from 0 to 100. Do not include Markdown or extra commentary.
\par\vspace{0.55em}\noindent\textbf{User.}
\par\vspace{0.35em}\noindent\textbf{Original teaching requirement:} \texttt{\{\{requirement\}\}}
\par\vspace{0.35em}\noindent\textbf{Structured teaching plan:} \texttt{\{\{teaching\_plan\}\}}
\par\vspace{0.35em}\noindent\textbf{Uniformly sampled video frames with timestamps:} \texttt{\{\{sampled\_frames\}\}}
\par\vspace{0.28em}\noindent Evaluate this educational animation independently.
\par\vspace{0.45em}\noindent\textbf{Return JSON:}
\par\vspace{0.25em}{\ttfamily\small \hspace*{0.0em}\{\\\hspace*{1.0em}"CA": 0,\\\hspace*{1.0em}"PC": 0,\\\hspace*{1.0em}"EN": 0,\\\hspace*{1.0em}"uncertain": false,\\\hspace*{1.0em}"evidence": \{"CA": "...", "PC": "...", "EN": "..."\}\\\hspace*{0.0em}\}\par}
\par\vspace{0.28em}\noindent Return JSON only.
\end{promptcard}


{\small
\begin{thebibliography}{54}
\providecommand{\natexlab}[1]{#1}
\providecommand{\url}[1]{\texttt{#1}}
\expandafter\ifx\csname urlstyle\endcsname\relax
  \providecommand{\doi}[1]{doi: #1}\else
  \providecommand{\doi}{doi: \begingroup \urlstyle{rm}\Url}\fi

\bibitem[Bar-Tal et~al.(2024)Bar-Tal, Chefer, Tov, Herrmann, Paiss, Zada,
  Ephrat, Hur, Liu, Raj, Li, Rubinstein, Michaeli, Wang, Sun, Dekel, and
  Mosseri]{bar2024lumiere}
Omer Bar-Tal, Hila Chefer, Omer Tov, Charles Herrmann, Roni Paiss, Shiran Zada,
  Ariel Ephrat, Junhwa Hur, Guanghui Liu, Amit Raj, Yuanzhen Li, Michael
  Rubinstein, Tomer Michaeli, Oliver Wang, Deqing Sun, Tali Dekel, and Inbar
  Mosseri.
\newblock {Lumiere}: A space-time diffusion model for video generation.
\newblock In \emph{SIGGRAPH Asia 2024 Conference Papers}, pages 1--11, 2024.

\bibitem[Belouadi et~al.(2024)Belouadi, Ponzetto, and
  Eger]{belouadi2024detikzify}
Jonas Belouadi, Simone~Paolo Ponzetto, and Steffen Eger.
\newblock {DeTikZify}: Synthesizing graphics programs for scientific figures
  and sketches with {TikZ}.
\newblock In \emph{Advances in Neural Information Processing Systems}, pages
  85074--85108. Curran Associates, Inc., 2024.

\bibitem[Blattmann et~al.(2023)Blattmann, Rombach, Ling, Dockhorn, Kim, Fidler,
  and Kreis]{blattmann2023alignyourlatents}
Andreas Blattmann, Robin Rombach, Huan Ling, Tim Dockhorn, Seung~Wook Kim,
  Sanja Fidler, and Karsten Kreis.
\newblock Align your latents: High-resolution video synthesis with latent
  diffusion models.
\newblock In \emph{Proceedings of the IEEE/CVF Conference on Computer Vision
  and Pattern Recognition}, pages 22563--22575, 2023.

\bibitem[Chai et~al.(2023)Chai, Zhuang, and Yan]{chai2023layoutdm}
Shang Chai, Liansheng Zhuang, and Fengying Yan.
\newblock {LayoutDM}: Transformer-based diffusion model for layout generation.
\newblock In \emph{Proceedings of the IEEE/CVF Conference on Computer Vision
  and Pattern Recognition}, pages 18349--18358, 2023.

\bibitem[Chen et~al.(2025)Chen, Shi, Huang, Li, Pan, Chen, and
  He]{chen2025visualedu}
Hao Chen, Tianyu Shi, Pengran Huang, Zeyuan Li, Jiahui Pan, Qianglong Chen, and
  Lewei He.
\newblock {VisualEDU}: A benchmark for assessing coding and visual
  comprehension through educational problem-solving video generation.
\newblock In \emph{Findings of the Association for Computational Linguistics:
  EMNLP 2025}, pages 16363--16394. Association for Computational Linguistics,
  2025.

\bibitem[Chen et~al.(2024{\natexlab{a}})Chen, Zhang, Zhou, and
  Chen]{chen2024lace}
Jian Chen, Ruiyi Zhang, Yufan Zhou, and Changyou Chen.
\newblock Towards aligned layout generation via diffusion model with aesthetic
  constraints.
\newblock In \emph{International Conference on Learning Representations}, pages
  9094--9111, 2024{\natexlab{a}}.

\bibitem[Chen et~al.(2024{\natexlab{b}})Chen, Lin, Schaerli, and
  Zhou]{chen2023teaching}
Xinyun Chen, Maxwell Lin, Nathanael Schaerli, and Denny Zhou.
\newblock Teaching large language models to self-debug.
\newblock In \emph{International Conference on Learning Representations}, pages
  8746--8825, 2024{\natexlab{b}}.

\bibitem[Chen et~al.(2026)Chen, Lin, and Shou]{chen2025code2video}
Yanzhe Chen, Kevin~Qinghong Lin, and Mike~Zheng Shou.
\newblock {C}ode2{V}ideo: A code-centric paradigm for educational video
  creation.
\newblock In \emph{Proceedings of the 43rd International Conference on Machine
  Learning}, pages 15370--15396. PMLR, 2026.

\bibitem[Feng et~al.(2023)Feng, Zhu, Fu, Jampani, Akula, He, Basu, Wang, and
  Wang]{feng2023layoutgpt}
Weixi Feng, Wanrong Zhu, Tsu-Jui Fu, Varun Jampani, Arjun Akula, Xuehai He,
  Sugato Basu, Xin~Eric Wang, and William~Yang Wang.
\newblock {LayoutGPT}: Compositional visual planning and generation with large
  language models.
\newblock In \emph{Advances in Neural Information Processing Systems}, pages
  18225--18250. Curran Associates, Inc., 2023.

\bibitem[Gehring et~al.(2025)Gehring, Zheng, Copet, Mella, Cohen, and
  Synnaeve]{gehring2025rlef}
Jonas Gehring, Kunhao Zheng, Jade Copet, Vegard Mella, Taco Cohen, and Gabriel
  Synnaeve.
\newblock {RLEF}: Grounding code {LLM}s in execution feedback with
  reinforcement learning.
\newblock In \emph{Proceedings of the 42nd International Conference on Machine
  Learning}, pages 19034--19055. PMLR, 2025.

\bibitem[{Google DeepMind}(2025)]{google2025gemma3card}
{Google DeepMind}.
\newblock {Gemma 3 12B Instruction-Tuned}: Official model card.
\newblock Hugging Face, 2025.

\bibitem[Gupta et~al.(2021)Gupta, Lazarow, Achille, Davis, Mahadevan, and
  Shrivastava]{gupta2021layouttransformer}
Kamal Gupta, Justin Lazarow, Alessandro Achille, Larry~S. Davis, Vijay
  Mahadevan, and Abhinav Shrivastava.
\newblock {LayoutTransformer}: Layout generation and completion with
  self-attention.
\newblock In \emph{Proceedings of the IEEE/CVF International Conference on
  Computer Vision}, pages 1004--1014, 2021.

\bibitem[Ho et~al.(2022)Ho, Chan, Saharia, Whang, Gao, Gritsenko, Kingma,
  Poole, Norouzi, Fleet, and Salimans]{ho2022imagenvideo}
Jonathan Ho, William Chan, Chitwan Saharia, Jay Whang, Ruiqi Gao, Alexey
  Gritsenko, Diederik~P. Kingma, Ben Poole, Mohammad Norouzi, David~J. Fleet,
  and Tim Salimans.
\newblock {Imagen Video}: High definition video generation with diffusion
  models.
\newblock \emph{arXiv preprint arXiv:2210.02303}, 2022.

\bibitem[Hong et~al.(2024)Hong, Zhuge, Chen, Zheng, Cheng, Wang, Zhang, Wang,
  Yau, Lin, Zhou, Ran, Xiao, Wu, and Schmidhuber]{hong2024metagpt}
Sirui Hong, Mingchen Zhuge, Jonathan Chen, Xiawu Zheng, Yuheng Cheng, Jinlin
  Wang, Ceyao Zhang, Zili Wang, Steven Yau, Zijuan Lin, Liyang Zhou, Chenyu
  Ran, Lingfeng Xiao, Chenglin Wu, and J\"{u}rgen Schmidhuber.
\newblock {MetaGPT}: Meta programming for a multi-agent collaborative
  framework.
\newblock In \emph{International Conference on Learning Representations}, pages
  23247--23275, 2024.

\bibitem[Huang et~al.(2024)Huang, He, Yu, Zhang, Si, Jiang, Zhang, Wu, Jin,
  Chanpaisit, Wang, Chen, Wang, Lin, Qiao, and Liu]{huang2024vbench}
Ziqi Huang, Yinan He, Jiashuo Yu, Fan Zhang, Chenyang Si, Yuming Jiang, Yuanhan
  Zhang, Tianxing Wu, Qingyang Jin, Nattapol Chanpaisit, Yaohui Wang, Xinyuan
  Chen, Limin Wang, Dahua Lin, Yu Qiao, and Ziwei Liu.
\newblock {VBench}: Comprehensive benchmark suite for video generative models.
\newblock In \emph{Proceedings of the IEEE/CVF Conference on Computer Vision
  and Pattern Recognition}, pages 21807--21818, 2024.

\bibitem[Inoue et~al.(2023)Inoue, Kikuchi, Simo-Serra, Otani, and
  Yamaguchi]{inoue2023layoutdm}
Naoto Inoue, Kotaro Kikuchi, Edgar Simo-Serra, Mayu Otani, and Kota Yamaguchi.
\newblock {LayoutDM}: Discrete diffusion model for controllable layout
  generation.
\newblock In \emph{Proceedings of the IEEE/CVF Conference on Computer Vision
  and Pattern Recognition}, pages 10167--10176, 2023.

\bibitem[Ji et~al.(2026)Ji, Qiu, Xin, Han, Chen, Zhang, Wang, and
  Yao]{ji2025eduvisagent}
Haonian Ji, Shi Qiu, Siyang Xin, Siwei Han, Zhaorun Chen, Dake Zhang, Hongyi
  Wang, and Huaxiu Yao.
\newblock From {EduVisBench} to {EduVisAgent}: A benchmark and multi-agent
  framework for reasoning-driven pedagogical visualization.
\newblock In \emph{International Conference on Learning Representations}, pages
  62341--62369, 2026.

\bibitem[Jyothi et~al.(2019)Jyothi, Durand, He, Sigal, and
  Mori]{jyothi2019layoutvae}
Akash~Abdu Jyothi, Thibaut Durand, Jiawei He, Leonid Sigal, and Greg Mori.
\newblock {LayoutVAE}: Stochastic scene layout generation from a label set.
\newblock In \emph{Proceedings of the IEEE/CVF International Conference on
  Computer Vision}, pages 9894--9903, 2019.

\bibitem[Kondratyuk et~al.(2024)Kondratyuk, Yu, Gu, Lezama, Huang, Schindler,
  Hornung, Birodkar, Yan, Chiu, Somandepalli, Akbari, Alon, Cheng, Dillon,
  Gupta, Hahn, Hauth, Hendon, Martinez, Minnen, Sirotenko, Sohn, Yang, Adam,
  Yang, Essa, Wang, Ross, Seybold, and Jiang]{kondratyuk2023videopoet}
Dan Kondratyuk, Lijun Yu, Xiuye Gu, Jose Lezama, Jonathan Huang, Grant
  Schindler, Rachel Hornung, Vighnesh Birodkar, Jimmy Yan, Ming-Chang Chiu,
  Krishna Somandepalli, Hassan Akbari, Yair Alon, Yong Cheng, Joshua~V. Dillon,
  Agrim Gupta, Meera Hahn, Anja Hauth, David Hendon, Alonso Martinez, David
  Minnen, Mikhail Sirotenko, Kihyuk Sohn, Xuan Yang, Hartwig Adam, Ming-Hsuan
  Yang, Irfan Essa, Huisheng Wang, David~A Ross, Bryan Seybold, and Lu Jiang.
\newblock {V}ideo{P}oet: A large language model for zero-shot video generation.
\newblock In \emph{Proceedings of the 41st International Conference on Machine
  Learning}, pages 25105--25124. PMLR, 2024.

\bibitem[Ku et~al.(2025)Ku, Chong, Leung, Shah, Yu, and
  Chen]{ku2025theoremexplainagent}
Max Ku, Cheuk~Hei Chong, Jonathan Leung, Krish Shah, Alvin Yu, and Wenhu Chen.
\newblock {TheoremExplainAgent}: Towards video-based multimodal explanations
  for {LLM} theorem understanding.
\newblock In \emph{Proceedings of the 63rd Annual Meeting of the {A}ssociation
  for {C}omputational {L}inguistics (Volume 1: Long Papers)}, pages 6663--6684.
  Association for Computational Linguistics, 2025.

\bibitem[Le et~al.(2022)Le, Wang, Gotmare, Savarese, and Hoi]{le2022coderl}
Hung Le, Yue Wang, Akhilesh~Deepak Gotmare, Silvio Savarese, and Steven
  Chu~Hong Hoi.
\newblock {CodeRL}: Mastering code generation through pretrained models and
  deep reinforcement learning.
\newblock In \emph{Advances in Neural Information Processing Systems}, pages
  21314--21328. Curran Associates, Inc., 2022.

\bibitem[Li et~al.(2019)Li, Yang, Hertzmann, Zhang, and Xu]{li2019layoutgan}
Jianan Li, Jimei Yang, Aaron Hertzmann, Jianming Zhang, and Tingfa Xu.
\newblock {LayoutGAN}: Generating graphic layouts with wireframe
  discriminators.
\newblock In \emph{International Conference on Learning Representations}, 2019.

\bibitem[Lian et~al.(2024)Lian, Shi, Yala, Darrell, and Li]{lian2024lvd}
Long Lian, Baifeng Shi, Adam Yala, Trevor Darrell, and Boyi Li.
\newblock {LLM}-grounded video diffusion models.
\newblock In \emph{International Conference on Learning Representations}, pages
  50207--50227, 2024.

\bibitem[Lin et~al.(2024)Lin, Zala, Cho, and Bansal]{lin2024videodirectorgpt}
Han Lin, Abhay Zala, Jaemin Cho, and Mohit Bansal.
\newblock {VideoDirectorGPT}: Consistent multi-scene video generation via
  {LLM}-guided planning.
\newblock In \emph{Conference on Language Modeling}, 2024.

\bibitem[Lin et~al.(2023)Lin, Guo, Sun, Yang, Lou, and
  Zhang]{lin2023layoutprompter}
Jiawei Lin, Jiaqi Guo, Shizhao Sun, Zijiang Yang, Jian-Guang Lou, and Dongmei
  Zhang.
\newblock {LayoutPrompter}: Awaken the design ability of large language models.
\newblock In \emph{Advances in Neural Information Processing Systems}, pages
  43852--43879. Curran Associates, Inc., 2023.

\bibitem[Lipman et~al.(2023)Lipman, Chen, Ben-Hamu, Nickel, and
  Le]{lipman2023flowmatching}
Yaron Lipman, Ricky T.~Q. Chen, Heli Ben-Hamu, Maximilian Nickel, and Matt Le.
\newblock Flow matching for generative modeling.
\newblock In \emph{International Conference on Learning Representations}, 2023.

\bibitem[Liu et~al.(2023{\natexlab{a}})Liu, Li, Wu, and Lee]{liu2023llava}
Haotian Liu, Chunyuan Li, Qingyang Wu, and Yong~Jae Lee.
\newblock Visual instruction tuning.
\newblock In \emph{Advances in Neural Information Processing Systems}, pages
  34892--34916. Curran Associates, Inc., 2023{\natexlab{a}}.

\bibitem[Liu et~al.(2023{\natexlab{b}})Liu, Zhu, Xiao, Fu, Han, Yang, and
  Ye]{liu2023rltf}
Jiate Liu, Yiqin Zhu, Kaiwen Xiao, Qiang Fu, Xiao Han, Wei Yang, and Deheng Ye.
\newblock {RLTF}: Reinforcement learning from unit test feedback.
\newblock \emph{Transactions on Machine Learning Research}, 2023{\natexlab{b}}.

\bibitem[Liu et~al.(2023{\natexlab{c}})Liu, Gong, and
  Liu]{liu2023rectifiedflow}
Xingchao Liu, Chengyue Gong, and Qiang Liu.
\newblock Flow straight and fast: Learning to generate and transfer data with
  rectified flow.
\newblock In \emph{International Conference on Learning Representations},
  2023{\natexlab{c}}.

\bibitem[Liu et~al.(2023{\natexlab{d}})Liu, Iter, Xu, Wang, Xu, and
  Zhu]{liu2023geval}
Yang Liu, Dan Iter, Yichong Xu, Shuohang Wang, Ruochen Xu, and Chenguang Zhu.
\newblock {G-Eval}: {NLG} evaluation using {GPT-4} with better human alignment.
\newblock In \emph{Proceedings of the 2023 Conference on Empirical Methods in
  Natural Language Processing}, pages 2511--2522. Association for Computational
  Linguistics, 2023{\natexlab{d}}.

\bibitem[Madaan et~al.(2023)Madaan, Tandon, Gupta, Hallinan, Gao, Wiegreffe,
  Alon, Dziri, Prabhumoye, Yang, Gupta, Majumder, Hermann, Welleck,
  Yazdanbakhsh, and Clark]{madaan2023selfrefine}
Aman Madaan, Niket Tandon, Prakhar Gupta, Skyler Hallinan, Luyu Gao, Sarah
  Wiegreffe, Uri Alon, Nouha Dziri, Shrimai Prabhumoye, Yiming Yang, Shashank
  Gupta, Bodhisattwa~Prasad Majumder, Katherine Hermann, Sean Welleck, Amir
  Yazdanbakhsh, and Peter Clark.
\newblock {Self-Refine}: Iterative refinement with self-feedback.
\newblock In \emph{Advances in Neural Information Processing Systems}, pages
  46534--46594. Curran Associates, Inc., 2023.

\bibitem[{Meta}(2024)]{meta2024llama31card}
{Meta}.
\newblock {Llama 3.1 8B Instruct}: Official model card.
\newblock Hugging Face, 2024.

\bibitem[Oli(2026)]{nabin2026manibench}
Nabin Oli.
\newblock {ManiBench}: A benchmark for testing visual-logic drift and syntactic
  hallucinations in {Manim} code generation.
\newblock \emph{arXiv preprint arXiv:2603.13251}, 2026.

\bibitem[Qian et~al.(2024)Qian, Liu, Liu, Chen, Dang, Li, Yang, Chen, Su, Cong,
  Xu, Li, Liu, and Sun]{qian2024chatdev}
Chen Qian, Wei Liu, Hongzhang Liu, Nuo Chen, Yufan Dang, Jiahao Li, Cheng Yang,
  Weize Chen, Yusheng Su, Xin Cong, Juyuan Xu, Dahai Li, Zhiyuan Liu, and
  Maosong Sun.
\newblock {C}hat{D}ev: Communicative agents for software development.
\newblock In \emph{Proceedings of the 62nd Annual Meeting of the {A}ssociation
  for {C}omputational {L}inguistics (Volume 1: Long Papers)}, pages
  15174--15186. Association for Computational Linguistics, 2024.

\bibitem[Rammuni~Silva et~al.(2026)Rammuni~Silva, Lotfi, Ihianle,
  Shahtahmassebi, and Bird]{silva2026manim}
Ravidu~Suien Rammuni~Silva, Ahmad Lotfi, Isibor~Kennedy Ihianle, Golnaz
  Shahtahmassebi, and Jordan~J. Bird.
\newblock Large language model approaches to educational video generation using
  {Manim}.
\newblock In \emph{Advances in Computational Intelligence Systems}, pages
  306--317. Springer, 2026.

\bibitem[Rodriguez et~al.(2025)Rodriguez, Puri, Agarwal, Laradji, Rodriguez,
  Rajeswar, Vazquez, Pal, and Pedersoli]{rodriguez2023starvector}
Juan~A. Rodriguez, Abhay Puri, Shubham Agarwal, Issam~H. Laradji, Pau
  Rodriguez, Sai Rajeswar, David Vazquez, Christopher Pal, and Marco Pedersoli.
\newblock {StarVector}: Generating scalable vector graphics code from images
  and text.
\newblock In \emph{Proceedings of the IEEE/CVF Conference on Computer Vision
  and Pattern Recognition}, pages 16175--16186, 2025.

\bibitem[Schick et~al.(2023)Schick, Dwivedi-Yu, Dessi, Raileanu, Lomeli,
  Hambro, Zettlemoyer, Cancedda, and Scialom]{schick2023toolformer}
Timo Schick, Jane Dwivedi-Yu, Roberto Dessi, Roberta Raileanu, Maria Lomeli,
  Eric Hambro, Luke Zettlemoyer, Nicola Cancedda, and Thomas Scialom.
\newblock {Toolformer}: Language models can teach themselves to use tools.
\newblock In \emph{Advances in Neural Information Processing Systems}, pages
  68539--68551. Curran Associates, Inc., 2023.

\bibitem[Shao et~al.(2024)Shao, Wang, Zhu, Xu, Song, Bi, Zhang, Zhang, Li, Wu,
  and Guo]{shao2024deepseekmath}
Zhihong Shao, Peiyi Wang, Qihao Zhu, Runxin Xu, Junxiao Song, Xiao Bi, Haowei
  Zhang, Mingchuan Zhang, Y.~K. Li, Y. Wu, and Daya Guo.
\newblock {DeepSeekMath}: Pushing the limits of mathematical reasoning in open
  language models.
\newblock \emph{arXiv preprint arXiv:2402.03300}, 2024.

\bibitem[Shinn et~al.(2023)Shinn, Cassano, Gopinath, Narasimhan, and
  Yao]{shinn2023reflexion}
Noah Shinn, Federico Cassano, Ashwin Gopinath, Karthik Narasimhan, and Shunyu
  Yao.
\newblock {Reflexion}: language agents with verbal reinforcement learning.
\newblock In \emph{Advances in Neural Information Processing Systems}, pages
  8634--8652. Curran Associates, Inc., 2023.

\bibitem[Shojaee et~al.(2023)Shojaee, Jain, Tipirneni, and
  Reddy]{shojaee2023ppocoder}
Parshin Shojaee, Aneesh Jain, Sindhu Tipirneni, and Chandan~K. Reddy.
\newblock Execution-based code generation using deep reinforcement learning.
\newblock \emph{Transactions on Machine Learning Research}, 2023.

\bibitem[Singer et~al.(2023)Singer, Polyak, Hayes, Yin, An, Zhang, Hu, Yang,
  Ashual, Gafni, Parikh, Gupta, and Taigman]{singer2023makeavideo}
Uriel Singer, Adam Polyak, Thomas Hayes, Xi Yin, Jie An, Songyang Zhang, Qiyuan
  Hu, Harry Yang, Oron Ashual, Oran Gafni, Devi Parikh, Sonal Gupta, and Yaniv
  Taigman.
\newblock {Make-A-Video}: Text-to-video generation without text-video data.
\newblock In \emph{International Conference on Learning Representations}, 2023.

\bibitem[{The Manim Community Developers}(2025)]{manim2025}
{The Manim Community Developers}.
\newblock {Manim -- Mathematical Animation Framework}.
\newblock Version 0.19.0, 2025.

\bibitem[Wang et~al.(2023)Wang, Wei, Schuurmans, Le, Chi, Narang, Chowdhery,
  and Zhou]{wang2023selfconsistency}
Xuezhi Wang, Jason Wei, Dale Schuurmans, Quoc Le, Ed Chi, Sharan Narang,
  Aakanksha Chowdhery, and Denny Zhou.
\newblock Self-consistency improves chain of thought reasoning in language
  models.
\newblock In \emph{International Conference on Learning Representations}, 2023.

\bibitem[Wei et~al.(2022)Wei, Wang, Schuurmans, Bosma, Ichter, Xia, Chi, Le,
  and Zhou]{wei2022cot}
Jason Wei, Xuezhi Wang, Dale Schuurmans, Maarten Bosma, Brian Ichter, Fei Xia,
  Ed Chi, Quoc~V Le, and Denny Zhou.
\newblock Chain-of-thought prompting elicits reasoning in large language
  models.
\newblock In \emph{Advances in Neural Information Processing Systems}, pages
  24824--24837. Curran Associates, Inc., 2022.

\bibitem[Wu et~al.(2025)Wu, Liang, Ge, Guo, Lu, Wang, Shan, and
  Luo]{wu2025plot2code}
Chengyue Wu, Zhixuan Liang, Yixiao Ge, Qiushan Guo, Zeyu Lu, Jiahao Wang, Ying
  Shan, and Ping Luo.
\newblock {P}lot2{C}ode: A comprehensive benchmark for evaluating multi-modal
  large language models in code generation from scientific plots.
\newblock In \emph{Findings of the {A}ssociation for {C}omputational
  {L}inguistics: {NAACL} 2025}, pages 3006--3028. Association for Computational
  Linguistics, 2025.

\bibitem[Wu et~al.(2024)Wu, Bansal, Zhang, Wu, Li, Zhu, Jiang, Zhang, Zhang,
  Liu, Awadallah, White, Burger, and Wang]{wu2023autogen}
Qingyun Wu, Gagan Bansal, Jieyu Zhang, Yiran Wu, Beibin Li, Erkang Zhu, Li
  Jiang, Xiaoyun Zhang, Shaokun Zhang, Jiale Liu, Ahmed~Hassan Awadallah,
  Ryen~W White, Doug Burger, and Chi Wang.
\newblock {AutoGen}: Enabling next-gen {LLM} applications via multi-agent
  conversations.
\newblock In \emph{Conference on Language Modeling}, 2024.

\bibitem[Yang et~al.(2025{\natexlab{a}})Yang, Li, Yang, Zhang, Hui, Zheng, Yu,
  Gao, Huang, Lv, Zheng, Liu, Zhou, Huang, Hu, Ge, Wei, Lin, Tang, Yang, Tu,
  Zhang, Yang, Yang, Zhou, Zhou, Lin, Dang, Bao, Yang, Yu, Deng, Li, Xue, Li,
  Zhang, Wang, Zhu, Men, Gao, Liu, Luo, Li, Tang, Yin, Ren, Wang, Zhang, Ren,
  Fan, Su, Zhang, Zhang, Wan, Liu, Wang, Cui, Zhang, Zhou, and
  Qiu]{yang2025qwen3}
An Yang, Anfeng Li, Baosong Yang, Beichen Zhang, Binyuan Hui, Bo Zheng, Bowen
  Yu, Chang Gao, Chengen Huang, Chenxu Lv, Chujie Zheng, Dayiheng Liu, Fan
  Zhou, Fei Huang, Feng Hu, Hao Ge, Haoran Wei, Huan Lin, Jialong Tang, Jian
  Yang, Jianhong Tu, Jianwei Zhang, Jianxin Yang, Jiaxi Yang, Jing Zhou,
  Jingren Zhou, Junyang Lin, Kai Dang, Keqin Bao, Kexin Yang, Le Yu, Lianghao
  Deng, Mei Li, Mingfeng Xue, Mingze Li, Pei Zhang, Peng Wang, Qin Zhu, Rui
  Men, Ruize Gao, Shixuan Liu, Shuang Luo, Tianhao Li, Tianyi Tang, Wenbiao
  Yin, Xingzhang Ren, Xinyu Wang, Xinyu Zhang, Xuancheng Ren, Yang Fan, Yang
  Su, Yichang Zhang, Yinger Zhang, Yu Wan, Yuqiong Liu, Zekun Wang, Zeyu Cui,
  Zhenru Zhang, Zhipeng Zhou, and Zihan Qiu.
\newblock {Qwen3} technical report.
\newblock \emph{arXiv preprint arXiv:2505.09388}, 2025{\natexlab{a}}.

\bibitem[Yang et~al.(2025{\natexlab{b}})Yang, Shi, Liu, Shui, Wang, Jing, Xu,
  Zhu, Li, Zhang, Liu, Nie, Cai, and Yang]{shi2025chartmimic}
Cheng Yang, Chufan Shi, Yaxin Liu, Bo Shui, Junjie Wang, Mohan Jing, Linran Xu,
  Xinyu Zhu, Siheng Li, Yuxiang Zhang, Gongye Liu, Xiaomei Nie, Deng Cai, and
  Yujiu Yang.
\newblock {ChartMimic}: Evaluating {LMM}'s cross-modal reasoning capability via
  chart-to-code generation.
\newblock In \emph{International Conference on Learning Representations}, pages
  26590--26646, 2025{\natexlab{b}}.

\bibitem[Yang et~al.(2024)Yang, Zhou, Wang, Cong, Han, Yan, Liu, Tan, Liu, Yu,
  Liu, Shi, and Sun]{yang2024matplotagent}
Zhiyu Yang, Zihan Zhou, Shuo Wang, Xin Cong, Xu Han, Yukun Yan, Zhenghao Liu,
  Zhixing Tan, Pengyuan Liu, Dong Yu, Zhiyuan Liu, Xiaodong Shi, and Maosong
  Sun.
\newblock {M}at{P}lot{A}gent: Method and evaluation for {LLM}-based agentic
  scientific data visualization.
\newblock In \emph{Findings of the {A}ssociation for {C}omputational
  {L}inguistics: {ACL} 2024}, pages 11789--11804. Association for Computational
  Linguistics, 2024.

\bibitem[Yao et~al.(2023)Yao, Zhao, Yu, Du, Shafran, Narasimhan, and
  Cao]{yao2023react}
Shunyu Yao, Jeffrey Zhao, Dian Yu, Nan Du, Izhak Shafran, Karthik Narasimhan,
  and Yuan Cao.
\newblock {ReAct}: Synergizing reasoning and acting in language models.
\newblock In \emph{International Conference on Learning Representations}, 2023.

\bibitem[Zala et~al.(2024)Zala, Lin, Cho, and Bansal]{zala2023diagrammergpt}
Abhay Zala, Han Lin, Jaemin Cho, and Mohit Bansal.
\newblock {DiagrammerGPT}: Generating open-domain, open-platform diagrams via
  {LLM} planning.
\newblock In \emph{Conference on Language Modeling}, 2024.

\bibitem[Zhang et~al.(2023)Zhang, Guo, Sun, Lou, and
  Zhang]{zhang2023graphiclayoutdiffusion}
Junyi Zhang, Jiaqi Guo, Shizhao Sun, Jian-Guang Lou, and Dongmei Zhang.
\newblock {LayoutDiffusion}: Improving graphic layout generation by discrete
  diffusion probabilistic models.
\newblock In \emph{Proceedings of the IEEE/CVF International Conference on
  Computer Vision}, pages 7226--7236, 2023.

\bibitem[Zheng et~al.(2023{\natexlab{a}})Zheng, Zhou, Li, Qi, Shan, and
  Li]{zheng2023layoutdiffusion}
Guangcong Zheng, Xianpan Zhou, Xuewei Li, Zhongang Qi, Ying Shan, and Xi Li.
\newblock {LayoutDiffusion}: Controllable diffusion model for layout-to-image
  generation.
\newblock In \emph{Proceedings of the IEEE/CVF Conference on Computer Vision
  and Pattern Recognition}, pages 22490--22499, 2023{\natexlab{a}}.

\bibitem[Zheng et~al.(2023{\natexlab{b}})Zheng, Chiang, Sheng, Zhuang, Wu,
  Zhuang, Lin, Li, Li, Xing, Zhang, Gonzalez, and Stoica]{zheng2023judging}
Lianmin Zheng, Wei-Lin Chiang, Ying Sheng, Siyuan Zhuang, Zhanghao Wu, Yonghao
  Zhuang, Zi Lin, Zhuohan Li, Dacheng Li, Eric Xing, Hao Zhang, Joseph
  Gonzalez, and Ion Stoica.
\newblock Judging {LLM}-as-a-judge with {MT-Bench} and {Chatbot Arena}.
\newblock In \emph{Advances in Neural Information Processing Systems}, pages
  46595--46623. Curran Associates, Inc., 2023{\natexlab{b}}.

\end{thebibliography}
}
\end{document}